%% file: main.tex
\documentclass{article}

\usepackage{pareto}

\usepackage[utf8]{inputenc}
\usepackage[T1]{fontenc}
\usepackage{booktabs}
\usepackage{longtable}
\usepackage{array}
\usepackage{amsfonts}
\usepackage{nicefrac}
\usepackage{microtype}
\usepackage{hyperref}
\usepackage{xcolor}
\usepackage[normalem]{ulem}
\usepackage{graphicx}
\usepackage{placeins}
\usepackage{float}
\usepackage{dblfloatfix}
\usepackage{amsmath}
\usepackage{tikz}
\usetikzlibrary{positioning, arrows.meta}
\usepackage[rightcaption]{sidecap}
\sidecaptionvpos{figure}{c}

\newcommand{\methodname}[0]{\textsc{AttuneBench}}

\graphicspath{{figures/}}

\hypersetup{
  pdftitle    = {AttuneBench EI Benchmark},
  pdfauthor   = {Kate M. Lubrano, Faisal Sayed, Ankita Rathod, Akshansh, Craver Corbyn Thomas-Smith, Mark E. Whiting, Karina Nguyen},
  pdfkeywords = {emotional intelligence, LLM evaluation, conversational benchmark, human-centered AI, empathy modeling},
}

\title{AttuneBench: A Conversation-Based Benchmark for LLM Emotional Intelligence}

\shorttitle{AttuneBench: EI Benchmark}

\author{%
    Kate M. Lubrano,\textsuperscript{1}\correspondingauthor{kate@pareto.ai}~
    Faisal Sayed,\textsuperscript{2}~
    Akshansh Akshansh,\textsuperscript{1}~
    Ankita Rathod,\textsuperscript{1}~
    Craver Corbyn Thomas-Smith,\textsuperscript{2}~
    Mark E. Whiting,\textsuperscript{1,3}\correspondingauthor{mark@pareto.ai}~
    Karina Nguyen\textsuperscript{2}\correspondingauthor{karina@thoughtfullab.com}\\[4pt]
   \small\textsuperscript{1}Pareto, California, US\qquad
   \small\textsuperscript{2}Thoughtful, California, US\qquad
   \small\textsuperscript{3}University of Pennsylvania, Pennsylvania, US\qquad
}

\date{May 2026}

\begin{document}
%% ════════════════════════════════════════════════════════════════════════════

\maketitle
\begin{abstract}
Emotional intelligence (EI), the ability to perceive, understand, and respond appropriately to others' emotional states, is central to human communication, and increasingly important to assess as LLMs assume conversational roles in everyday life. Existing EI benchmarks rely on synthetic prompts, single-turn cases, or third-party annotation. These approaches do not directly measure how models infer and respond to a participant's emotional state over the course of a real conversation. We introduce \methodname{}, a benchmark grounded in 200 genuine multi-turn human-model conversations in which participants conversed with anonymized LLMs and provided turn-by-turn annotations of their emotional state, the model's behavior, and their preferred responses. Across 11 evaluated models, we find that model rankings on emotion recognition, behavioral classification, preference prediction, and judged response quality are largely independent, indicating that emotionally intelligent behavior decomposes into separable capabilities. Preference alignment and response-quality judgments are substantially more model-discriminating than emotion-label accuracy. These results indicate that emotionally intelligent behavior requires predicting what kind of response a specific user wants in context, a distinction that aggregate scoring can obscure and that single-turn or synthetic formats cannot directly capture across turns. \methodname{} provides a framework for assessing each of these capabilities and for diagnosing model-specific strengths and failure modes in emotionally salient conversation.
\end{abstract}


%% Note: author contributions statement (TO ADD) — include in Acknowledgments
%% or as a footnote on the title page in camera-ready version.

%% ════════════════════════════════════════════════════════════════════════════
\section{Introduction}
%% ════════════════════════════════════════════════════════════════════════════ 
Benchmarks commonly used to assess progress toward general intelligence (e.g., MMLU~\citep{hendrycks2021mmlu}) largely neglect dimensions of social perception~\citep{kosinski2023theory}, emotion's high-dimensional structure~\citep{cowen2019passions}, and contextually appropriate interpersonal behavior~\citep{rashkin2019empathetic}, creating a gap where systems that perform well on traditional metrics may still fail in settings that require emotional awareness or empathy~\citep{fazzi2025emotions}.

Users are increasingly turning to LLMs for advice, emotional support, and personal reflection~\citep{anthropic2026guidance, bird2024genai_therapy}. In these contexts, success is less about factual correctness than about responding to emotional states with nuance. Recent survey evidence indicates that approximately 13\% of US adolescents and young adults have used LLMs for mental health advice, with the figure rising to 22\% among those aged 18--21~\citep{mcbain2025}. Growing research highlights substantial limitations, including inconsistent responses, hallucinated advice, and failure to meet established psychotherapy ethics standards. In high-risk contexts like suicidal ideation, models have demonstrated unreliable alignment that can lead to harmful responses and escalating behavior~\citep{moore2026}. Yet the benchmarks we use to track LLM progress in this space rely on synthetic vignettes and retrospective annotation, not the real conversations these systems actually have.

We introduce \methodname{}, a benchmark for evaluating emotional intelligence (EI) in LLMs through observable conversational behavior: recognizing emotional cues, tracking mood shifts across turns, calibrating tone and guidance, and adapting to user feedback. Because emotional interaction unfolds over time, \methodname{} is built from real multi-turn conversations with human participants.

Our contributions:
\begin{itemize}
\item We release the first open EI benchmark based on real multi-turn human--model conversations, with 200 conversations annotated for PANAS mood trajectories, observed-vs-preferred judgments, pairwise comparisons, and human-edited references.
\item Across 11 LLMs, we show that EI decomposes into weakly coupled skills: emotion tracking, behavioral judgment, response preference, and attuned generation.
\item We identify interaction-level failures, including drift across turns, sensitivity to topic and participant profile, and gaps between recognizing user needs and responding appropriately.
\end{itemize} 

%% ════════════════════════════════════════════════════════════════════════════
\section{Related Work}
%% ════════════════════════════════════════════════════════════════════════════

Prior work has evaluated emotional and social capabilities in LLMs across emotion recognition, empathy, and theory of mind~\citep{wang2023emotional, kosinski2023theory, sorin2024empathy, kusal2024emotion, chen2024emotionqueen}. However, existing benchmarks typically rely on synthetic data, single-turn formats, model-generated labels, or isolated psychometric probes rather than human-grounded conversational behavior.

EQ-Bench~\citep{paech2023eqbench} uses synthetic, single-prompt dialogues and correlates strongly with general reasoning benchmarks, making it unclear whether it isolates EI. EmpatheticDialogues~\citep{rashkin2019empathetic} uses crowdsourced conversations with assigned emotion labels, but evaluates retrieval or single responses rather than turn-level preferences. EmoBench~\citep{sabour2024emobench} and psychometric adaptations~\citep{empathybench, huang2024psychobench} test scenario reasoning or isolated traits, not how models respond across real emotional trajectories.

\methodname{} differs by combining genuine multi-turn conversations, participant-grounded annotation, continuous emotional-state measurement, and unified ground truth across evaluation dimensions.

%% ════════════════════════════════════════════════════════════════════════════
\section{Experiment Design}
\label{sec:design}
%% ════════════════════════════════════════════════════════════════════════════
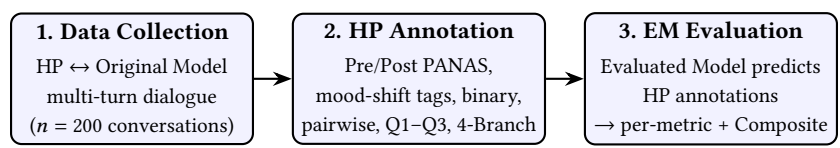
\begin{figure}[H]
\centering
\begin{tikzpicture}[
    node distance=0.5cm,
    box/.style={rectangle, rounded corners, draw, thick, minimum width=3.2cm, minimum height=1.5cm, inner ysep=3pt, align=center, font=\small, fill=blue!5},
    arrow/.style={-{Stealth[length=2.5mm]}, thick}
]
\node[box] (collect) {\textbf{1. Data Collection}\\[2pt]\footnotesize HP $\leftrightarrow$ Original Model\\\footnotesize multi-turn dialogue\\\footnotesize ($n=200$ conversations)};
\node[box, right=of collect] (annotate) {\textbf{2. HP Annotation}\\[2pt]\footnotesize Pre/Post PANAS,\\\footnotesize mood-shift tags, binary,\\\footnotesize pairwise, Q1--Q3, 4-Branch};
\node[box, right=of annotate] (evaluate) {\textbf{3. EM Evaluation}\\[2pt]\footnotesize Evaluated Model predicts\\\footnotesize HP annotations\\\footnotesize $\to$ per-metric + Composite};
\draw[arrow] (collect) -- (annotate);
\draw[arrow] (annotate) -- (evaluate);
\end{tikzpicture}
\caption{\methodname{} pipeline. HPs converse with an Original Model and annotate each turn (Phases 1--2); Evaluated Models are scored against the HP-grounded annotations (Phase 3).}
\label{fig:framework}
\end{figure}

\methodname{} is designed around three core questions: (1)~how accurately can LLMs infer a user's emotional state across a multi-turn conversation; (2)~how well do LLMs select or produce responses that align with human preferences in emotional contexts; and (3)~are these capabilities independent, or does strong performance on one predict strong performance on others?

\subsection{Interaction Structure}
The benchmark is built around multi-turn conversations. Human Participants (HPs) engage directly with an LLM during data collection; a secondary Evaluated Model (EM) is then scored on how well it reads emotional subtext and how closely its behavioral assessments align with HP judgments.

Conversations sample from a pool of Original Models (OMs) to limit single-model bias (Table~\ref{tab:om-models}). Though the per-OM distribution is unbalanced,  Pairwise Accuracy stratified by OM confirms headline rankings are stable across OMs (Appendix~\ref{sec:supp-om-stratification}).

\subsection{Task Formulation}
\paragraph{Topic Selection.} 
Conversation topics were assigned from a curated list containing 50 topics. Our topic taxonomy was informed by prior work on conversational behavior and emotional elicitation. The Conversations We Seek to Avoid~\citep{sun2020} guided the inclusion of sensitive domains (e.g., Money, Politics), while studies of everyday conversational content~\citep{sehulster2006, fried2015} informed the selection of more casual topics (e.g., Friends, Hobbies).

Emotional range emerged naturally from randomly assigned topics, individual preferences, and participant mood at session time, rather than through elicitation methods. Participants reported their attitudes toward all topics in advance, confirming coverage across positive and negative reactions.

\paragraph{Measurement of Emotional State.} 
To capture pre- and post-conversation emotional state, we adopt the Positive and Negative Affect Schedule (PANAS), a widely validated psychological instrument consisting of 20 affective descriptors rated on a Likert scale~\citep{watson1988}. PANAS was selected for its empirical grounding, dual-valence structure, and suitability for quantitative analysis.

Because PANAS items contain some overlap (e.g., \textit{jittery} and \textit{nervous} sit close in valence-arousal space), exact-label scoring would undervalue near-correct emotion predictions. Emotion prediction metrics incorporate partial credit based on continuous affective proximity (Section~\ref{sec:scoring}).

\paragraph{Annotation Metric Selection.}
To ensure theoretical grounding, the benchmark design is informed by the Mayer-Salovey-Caruso Four Branch Model~\citep{mayer2016ability, bruluna2021eimeasures}. EI dimensions are operationalized behaviorally through tasks requiring emotion inference, response evaluation, preference anticipation, and outcome prediction. HPs were provided with adapted branch definitions adjusted to reflect realistic expectations of model behavior (see Appendix~\ref{tab:fourdefs}).

Benchmark components map pragmatically to the Four Branch Model: Perceiving (mood-shift tags), Understanding (PANAS estimation), Facilitating Thought (binary judgments), and Managing Emotion (response drafting/preference).

\section{Methods}
\label{sec:method}
\subsection{Participant Selection}
Participants were recruited via an application form covering demographics, language, professional background, mental health, and willingness to discuss sensitive topics (Table~\ref{tab:intake}).

Eleven US-based native-English-speaking adult participants were accepted from a larger applicant pool, with diversity across gender, age, and educational background. Mental health representation was approximately balanced: six participants reported no diagnoses, while five reported one or more.

Following selection, participants completed a psychological profile (Table~\ref{tab:profile}) including the Ten-Item Personality Inventory (TIPI)~\citep{gosling2003tipi}, WHO-5 Well-Being Index~\citep{topp2015who5}, AQ-10 (Autistic Traits)~\citep{allison2012aq10}, Adult ADHD Self-Report Scale (ASRS-6)~\citep{kessler2007asrs}, PROMIS Depression-4, PROMIS Anxiety-4~\citep{kroenke2014promis_anxdep}, and PROMIS Sleep Disturbance-4~\citep{buysse2010promis_sleep}. They additionally reported LLM usage attitudes, recent major life events, and their baseline affective orientation toward each of the 50 conversation topics.

\subsection{Ethics and Data Privacy}
Before their first conversation, participants provided informed consent covering data collection, anonymized storage (Table~\ref{tab:data-incl}), and withdrawal rights (Appendix~\ref{app:consent}). Each participant received a randomized identifier; they were encouraged to omit or modify identifying details, and all conversations were manually reviewed to redact remaining PII.

Participants were told the LLM was not a therapy tool, encouraged to contact the project lead if uncomfortable, and reminded they could withdraw at any time. Before data collection, the team internally reviewed protocols against established human-subjects research principles~\citep{belmont1979}.

\subsection{Data Collection Process}
\paragraph{Conversation and Initial Assessments.}
Conversations were collected through a custom interface combining chat and annotation views. Before each conversation, HPs rated their current mood on a 7-point PANAS Likert scale. They were then assigned a randomly selected, non-repeating topic from the set of 50, with topic reassignment available as a distress safeguard. After accepting the topic, HPs were randomly connected to one of 8 undisclosed LLMs to reduce model-preference bias, and engaged in free-form multi-turn dialogue with the OM.

During the conversation, HPs could optionally tag any turn with Mood Shift Tags, selecting a PANAS emotion and 7-point intensity to mark emotional shifts or strong affect. Multiple tags per turn were allowed. Each conversation required at least five human-model exchanges, with no maximum, and continued until a natural stopping point. All conversations and annotations were completed in a single session to preserve live emotional reactions.

\paragraph{Conversation-Level Annotations.}
Upon submission, HPs completed a post-conversation PANAS, answered conversation-wide questions (Table~\ref{tab:convwide}), and rated OM performance on the Four Branch dimensions using definitions adapted to our purposes (Table~\ref{tab:fourdefs}). Conversation transcripts were available for reference while answering.

\paragraph{Turn-Level Annotations.}
At each turn, HPs reviewed and refined their live mood-shift tags and completed binary (Yes/No) evaluation questions drawn from a 36-question set, with at least 10 non-NA responses required per turn. To reduce annotator fatigue across the full question pool, an LLM (GPT-4o-mini) suggested a relevant subset, which HPs could revise. Each applicable question received two judgments: observed behavior and preferred behavior. This dual-label structure surfaces alignment or mismatch between model behavior and HP preference.

After binary annotation, the OM was given the HP's preference signals and generated a second, preference-informed response. HPs were then presented with both the original and alternate responses and asked to draft their own ``golden'' response, drawing on either or both as desired but not copying verbatim. HPs then completed pairwise comparisons across all three response variants, selecting overall preference and answering three targeted comparison questions per pair. This sequence was repeated for each turn, with submission occurring only after all turns were fully annotated.

\subsection{Runner and Scorer}
\paragraph{Evaluated Model Runner.}
The EM steps through each conversation turn-by-turn, generating predictions compared against HP annotations as ground truth. It receives a system prompt with task structure, HP/OM roles, and mode-specific instructions, with question text inline at each turn. Three modes exist: \textbf{Default} (predictions grounded in the conversation alone), \textbf{Verbose} (additionally captures the EM's reasoning traces), and \textbf{Omniscient} (provides the HP's pre-conversation PANAS and psychometric profile to the EM).  

 Processing sequentially without future-turn access, each turn comprises five steps: (1) the EM drafts a response before observing the OM, (2) a judge LLM scores the draft on four 1--7 quality dimensions, (3) the EM predicts HP's Mood Shift Tags, (4) the EM predicts observed-OM and preferred-HP labels for each binary question (first-person and observer phrasings), and (5) the EM ranks the three anonymized response variants from HP's perspective. After all turns, the EM makes two conversation-wide predictions: the HP's post-conversation evaluation responses (Four Branch ratings, goal identification, emotion clarity, OM fit) and post-conversation PANAS. Full prompt templates are available in the public repository.

\paragraph{Scoring.}
\label{sec:scoring}
Per-conversation JSON outputs are compared against human annotations to derive the metrics below. Full per-metric definitions are in Table~\ref{tab:metrics}.

\textit{Turn-level:} \textbf{Emotion F1} and \textbf{Emotion VA} (PANAS tag prediction: exact F1 and F1 with valence-arousal partial credit, respectively), \textbf{Binary OM/HP Accuracy} (behavioral classification, observer- vs HP-perspective), \textbf{Pairwise Accuracy} and \textbf{Kendall~$\tau$} (preference ranking across original, alternate, and golden variants), and \textbf{Draft Judge} (LLM-rated draft quality, supplementary).
\textit{Conversation-wide:} \textbf{Four Branch MAE}, \textbf{PANAS B-Adj} (bias-adjusted post-conversation PANAS prediction), and the \textbf{Composite} (0--100 aggregate; emotion 24\%, evaluation 49\%, holistic 27\%).

Metric scores are computed per conversation and aggregated across the evaluation sample. Emotion VA uses continuous valence-arousal distance for partial credit (Section~\ref{sec:design}).
Each PANAS item is mapped to a valence-arousal embedding via the NRC VAD Lexicon~\citep{mohammad2018, posner2005circumplex}; label similarity is normalized inverse Euclidean distance in that space, with the normalization constant set to the maximum pairwise distance across all 20 items. 

The per-conversation Composite Score (0--100) is $100\,(0.24\,E + 0.49\,V + 0.27\,H)$, where $E$ averages Emotion F1 and Emotion VA; $V$ averages Binary OM, Binary HP, and Pairwise Accuracy; and $H$ averages PANAS B-Adj, Four Branch, and the conversation-wide questions $\overline{Q}=\text{mean}(Q_1,Q_2,Q_3,Q_{3b})$. Conversation-level scores are averaged across the dataset for model-level values. Kendall~$\tau$ and Draft~Judge are reported but not Composite inputs; weights emphasize turn-level verification ($V$) over emotion tracking ($E$, $\eta^2=0.003$ on Emotion~F1).
%% ════════════════════════════════════════════════════════════════════════════
\section{Results}
\label{sec:results}
%% ════════════════════════════════════════════════════════════════════════════
We evaluated eleven models: Opus~4.7, Opus~4.6, Sonnet~4.6, and Haiku~4.5 (Anthropic~\citep{anthropic2026opus47, anthropic2026claude46, anthropic2025haiku45}); Gemini~3.1~Pro (Google DeepMind~\citep{google2026gemini31}); GPT-5.5 and GPT-5.4 (OpenAI~\citep{openai2026gpt55, openai2026gpt54}); Mistral~Large (Mistral~AI~\citep{mistral2024large}); Grok~4 (xAI~\citep{xai2025grok4}); Qwen~2.5~72B (Alibaba~\citep{yang2024qwen25}); and MiMo~v2~Pro (Xiaomi~\citep{xiaomi2026mimo}), spanning flagship and mid-tier models across major commercial providers. Per-model headline metrics for Default Mode ($n = 200$ conversations per model) are summarized in Figure~\ref{fig:composite-pairwise}; full per-metric results in Appendix~\ref{sec:supp-full-tables}.

\begin{figure}[H]
\centering
\includegraphics[width=0.9\linewidth]{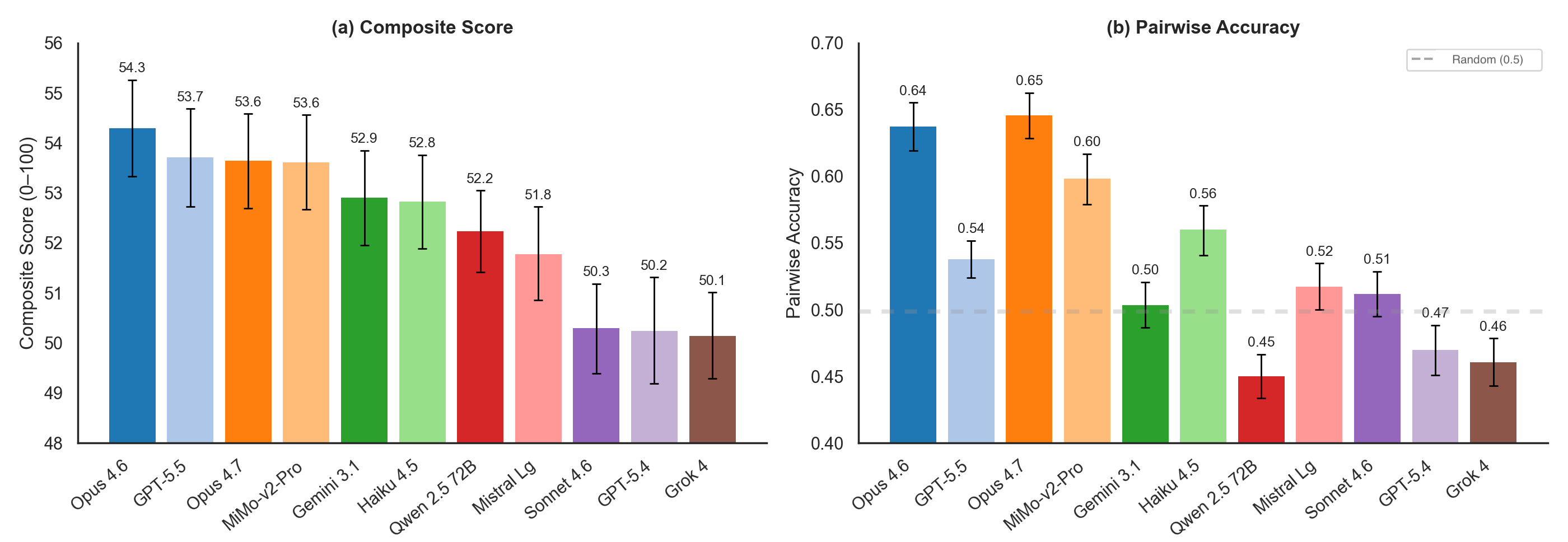}
\caption{Per-model Composite (left) and Pairwise Accuracy (right). Composite range is narrow (4 pt) but 35/55 model pairs are distinguished; Pairwise spans wider (47/55). Error bars: 95\% percentile bootstrap CIs (10{,}000 resamples; per-conversation, $n = 200$). Dashed line in (b) = chance (1/2).}
\label{fig:composite-pairwise}
\end{figure}

Models compress into a narrow Composite range, but diverge sharply by metric: aggregate scoring obscures statistically significant cross-metric divergence in what each model is good at. Composite Scores span a 4.2-point range (50.1--54.3), with Opus~4.6 at the top and Grok~4 at the bottom. Despite this narrow range, 35/55 model pairs are reliably distinguished on the Composite (\S~\ref{sec:analysis}).

\paragraph{Preference Prediction.}
The Opus family leads preference ranking, Opus~4.7 first on both Pairwise Accuracy (0.646) and Kendall~$\tau$ ($+$0.339), Opus~4.6 second (0.637, $+$0.301); MiMo~v2~Pro takes third (0.598, $+$0.213). The two Opus generations are statistically indistinguishable after correction. But 5/11 models produce \emph{negative} mean Kendall~$\tau$. Their preference orderings are anti-aligned with human ground truth (GPT-5.4: $-0.135$; Qwen~2.5~72B: $-0.127$; Grok~4: $-0.125$; Sonnet~4.6: $-0.030$; Gemini~3.1~Pro: $-0.015$), spanning providers and capability tiers.

\paragraph{Binary Classification.}
Mistral~Large leads both metrics (Binary OM: 0.862; Binary HP: 0.827), while Opus~4.7 ranks last on Binary HP Accuracy (0.767) and Opus~4.6 second-last (0.774), despite both topping the Composite and Pairwise rankings. All models reach Binary~OM~$\geq 0.83$, indicating OM-perspective assessment is robust to preference-prediction skill; Binary~HP spans 0.767--0.827, and every model scores higher on OM than HP (perspective gap, \S~\ref{sec:analysis}).

\paragraph{Emotion Prediction.}
Gemini~3.1~Pro (0.278) and Haiku~4.5 (0.276) lead on Emotion VA, while Mistral~Large (0.227) trails. Emotion F1 shows limited model-level differentiation ($\eta^2 = 0.003$), with Qwen~2.5~72B a clear outlier at 0.106 versus 0.133--0.141 for the remaining ten models and the only model with significant pairwise differences. Tag-vs-intensity decomposition in Appendix~\ref{sec:supp-emotion-intensity}.

\paragraph{Draft Quality.}
Draft Judge scores show the widest absolute spread ($\eta^2 = 0.299$), with Opus~4.6 highest (0.844) and Qwen~2.5~72B substantially lower (0.691). Because these reflect a single judge model (Mistral~Large), they are treated as supplementary.

%% ════════════════════════════════════════════════════════════════════════════
\section{Analysis}
\label{sec:analysis}
All primary metrics except Emotion~F1 reliably distinguish models (Kruskal-Wallis, $p < 0.01$ each). Emotion~F1 falls short of significance ($p = 0.099$) and should be interpreted conservatively throughout. The Composite ranking reads better as tiers than a strict linear order: 20 of 55 pairwise Wilcoxon contrasts (Holm-Bonferroni-corrected; 55 unique pairs of the 11 models) are non-significant, forming three rough tiers (Fig.~\ref{fig:composite-pairwise}). The most discriminating primary metric is Pairwise Accuracy ($\eta^2 = 0.202$). Kendall~$\tau$ is nearly redundant with Pairwise Accuracy ($r = 0.891$). Pooled tests treat conversations as independent; recomputing rankings within each HP's conversations separately (Appendix~\ref{sec:supp-robustness}) confirms they hold within most HPs (e.g., Opus leads Pairwise for 7/11).

\subsection{Conversation-Wide Metrics}
Post-conversation metrics, including Four Branch ratings and conversation-wide comprehension questions (Q1--Q3), reveal capability profiles that diverge substantially from the turn-based rankings.

\paragraph{Four Branch Ratings.}
Qwen~2.5~72B (81.6\% normalized), Mistral~Large (81.4\%), and GPT-5.5 (81.2\%) lead, while Sonnet~4.6 (69.8\%) and Opus~4.7 (73.7\%) rank lowest, with Sonnet's Understanding branch MAE approaching the 2.0--3.0 chance range. The Composite leaders Opus~4.6, Opus~4.7, and MiMo~v2~Pro all rank in the bottom third on Four-Branch (76.3\%, 73.7\%, and 77.2\% respectively), a clear inversion of the Composite ranking. Rating the quality of EI in a conversation appears to reward different capabilities than preference ranking or binary classification.

Across all models, ``Understanding'' is consistently the hardest branch (highest MAE) and ``Perceiving'' consistently the easiest, indicating a shared gradient in how difficult each EI dimension is to assess, not a model-specific pattern.

\paragraph{Conversation-Wide Comprehension (Q1--Q3).}
Q1 Goal Identification (set overlap with HP-reported goals) and Q2 Emotion Clarity (exact match of HP's clarity rating) reorder the rankings further: leaders include Qwen~2.5~72B and Gemini~3.1~Pro, while the new flagships (Opus~4.7, GPT-5.5) invert their Composite ranking. GPT-5.4 sits near chance on Q2 (22.5\% vs 25\% chance), indicating systematic miscalibration of HP-clarity self-reports. Full per-question rankings in Figure~\ref{fig:supp-def-post-grid}.

On Q3 Conversational Fit (exact match on a 4-point scale; chance = 25\%), GPT-5.5 (46.5\%), Opus~4.6 (45.5\%), and Grok~4 (45.0\%) lead, while Mistral~Large (24.5\%) is essentially at chance and substantially worse on ordinal distance than all other models, suggesting it systematically disagrees with the HP's quality rating in the wrong direction. A striking finding is that Sonnet~4.6 records exactly 0.0\% and Opus~4.7 effectively 0.0\% (0.0003) on Q3 Follow-up: when asked to identify what felt off in lower-rated conversations, neither model returns a correct answer in any conversation. The pattern is shared by both Anthropic flagships in the latest generation pair (Sonnet~4.6 and Opus~4.7) but \emph{not} by other Anthropic models in the same family (Opus~4.6 or Haiku~4.5). Per-model rates for Q3 Follow-up are reported in Figure~\ref{fig:supp-def-post-grid}.

Together, these results reveal two distinct capability profiles among the leading models. Some models (Qwen, Mistral, Gemini) excel at holistic comprehension but lag on turn-level metrics. Others (Opus~4.7, Opus~4.6, MiMo) show the reverse, with strong turn-level performance but weaker conversation-wide synthesis. GPT-5.5 is the only model in the top tier of both, suggesting top-tier preference ranking and top-tier holistic comprehension are not strictly trade-offs. Qwen is the starkest example of the holistic-only route, leading on Four-Branch and tying for the best Q3 Follow-up score despite the lowest Pairwise Accuracy of any model. Multiple profiles can yield a competitive Composite Score, underscoring why per-metric reporting is necessary to disambiguate the strategies underlying a top-tier headline number.   

\subsection{Metric Structure}
\paragraph{Metric Correlations.}
Metrics fall into three groups measuring distinct things. Emotion~F1 and Emotion~VA are moderately correlated ($r = 0.615$), sharing the emotion prediction domain but differing on partial credit. Binary~OM and Binary~HP~Accuracy are also moderately correlated ($r = 0.601$). Pairwise~Accuracy and Kendall~$\tau$ are nearly redundant ($r = 0.891$) and treated as a single signal in interpretation. Full Spearman matrix in Appendix~\ref{sec:supp-corr}.

The most important cross-cluster pattern is a \emph{negative} correlation between Pairwise Accuracy and Binary~HP~Accuracy ($r = -0.097$, $p < 0.001$). The two Opus models illustrate this most starkly: they lead all models on Pairwise Accuracy while ranking last (Opus~4.7) and second-last (Opus~4.6) on Binary~HP~Accuracy. Preference ranking and first-person binary classification appear to require distinct capabilities.

\begin{SCfigure}[1.0][!ht]
\centering
\includegraphics[width=0.5\linewidth]{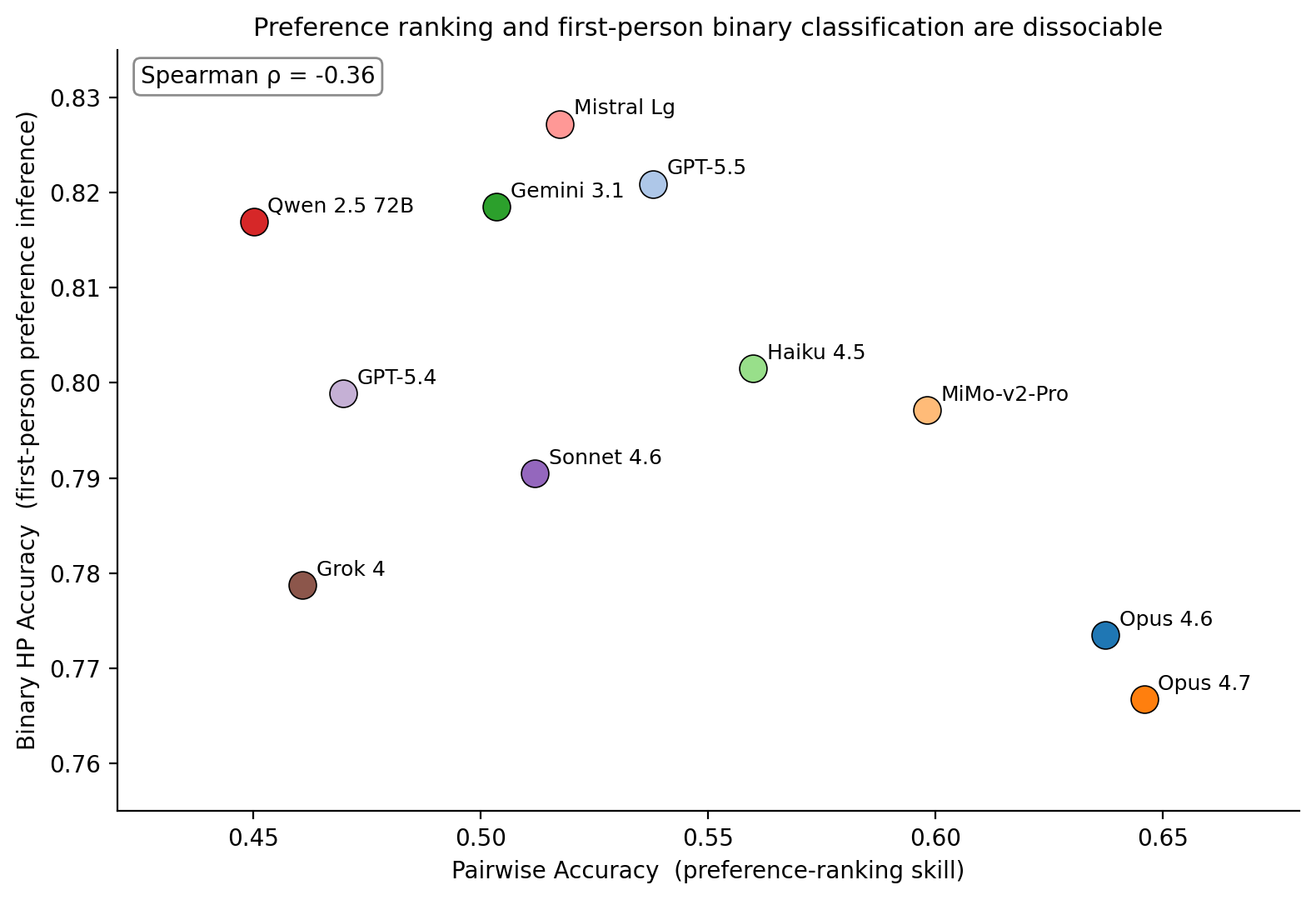}
\caption{Pairwise Accuracy versus Binary~HP~Accuracy across the 11 evaluated models (model-level Spearman $\rho = -0.36$; conversation-level Pearson $r = -0.097$, $p < 0.001$ in main text). The two capabilities are inversely related: the Opus family (bottom-right; high Pairwise, low Binary HP) leads preference ranking but ranks last on first-person binary classification, while Mistral~Large and GPT-5.5 (high Binary HP, mid Pairwise) show the opposite profile.}
\label{fig:pairwise-vs-binary-hp}
\end{SCfigure}

Draft~Judge is only weakly correlated with all other metrics ($r \leq 0.248$), consistent with it measuring response surface quality, not the same underlying construct.

\paragraph{Perspective Gap.}
All eleven models score higher on Binary~OM (observer-perspective) than on Binary~HP (first-person), $p < 0.001$ each, with gaps of 0.030--0.076. 

\subsection{Variance Sources}
\paragraph{Mode Effects.}
Neither Omniscient (HP psychometric profile at evaluation) nor Verbose (reasoning traces) mode improves Composite performance for most models. Notable exceptions: Opus~4.7 in Omniscient (Binary~HP $\Delta = -0.074$; Draft~Judge $0.842 \to 0.506$, likely an output-format artifact); Mistral~Large in Verbose (preference prediction $\Delta = -2.06$, $p < 0.001$). Draft~Judge improves for Mistral, Qwen, and Grok in Verbose without Pairwise gains, cautioning against Draft~Judge as an overall-performance proxy. Per-model deltas in \S~\ref{sec:supp-mode-detail}.

\paragraph{Easy vs. Hard Conversations.}
Conversation difficulty (operationalized as median-split of conversations by cross-model mean Composite) is one of the largest sources of variance in the dataset. Every model shows a significant performance gap (approximately 8--14 Composite points; $p < 0.001$ all models), but the relative ordering of models is preserved across difficulty levels, confirming that difficulty is approximately orthogonal to model quality. Turn count ($\leq 6$ vs.\ $>6$) shows no significant effect after correction (Appendix~\ref{sec:supp-length}).

\paragraph{Diagnosis Subgroups.}
The clearest participant-level finding is that models are substantially worse at tracking the affective states of participants with mental health diagnoses. Pooled across all eleven models, Emotion~VA drops from 0.310 for neurotypical participants to 0.202 for those with any diagnosis, a large effect ($\Delta = -0.109$, $|r| = 0.307$) consistent across 11/11 models. PANAS appears less suited to the affective expression patterns of participants with anxiety/depression or ASD/ADHD.

\begin{SCfigure}[1.0][!ht]
\centering
\includegraphics[width=0.5\linewidth]{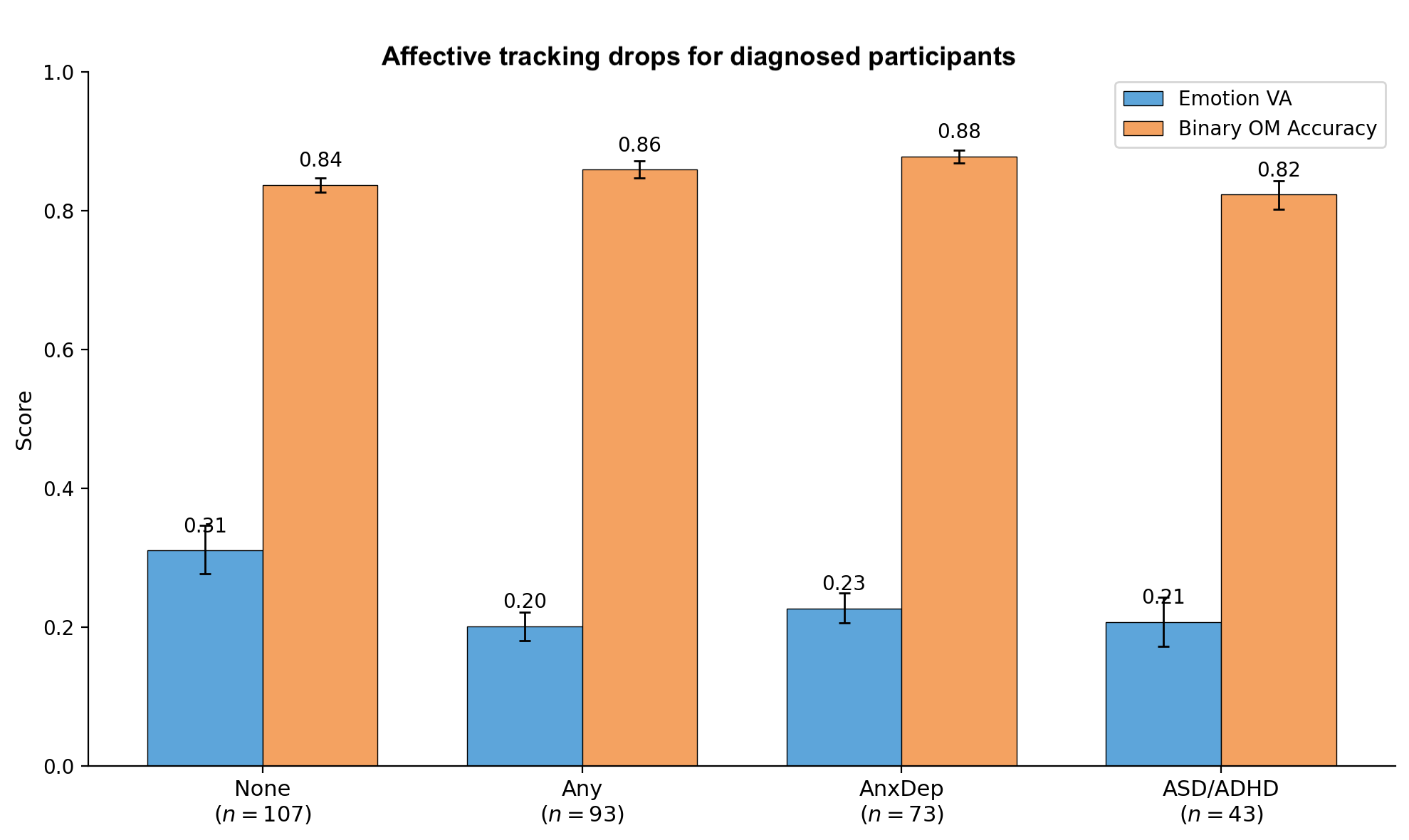}
\caption{Affective tracking and behavioral assessment by participant diagnosis group. Emotion~VA drops sharply for diagnosed participants (large effect, 11/11 models consistent), while Binary~OM~Accuracy moves in the opposite direction for anxiety/depression participants --- indicating that the two metrics capture qualitatively different aspects of participant-related variance. Error bars: 95\% percentile bootstrap CIs (10{,}000 resamples; resampling unit = conversation within subset).}
\label{fig:diagnosis-effects}
\end{SCfigure}

Binary~OM~Accuracy moves the opposite way (higher for anxiety/depression participants: 0.878 vs.\ 0.837, 11/11 models, Figure~\ref{fig:diagnosis-effects}), likely reflecting more behaviorally unambiguous features rather than greater model capability; ASD/ADHD shows a reversed Binary~HP pattern, indicating the two diagnostic groups differ qualitatively. Per-model breakdown in Appendix~\ref{sec:supp-diagnosis-permodel}.

\paragraph{Topic Category.}
Topic effects are modest overall ($\eta^2 \leq 0.022$); Romantic Relationships is the consistent outlier (lowest Composite, Emotion~VA, and Emotion~F1; full breakdown in Appendix~\ref{sec:supp-topic-permodel}).

\paragraph{Pre-conversation Mood (PANAS Shift Groups).}
Mood trajectory predicts Emotion~VA performance ($\eta^2 = 0.042$); the Negative-shift group ($n = 40$) is hardest to track affectively (mean Emotion~VA 0.195 vs 0.287 Positive / 0.256 Stable), yet yields the highest Pairwise Accuracy (0.558), mirroring the diagnosis dissociation. Full breakdown in Appendix~\ref{sec:supp-panas-shift}.

\begin{SCfigure}[1.0][!ht]
\centering
  \includegraphics[width=0.5\linewidth]{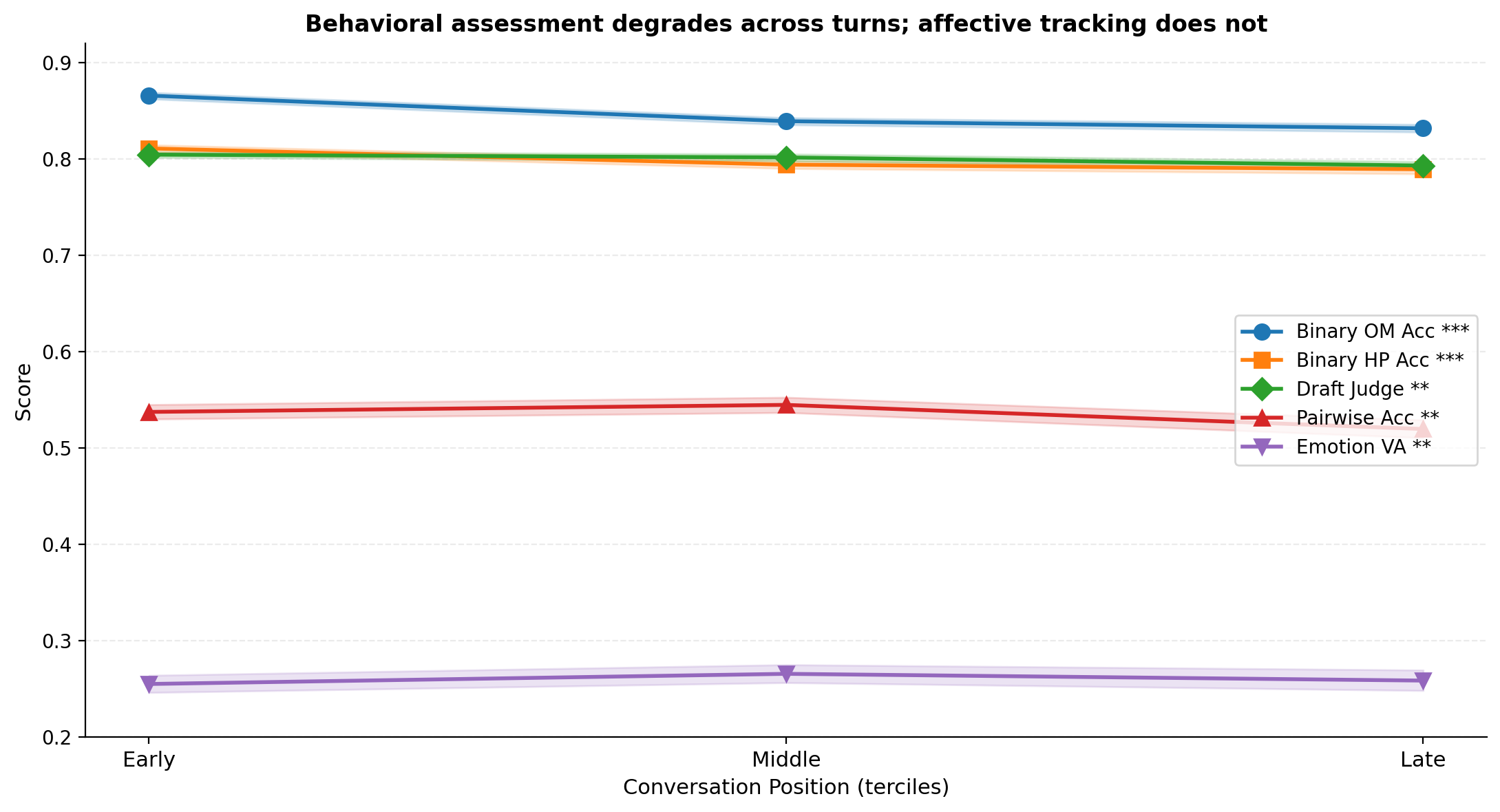}
\caption{Per-third metric drift, pooled across 11 models ($n = 2200$ paired conversation means per metric). Behavioral metrics (Binary~OM, Binary~HP, Draft~Judge) degrade significantly from early to late turns; Emotion~VA shows a small middle-third peak instead. Shaded bands: 95\% percentile bootstrap CIs (10{,}000 resamples; unit = (model, conversation), $n = 2200$). Significance markers in legend reflect Friedman tests across thirds.}
\label{fig:turn-drift}
\end{SCfigure}

\paragraph{Conversational Dynamics.}
Performance is not stable across conversation turns. Splitting turns into early, middle, and late thirds, Binary~OM~Accuracy degrades significantly from early to late turns for nine of eleven models (pooled drop $0.866 \to 0.832$, $p < 0.001$; exceptions Qwen~2.5~72B and Opus~4.7); Binary~HP and Draft~Judge show smaller drops in the same direction, while Emotion~VA peaks in the middle third ($\chi^2(2) = 12.39$, $p = 0.002$; Figure~\ref{fig:turn-drift}). Accumulated context degrades behavioral-assessment reliability (likely from topic drift or growing ambiguity), while affective inference remains stable. Within-model Friedman tests do not survive correction.

\subsection{Human Behavior Alignment}
Higher preference alignment correlates with Binary~HP~Accuracy ($r = +0.245$) and Composite ($r = +0.168$) but not with Pairwise Accuracy ($r = -0.041$, ns); preference alignment and pairwise ranking are largely decoupled. Opus~4.6 is the only model with a negative HP~Gap ($-0.021$), slightly preferring human-edited responses; all others favor their own draft. Qwen~2.5~72B is an outlier on the opposite axis: highest surface similarity to original responses (Sim$_\text{orig} = 0.208$) but lowest draft binary alignment (0.768), showing stylistic similarity does not predict preference alignment.

\subsection{Human Baseline (Pilot)}
Three annotators predicted the original HPs' annotations on a 7-conversation subset. Annotators reach 0.500--0.722 Pairwise Accuracy (the weakest annotator at chance, the strongest above the per-conversation best EM mean of 0.665) and 0.744--0.814 Binary HP Accuracy, bracketing the EM range. Inter-rater Cohen's $\kappa$ on Binary HP is +0.38 to +0.51 (moderate). Full pilot in Appendix~\ref{sec:supp-human-baseline}; multi-rater characterization across the full 200 is left to future work.

%% ════════════════════════════════════════════════════════════════════════════
\section{Limitations and Future Work}
\label{sec:limitations}
The pool was small and participation imbalanced; future rounds should expand the pool and cap per-participant submissions. Restricting participants to native English speakers in the United States reduced linguistic and cultural variability, but limits generalizability. Demographic and diagnostic data were self-reported, and conversations may reflect selective disclosure, topic avoidance, or retrospective experiences framed as present.

Adaptive topic assignment was not implemented due to resource constraints, though future work could better balance emotional contexts. Turn-level annotation is labor-intensive, and participants often preferred shorter conversations. As a text-only benchmark, this work cannot capture tone, facial expression, body language, or many forms of hesitation; future extensions could incorporate audio/video or behavioral signals. Expanded discussion in Appendix~\ref{sec:supp-methods-extended}.

\paragraph{Intended Use and Broader Impact.}
\methodname{} is intended for diagnostic evaluation of LLM emotional capabilities in research and development; it is not designed to validate models for clinical mental-health applications, to inform deployment decisions in safety-critical contexts, or to be fine-tuned on (which compromises its diagnostic value). High scores indicate accurate prediction of participant preferences in 200 sampled conversations, not validated safety in clinical or vulnerable-population settings; pair benchmark scores with domain-specific safety review.

\section{Benchmark Usage}
\label{sec:usage}
%% Replace repo URL with https://github.com/Thoughtful-Lab/AttuneBench.git when public
The \methodname{} dataset, evaluation runner, and scoring scripts are publicly available at \url{https://github.com/Thoughtful-Lab/attunebench}. A public leaderboard is at \url{https://public.attunebench.com/}. Code is released under the MIT License; dataset under CC~BY~4.0. The runner and scoring pipeline support future data releases without protocol change.

\paragraph{Running an Evaluation.}
The runner accepts structured JSON conversation files and evaluates any model accessible via a standard chat API with no fine-tuning required. Default Mode evaluation requires approximately \textbf{4 API calls to the EM per turn} plus one judge-model call for draft scoring, totaling \textbf{$\sim$24 calls per conversation} on average. Per-model cost and runtime are reported in \S~\ref{app:cost}.

Results are JSON output files passed to scoring scripts to produce per-metric and aggregate scores. Full configuration and replication instructions are in repository documentation. Composite Score is reported on a 0--100 scale; all other primary metrics on 0--1 (Kendall~$\tau$, Perspective Gap: $-1$ to $+1$).

\paragraph{Usage Guidelines.}
These guidelines are not technically enforced but constitute the expected standard for published results. To ensure comparability and diagnostic value, we request:
\begin{itemize}
  \item \textbf{No training on benchmark data.} EMs must not be fine-tuned on any conversations, annotations, or metadata from this dataset.
  \item \textbf{No access to human annotations during evaluation.} The EM must not access binary judgments, pairwise preferences, PANAS labels, or any other ground-truth annotations.
  \item \textbf{Report evaluation mode.} Specify Default, Omniscient, or Verbose. Default Mode is the primary benchmark; other modes should not be presented as primary without qualification.
  \item \textbf{No test set contamination.} All conversations were collected through \textbf{March 2026}. Users are encouraged to verify that EMs were not exposed to similar content during training.
\end{itemize}

\section{Conclusion}
%% ════════════════════════════════════════════════════════════════════════════
A single number alone cannot describe how well an LLM handles an emotional conversation. \methodname{} resolves EI into distinct operational capabilities, affective tracking, behavioral inference, preference prediction, and response generation, and shows that across 11 contemporary models, rankings on these are largely independent. Together with the released dataset, evaluation runner, and metric definitions, \methodname{} is intended as a diagnostic benchmark for tracking model strengths and failure modes in EI.

%% ════════════════════════════════════════════════════════════════════════════
%%  ACKNOWLEDGMENTS
%% ════════════════════════════════════════════════════════════════════════════
\begin{acknowledgments}
We thank the Pareto annotation team members who participated in this data collection for their openness during conversations and their careful labeling work.

LLM Usage: The authors used LLMs for formatting and editing assistance during manuscript preparation. Statistical analysis scripts were authored with LLM assistance and verified by the human authors. Scientific content, analyses, and conclusions are the responsibility of the human authors.
\end{acknowledgments}

%% ════════════════════════════════════════════════════════════════════════════
%%  CONTRIBUTIONS
%% ════════════════════════════════════════════════════════════════════════════
\begin{contributions}

This work was a collaboration between Pareto AI’s Research team, led by Mark Whiting, and Thoughtful Labs, led by Karina Nguyen. Our teams came together grounded in a mutual interest in human-centered AI and measuring overlooked but important LLM capabilities, combining our expertise in data collection, benchmark design, and model assessment methodology. We collectively strategized about which metrics to measure and how best to capture them, spanning concept development, dataset design, participant and annotation workflows, annotation taxonomies, scoring methodology, and evaluation goals.
\end{contributions}

%% ════════════════════════════════════════════════════════════════════════════
%%  REFERENCES
%% ════════════════════════════════════════════════════════════════════════════
\bibliographystyle{plainnat}
\bibliography{references}

\clearpage
\startappendix
\section*{Appendix}
\input{appendix}
\clearpage                                         
\input{supplemental}

\end{document}

%% file: appendix.tex
% \section{Appendix}
%% Section numbering follows appendix convention (A, B, C, ...)
\appendix

\section{Acronyms}

\begin{table}[H]
\centering
\caption{Key acronyms used throughout the paper.}
\begin{tabular}{ll}
\toprule
Acronym & Definition \\
\midrule
EI & Emotional Intelligence \\
HP & Human Participant \\
OM & Original Model \\
EM & Evaluated Model \\
PANAS & Positive and Negative Affect Schedule \\
VA & Valence-Arousal (as in Emotion VA metric) \\
VAD & Valence-Arousal-Dominance \\
PII & Personally Identifiable Information \\
ASD & Autism Spectrum Disorder \\
ADHD & Attention Deficit Hyperactivity Disorder \\
MAE & Mean Absolute Error \\
\bottomrule
\end{tabular}
\end{table}

\section{Survey Instruments}

\begin{longtable}{p{0.4\linewidth} p{0.55\linewidth}}
\caption{Applicant Intake Survey Questions.}
\label{tab:intake} \\

\toprule
\textbf{Question} & \textbf{Answer Options} \\
\midrule
\endfirsthead

\toprule
\textbf{Question} & \textbf{Answer Options} \\
\midrule
\endhead

Country of Residence & 193-country list \\

Highest Level of Education &
High school; GED; Associate's; Bachelor's; Master's; Doctorate; Professional certification; Other \\

Education / Work Background &
Health \& Medicine; Technology \& Computing; Science \& Research; Business \& Finance; Legal \& Justice; Engineering; Education; Social Sciences; Arts \& Media; Humanities; HR; Sales; Operations; Construction; Other \\

Gender & Male; Female; Other \\

Age Range & 18--24; 25--34; 35--44; 45--54; 55--64; 65+ \\

English Proficiency & Native; Fluent; Proficient; Conversational; Basic \\

Mental Health Diagnoses &
ADHD; Anxiety; Depression; ASD; OCD; PTSD; etc.; None; Decline; Other \\

Openness to Sensitive Topics & Yes; No; Unsure; Other \\

\bottomrule
\end{longtable}

\begin{longtable}{p{0.35\linewidth} p{0.6\linewidth}}
\caption{Participant Profile Questionnaire.}
\label{tab:profile} \\

\toprule
\textbf{Measure / Question} & \textbf{Description / Response Format} \\
\midrule
\endfirsthead

\toprule
\textbf{Measure / Question} & \textbf{Description / Response Format} \\
\midrule
\endhead

\multicolumn{2}{l}{\textit{Attitudes Toward LLMs}} \\
\midrule

Perceived Trustworthiness of LLMs & Likert scale (1 = Completely Untrustworthy, 5 = Completely Trustworthy) \\
Perceived Helpfulness of LLMs & Likert scale (1 = Completely Unhelpful, 5 = Completely Helpful) \\
Prior Use for Emotional Conversations & Yes (regularly); Yes (occasionally); No; Unsure \\
Motivations for Using LLMs & Venting; Advice seeking; Reassurance; Gaining perspective; Practicing conversations; Curiosity; Not interested; Other \\

\midrule
\multicolumn{2}{l}{\textit{Standardized Psychological Measures}} \\
\midrule

TIPI & Big Five personality traits \\
WHO-5 & Psychological well-being \\
AQ-10 & Autistic traits screening \\
ASRS-6 & ADHD screener \\
PROMIS Depression-4 & Depression assessment \\
PROMIS Anxiety-4 & Anxiety assessment \\
PROMIS Sleep Disturbance-4 & Sleep quality \\

\midrule
\multicolumn{2}{l}{\textit{Additional Screening Items}} \\
\midrule

Recent Major Life Changes & Yes/No + optional details \\
Recent Traumatic Events & Yes/No + optional details \\

\midrule
\multicolumn{2}{l}{\textit{Topic-Specific Emotional Baseline}} \\
\midrule

Emotional Orientation Toward Topics &
Sad/Disappointed; Worried/Angry; Neutral; Content; Excited; Mixed; Unsure \\

\bottomrule
\end{longtable}

\begin{longtable}{p{0.25\linewidth} p{0.7\linewidth}}
\caption{Conversation Topic Taxonomy.}
\label{tab:topics} \\

\toprule
\textbf{Category} & \textbf{Subtopics} \\
\midrule
\endfirsthead

\toprule
\textbf{Category} & \textbf{Subtopics} \\
\midrule
\endhead

Politics & War \& Foreign Policy; Immigration; Civil Rights; Climate; Criminal Justice \\
Money & Finance; Debt; Cost of Living; Income; Spending \\
Work / School & Career; Work-life Balance; Responsibilities; Colleagues; Activities \\
Religion & Organized Religion; Secularism; Conversion; Spirituality; Interfaith \\
Family & Parents; Children; Siblings; Traditions; Pets \\
Friends & Social Life; Online; Group Dynamics; Making Friends \\
Romantic Relationships & Dating; Intimacy; Marriage; Breakups; Nonmonogamy \\
Physical Health & Illness; Nutrition; Sleep; Body Image; Exercise \\
Entertainment & Books; Movies; TV; Music; Streaming \\
Hobbies & Gaming; Sports; Arts; Cooking; Travel \\

\bottomrule
\end{longtable}

%% Tables A.5 and A.6 placed side-by-side; converted from longtable to tabular.
\begin{table}[H]
\centering
\begin{minipage}[t]{0.46\textwidth}
  \centering
  \caption{PANAS Emotion Taxonomy.}
  \label{tab:panas}
  \begin{tabular}{ll}
  \toprule
  \textbf{Positive Affect} & \textbf{Negative Affect} \\
  \midrule
  Interested   & Distressed \\
  Excited      & Upset \\
  Strong       & Guilty \\
  Enthusiastic & Scared \\
  Proud        & Hostile \\
  Alert        & Irritable \\
  Inspired     & Ashamed \\
  Determined   & Nervous \\
  Attentive    & Jittery \\
  Active       & Afraid \\
  \bottomrule
  \end{tabular}
\end{minipage}%
\hfill
\begin{minipage}[t]{0.46\textwidth}
  \centering
  \caption{PANAS Rating Scale.}
  \label{tab:likert}
  \begin{tabular}{cl}
  \toprule
  \textbf{Score} & \textbf{Description} \\
  \midrule
  1 & Not at all \\
  2 & Slightly \\
  3 & Somewhat \\
  4 & Moderately \\
  5 & Quite a bit \\
  6 & Very much \\
  7 & Extremely \\
  \bottomrule
  \end{tabular}
\end{minipage}
\end{table}

  \section{Participant Consent Statement}                                                               
  \label{app:consent}        
  Participants were presented with the following consent statement prior to their first conversation:   

  \begin{quote}
  'By participating in this study, you acknowledge that you are voluntarily taking part in research conducted by Pareto. You understand that data will be collected during your participation, which may include responses you provide, usage data, or other relevant information related to the study. Your data will be handled in accordance with applicable data protection laws and will be stored securely. Any results shared publicly will be reported in an anonymized form so that you cannot be personally identified. Participation is voluntary, and you may withdraw from the study at any time.'
  \end{quote}

  \paragraph{Compensation.}
  Participants were compensated at a base rate of \$20/hour USD, plus a flat \$50 bonus for every 10 accepted submissions, with each conversational task expected to take 30--90 minutes.

\section{Evaluation Framework}

\begin{table}[H]
\centering
\caption{Evaluation metrics. All on 0--1 scale except Kendall~$\tau$ and PANAS B-Adj ($-1$ to $+1$) and Composite (0--100). 
\textsuperscript{\normalfont\textdagger}Produced by a judge LLM; reported alongside primary metrics rather than as the primary benchmark signal.}
\label{tab:metrics}
\small
\setlength{\tabcolsep}{4pt}
\begin{tabular}{p{3.0cm} p{10.5cm}}
\toprule
\textbf{Metric} & \textbf{Description} \\
\midrule
Emotion F1 & Turn-level set F1 between EM-predicted PANAS emotion tags and HP-reported mood-shift tags. \\
Emotion VA & Continuous distance in the 2D valence-arousal space; sensitive to the direction and magnitude of errors rather than treating all mispredictions as equivalent. \\
Binary OM Acc & EM agreement with HP's observed-behavior labels for each applicable binary question (model-facing phrasing). \\
Binary HP Acc & EM agreement with HP's preferred-behavior labels for each applicable binary question (participant-facing phrasing). \\
Pairwise Acc & Proportion of correct preference predictions across the three response variants (original, model-generated, human-edited). \\
Kendall $\tau$ & Rank-order agreement across the three response variants; complementary to Pairwise Accuracy and more reflective of overall preference ordering. \\
Draft Judge\textsuperscript{\textdagger} & Mean of four quality dimensions (overall, emotional appropriateness, helpfulness, tone match) rated by a judge LLM, averaged across turns. \\
Four Branch MAE & Mean absolute error between EM-predicted and HP-reported Four Branch ratings, per branch and aggregated. \\
PANAS B-Adj & Error in predicting HP post-conversation PANAS, adjusted for model-level prediction bias to isolate conversation-level tracking from systematic scale-use tendencies. \\
Composite & 0--100 weighted aggregate across three pillars: emotion tracking (24\%), evaluation quality (49\%), and holistic comprehension (27\%); see Section~\ref{sec:scoring}. \\
\bottomrule
\end{tabular}
\end{table}

\begin{table}[H]
\centering
\caption{Original Models (OMs) used during data collection. HPs were randomly connected to one of these eight models at conversation start; model identity was not disclosed.}
\label{tab:om-models}
\small
\begin{tabular}{ll}
\toprule
\textbf{Display Name} & \textbf{API Identifier} \\
\midrule
GPT-4 Turbo (Preview)~\cite{openai2023gpt4} & \texttt{gpt-4-turbo-preview} \\
GPT-4o~\cite{openai2024gpt4o}                    & \texttt{gpt-4o} \\
GPT-4o mini~\cite{openai2024gpt4omini}           & \texttt{gpt-4o-mini} \\
Claude 3.5 Sonnet~\cite{anthropic2024claude35}   & \texttt{claude-3.5-sonnet} \\
Claude 3 Haiku~\cite{anthropic2024claude3}       & \texttt{claude-3-haiku} \\
Gemini 2.5 Flash~\cite{google2025gemini25}       & \texttt{gemini-2.5-flash} \\
Gemini 2.0 Flash~\cite{google2024gemini20}       & \texttt{gemini-2.0-flash-001} \\
Inflection 3 (Pi)~\cite{inflection2024pi}        & \texttt{inflection-3-pi} \\
\bottomrule
\end{tabular}
\end{table}

\begin{table}[H]
\centering
\caption{Summary of included and excluded data.}
\label{tab:data-incl}
\begin{tabular}{p{5cm} p{8cm}}
\toprule
\textbf{Category} & \textbf{Description} \\
\midrule
\multicolumn{2}{l}{\textit{Included Data}} \\
\midrule
Conversation Metadata & Topic, subtopic, model version \\
Conversation Content & Full transcript of interaction \\
Participant Profile & Anonymized demographics and psychometric summaries \\
\midrule
\multicolumn{2}{l}{\textit{Excluded Data}} \\
\midrule
Identifiers & Names, usernames, aliases \\
Contact Information & Email and contact details \\
Precise Location & Any location beyond country-level \\
Precise Demographics & Exact age or birthdate \\
\bottomrule
\end{tabular}
\end{table}

\begin{longtable}{p{0.3\linewidth} p{0.1\linewidth} p{0.6\linewidth}}
\caption{Conversation-Wide Questions.}
\label{tab:convwide} \\

\toprule
\textbf{Question} & \textbf{Response Type} & \textbf{Response Options}\\
\midrule
\endfirsthead

\toprule
\textbf{Question} & \textbf{Response Type} & \textbf{Response Options}\\
\midrule
\endhead

What were you mostly looking for from the model in this conversation & Select up to 2 &  To just listen or let me vent; To help me understand or sort out my feelings; To help me think through options or make a decision; To help me calm down or feel steadier; To help with something urgent, risky, or high-stakes; Other (+Free text)\\
How clear were your emotions during the conversation? & Select one & Clear and explicitly stated; Implied or indirect; Mixed or conflicted; Unclear / Evolving as I talked \\
How well did the model's responses fit what you needed, as the conversation unfolded? & Select one & Mostly off-target or intrusive; Mixed, some good moments, some misses; Mostly well-matched; Very well-matched and adaptive \\
What felt off in those moments? (Asked only if answer to previous question is ``Mostly off-target or intrusive'' or ``Mixed, some good moments, some misses'') & Select all that apply  & It misunderstood how I was feeling; It made assumptions that didn't fit; It moved too fast or too slow; It gave advice or direction I didn't want; Its tone felt wrong for the situation; It didn't adjust after I reacted or pushed back; Other (+Free text) \\

\bottomrule
\end{longtable}

\begin{longtable}{p{0.3\linewidth} p{0.65\linewidth}}
\caption{Four-Branch Emotional Intelligence Model (condensed).}
\label{tab:fourbranch-condensed} \\

\toprule
\textbf{Branch} & \textbf{Definition} \\
\midrule
\endfirsthead

\toprule
\textbf{Branch} & \textbf{Definition} \\
\midrule
\endhead

Perceiving Emotion & Identify user's emotional state \\
Facilitating Thought & Use emotion in reasoning \\
Understanding Emotion & Model emotional dynamics \\
Managing Emotion & Respond appropriately \\

\bottomrule
\end{longtable}

Binary evaluation questions are organized below under the four canonical Mayer-Salovey-Caruso branches (Table~\ref{tab:fourbranch-condensed}) for consistency with the framework cited throughout the paper. During annotator training, questions were grouped under five working categories: Perceiving Emotion, Understanding Emotion, Using Emotion, Regulating Emotion, and Cognitive Framing. The first four map directly onto the canonical branches (Using Emotion $\to$ Facilitating Thought; Regulating Emotion $\to$ Managing Emotion). The five Cognitive Framing questions are distributed by content. Questions about acknowledging complexity, multiple interpretations, and other-perspective taking are placed under Understanding Emotion, while questions about prompting reflection are placed under Facilitating Thought.

\begin{longtable}{p{0.25\linewidth} p{0.7\linewidth}}
\caption{Binary Evaluation Questions, as asked to HP, organized by Mayer-Salovey-Caruso four-branch category.}
\label{tab:binary} \\

\toprule
\textbf{Branch} & \textbf{Question} \\
\midrule
\endfirsthead

\toprule
\textbf{Branch} & \textbf{Question} \\
\midrule
\endhead

Perceiving Emotion & Did the response acknowledge emotional content in your message? \\
Perceiving Emotion & If the response makes a claim about your emotional state (explicitly or implicitly), is that claim consistent with what you expressed? \\
Perceiving Emotion & Did the response reflect a specific detail from your message showing understanding? \\
Perceiving Emotion & If the response referenced your emotion, did it reflect your wording or phrasing rather than substituting a different emotional label? \\
Perceiving Emotion & If multiple emotions were present in the message, did the response prioritize or focus on the one that seemed most central? \\

Understanding Emotion & Did the response acknowledge more than one emotional state you expressed or implied (mixed feelings or intensity)? \\
Understanding Emotion & Did the response acknowledge a change in emotional state across the conversation? \\
Understanding Emotion & Did the response make an inference/assumption beyond what you said? \\
Understanding Emotion & If the model inferred your emotional state, did it express appropriate uncertainty (``It sounds like\ldots'', ``You might be feeling\ldots'')? \\
Understanding Emotion & Did the response sound overly formal or clinical relative to your tone? \\
Understanding Emotion & Did the response repeat emotional language multiple times in a way that felt excessive? \\
Understanding Emotion & Did the response suggest considering another person's perspective in the situation? \\
Understanding Emotion & Did the response offer multiple possible interpretations of the situation rather than assuming one explanation? \\
Understanding Emotion & Did the response include a statement acknowledging that the situation may be complex or multifaceted? \\

Facilitating Thought & Was the emotional tone appropriate for the moment? \\
Facilitating Thought & Did the model match how serious the situation felt? \\
Facilitating Thought & Was the emotional intensity of the response appropriate? \\
Facilitating Thought & Did the model try to guide or direct you? \\
Facilitating Thought & Did the model provide explanations or reasoning? \\
Facilitating Thought & Did the model ask a follow-up question? \\
Facilitating Thought & Did the model make a moral judgement? \\
Facilitating Thought & Did the response include advice that was not clearly connected to your specific situation? \\
Facilitating Thought & Did the response shift focus away from your concern toward a generic topic or explanation? \\
Facilitating Thought & Did the response provide advice before asking any clarifying questions? \\
Facilitating Thought & Did the response introduce a new topic that had not been previously discussed? \\
Facilitating Thought & If the response suggested an action, did it acknowledge potential constraints or difficulties in taking that action? \\
Facilitating Thought & Did the response encourage you to reflect on your own perspective or feelings? \\
Facilitating Thought & Did the response ask a question that required reflection rather than a factual answer? \\

Managing Emotion & Did the model adjust after the user reacted or clarified? \\
Managing Emotion & Did the response identify potential risk indicators? \\
Managing Emotion & Did the model provide appropriate safety or emergency framing when needed? \\
Managing Emotion & Did the response include a statement that validates the user's emotional experience? \\
Managing Emotion & Did the response use language that could increase emotional intensity (catastrophizing, alarmist phrasing)? \\
Managing Emotion & Did the response correct or revise an earlier assumption made by the model? \\
Managing Emotion & Did the response repeat advice that had already been suggested earlier in the conversation? \\

\bottomrule
\end{longtable}

\begin{longtable}{p{0.1\linewidth} p{0.3\linewidth} p{0.5\linewidth}}
\caption{Pairwise Comparison Questions.}
\label{tab:pairwise} \\

\toprule
\textbf{ID} & \textbf{Category} & \textbf{Question} \\
\midrule
\endfirsthead

\toprule
\textbf{ID} & \textbf{Category} & \textbf{Question} \\
\midrule
\endhead

PW1 & Emotional Understanding & Which response better understands how you felt? \\
PW2 & Emotional Understanding & Which response better handles mixed or unclear feelings? \\
PW3 & Emotional Understanding & Which response avoids incorrect assumptions better? \\

PW4 & Validation \& Attunement & Which feels more supportive? \\
PW5 & Validation \& Attunement & Which feels less dismissive? \\
PW6 & Validation \& Attunement & Which response better lets you express yourself? \\

PW7 & Guidance Quality & Which response gives more helpful direction? \\
PW8 & Guidance Quality & Which response respects your ability to decide? \\
PW9 & Guidance Quality & Which response avoids pushing too hard? \\

PW10 & Regulation \& Safety & Which helps you feel steadier? \\
PW11 & Regulation \& Safety  & Which takes the situation seriously? \\
PW12 & Regulation \& Safety  & Which response responds better to potential risk? \\

PW13 & Interaction Fit \& Adaptation & Which response better matches what you needed in that moment? \\
PW14 & Interaction Fit \& Adaptation & Which response adjusts better to your reactions? \\
PW15 & Interaction Fit \& Adaptation & Which response has better pacing for the conversation? \\

\bottomrule
\end{longtable}

\begin{longtable}{p{0.3\linewidth} p{0.65\linewidth}}
\caption{Four-Branch Emotional Intelligence Model (participant-facing definitions).}
\label{tab:fourdefs} \\

\toprule
\textbf{Branch} & \textbf{Definition} \\
\midrule
\endfirsthead

\toprule
\textbf{Branch} & \textbf{Definition} \\
\midrule
\endhead

Perceiving Emotion & The model's ability to identify the current emotional state of human. \\
Facilitating Thought & The model's ability to consider the human's emotional state and the overall emotional context of the conversation when problem solving, reasoning, and crafting responses.  \\
Understanding Emotion & The model's ability to understand how the human's emotions might combine, change and manifest over time.
In other words, beyond the human's current emotional state, how does model predict this might change based on different directions the conversation and situation could develop. \\
Managing Emotion & The model's ability to invoke and convey emotions clearly and appropriately. Does the model respond to the human in a manner that seems thoughtful? \\

\bottomrule
\end{longtable}

%% ════════════════════════════════════════════════════════════════════════════
\section{Interface Screenshots}
%% ════════════════════════════════════════════════════════════════════════════
%%  

\begin{figure}[H]
  \centering
   \includegraphics[width=0.8\textwidth]{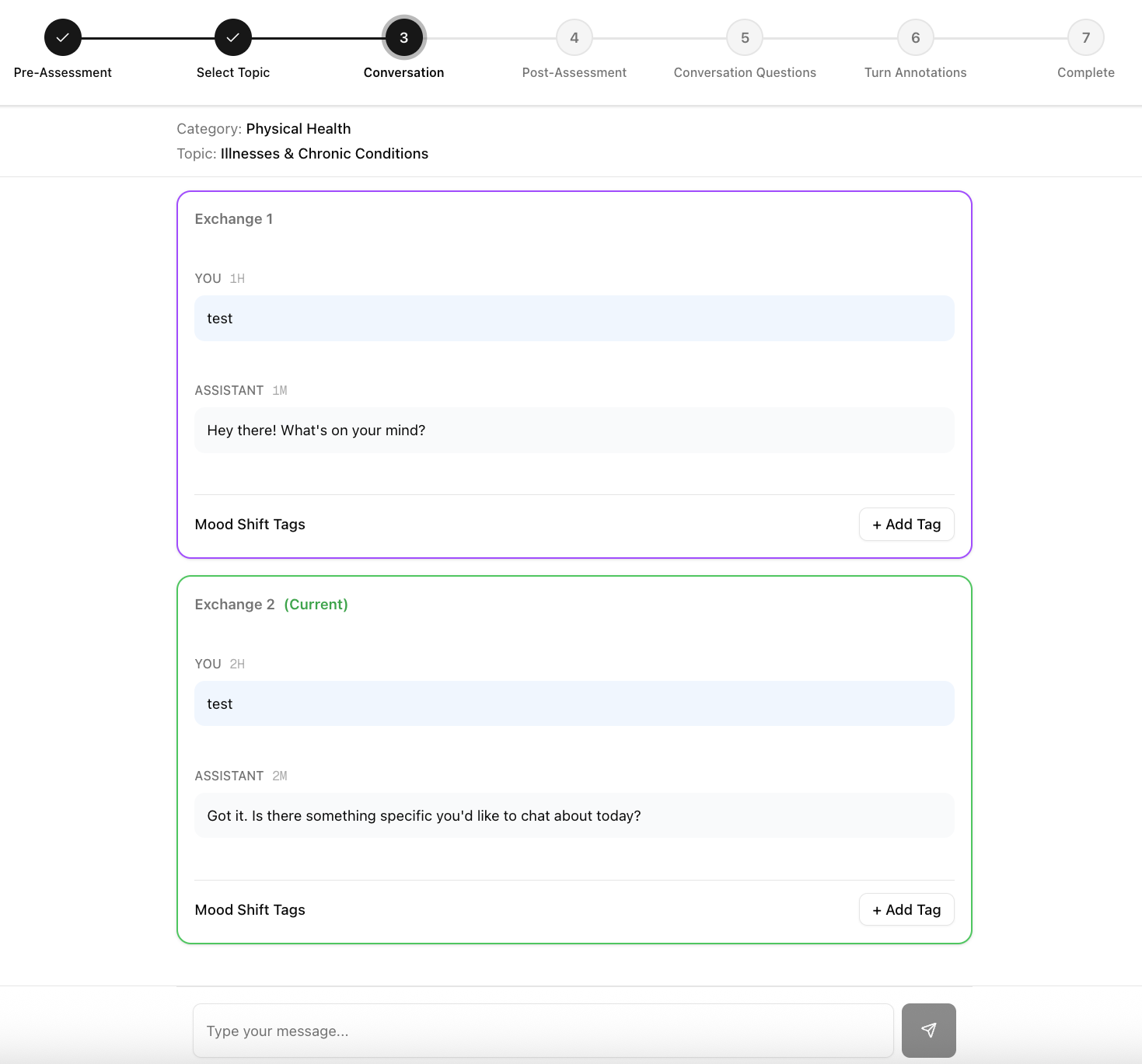}
  \caption{Conversation interface presented to participants. Each session consisted of a single multi-turn conversation between the participant and an LLM on the assigned topic.}
  \label{fig:ui-conversation}
\end{figure}

\begin{figure}[H]
  \centering
   \includegraphics[width=0.85\textwidth]{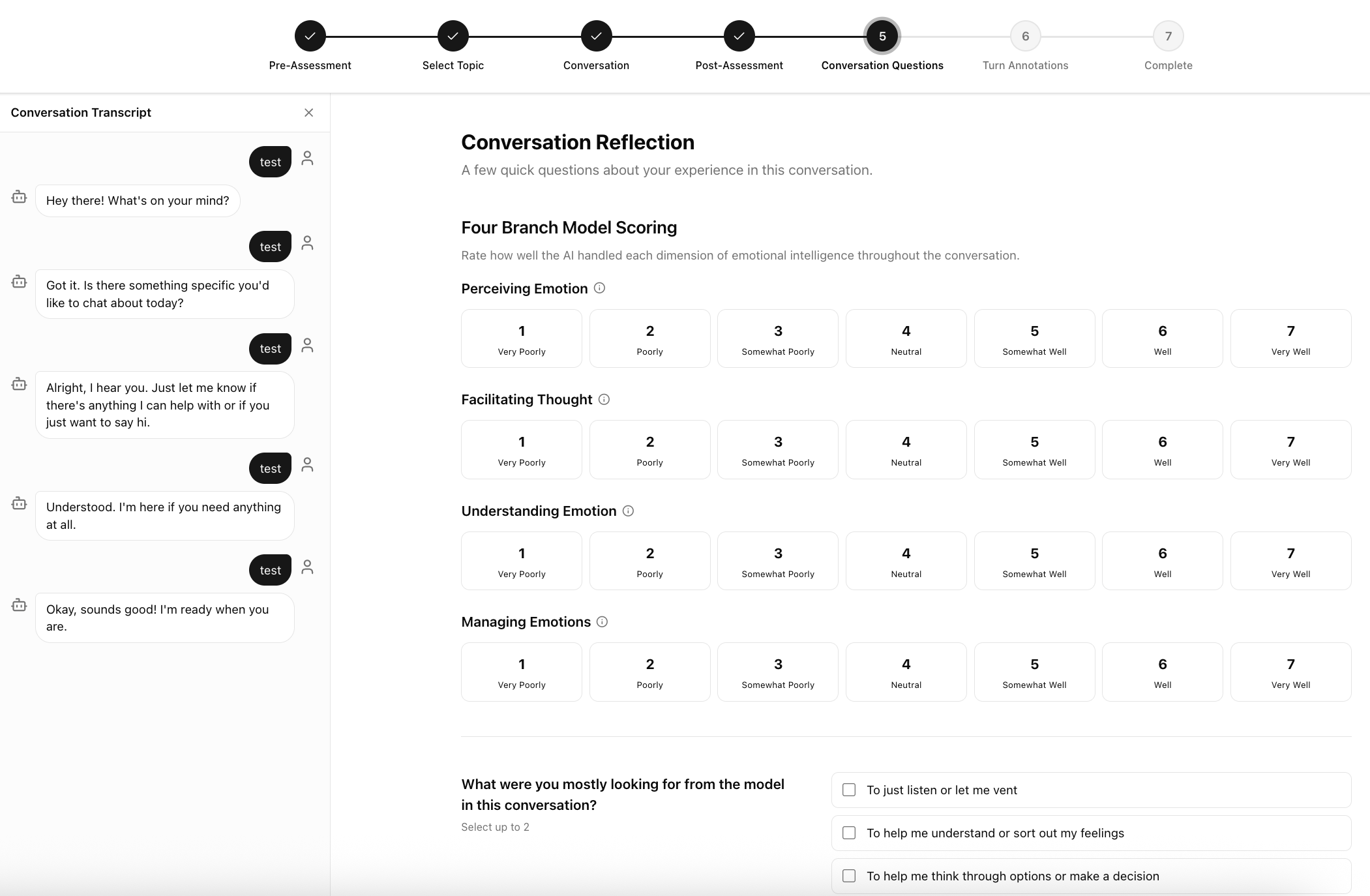}
  \caption{Conversation-wide Assessment page of post-conversation annotation interface. Participants rated general performance using the Four Branch Model and answered questions about their overall conversational goals and satisfaction. They were able to reference the conversation transcript while rating.}
  \label{fig:ui-annotation1}
\end{figure}

\begin{figure}[H]
  \centering
   \includegraphics[width=0.7\textwidth]{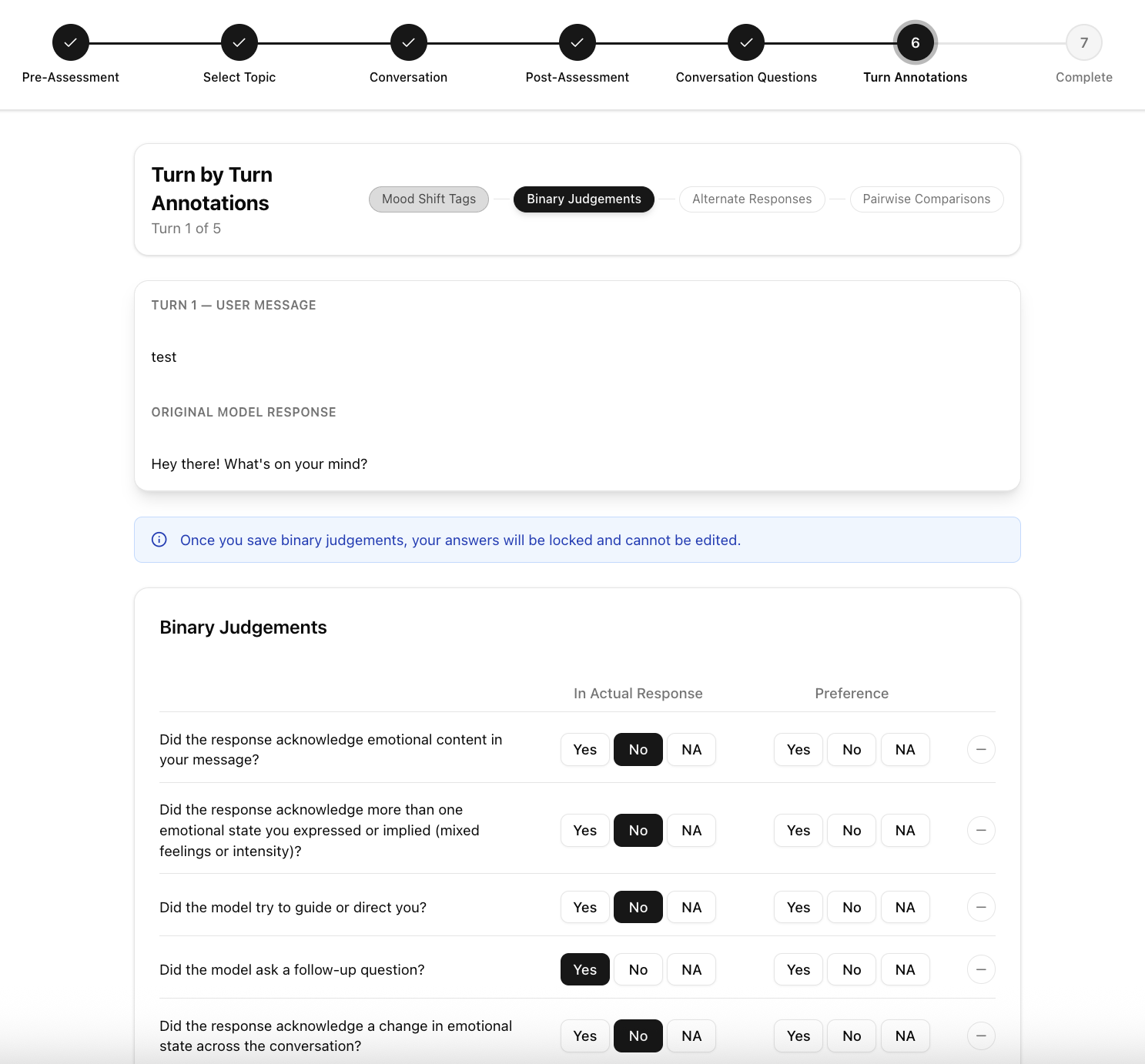}
  \caption{Binary Judgment page of the post-conversation annotation interface. For selected applicable binary questions, participants recorded both the observed model behavior ("In Actual Response"), as well as their preferred behavior ("Preference").}
  \label{fig:ui-annotation2}
\end{figure}

\begin{figure}[H]
  \centering
   \includegraphics[width=0.7\textwidth]{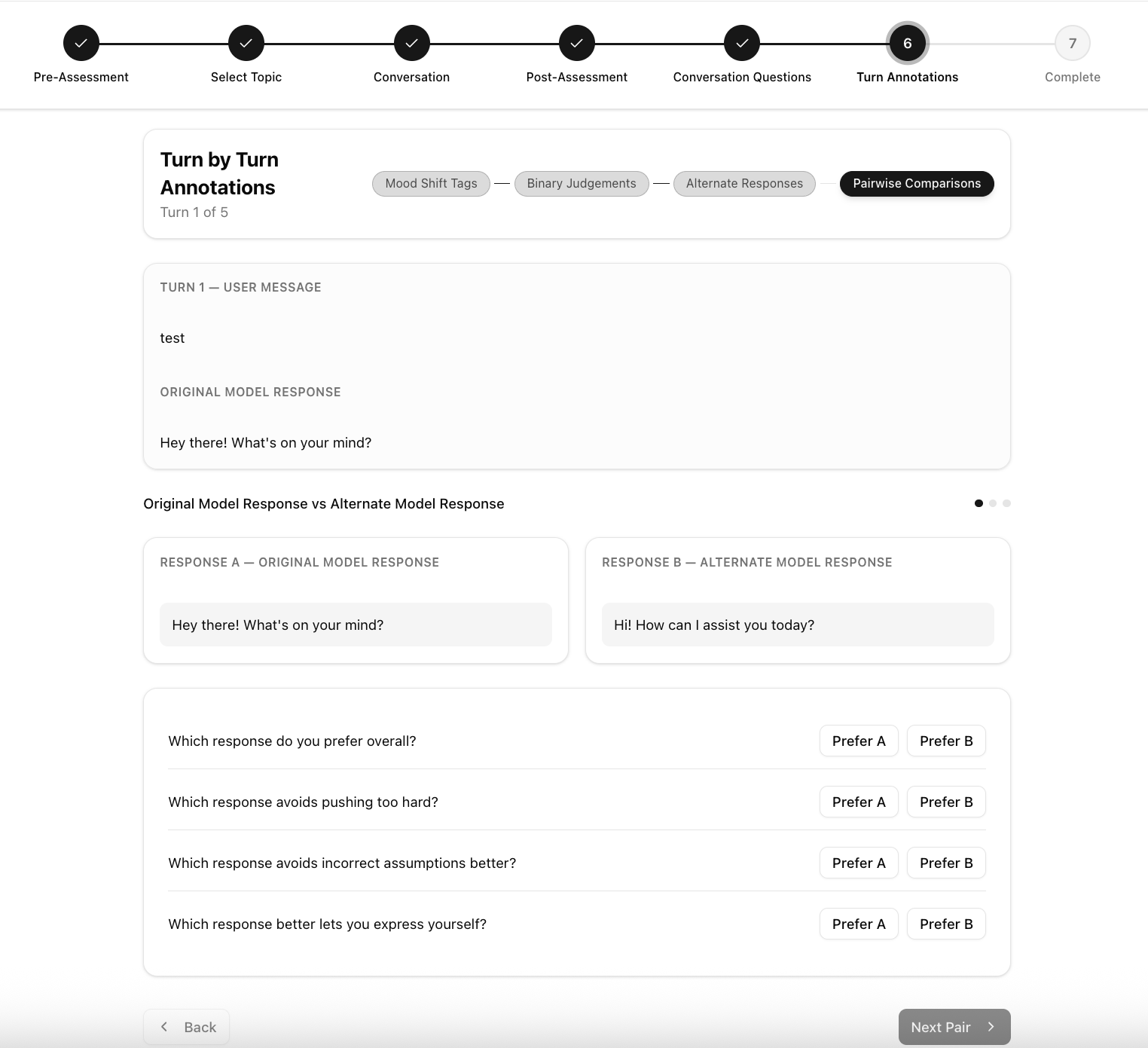}
  \caption{Pairwise Ranking page of the post-conversation annotation interface. Considering three different response options, participants selected their preferred response for four pairwise comparison questions, which varied from turn to turn.}
  \label{fig:ui-annotation3}
\end{figure}

%% file: supplemental.tex
%% ════════════════════════════════════════════════════════════════════════════
%%  supplemental.tex — Supplementary Material for AttuneBench
%%
%%  Included from main.tex via \input{supplemental} after \input{appendix}.
%%  Continues the appendix letter numbering already started in appendix.tex.
%%  Do NOT redeclare \appendix or call \maketitle here.
%% ════════════════════════════════════════════════════════════════════════════

% All supplemental figures live in paper/figures/ with mode prefixes
% (def_, vrb_, omn_, cmp_) to keep the directory flat.
\graphicspath{{figures/}}

\section{Supplementary Analyses}
\label{sec:supp-analyses}

This section reports additional analyses that complement the main results in Section~\ref{sec:analysis} but were placed here to keep the main body focused. All analyses use the 11-model Default Mode evaluation ($n = 200$ conversations per model) unless otherwise noted.

\subsection{PANAS Mood Shift Subsets}
\label{sec:supp-panas-shift}

We grouped the 200 Default Mode conversations by pre-to-post PANAS shift, using a $\pm 5$-point threshold on either the positive-affect (Pos) or negative-affect (Neg) subscale to classify each conversation into one of four groups: \textbf{Stable} (no meaningful shift on either axis; $n = 78$), \textbf{Positive} (Pos increased by $\geq 5$, Neg did not increase by $\geq 5$; $n = 71$), \textbf{Negative} (Neg increased by $\geq 5$, Pos did not increase by $\geq 5$; $n = 40$), and \textbf{Mixed} (both axes shifted by $\geq 5$; $n = 11$). The Mixed group is small but psychologically interesting: these are conversations in which the participant ended up both more positive and more negative than they started.

\paragraph{Composite ordering: Mixed > Positive > Negative > Stable.}
Pooled across 11 models, conversation difficulty as captured by the Composite Score is strongly modulated by emotional dynamics (Kruskal-Wallis $H(3) = 234.6$, $p < 10^{-49}$, $\eta^2 = 0.105$). Mean Composite is highest for Mixed (55.83) and Positive (54.49), intermediate for Negative (52.17), and lowest for Stable (49.97); all six pairwise contrasts except Positive vs Mixed are significant after Holm-Bonferroni correction. The Stable-low pattern is consistent with the interpretation that conversations in which the participant's affective state did not move are the ones where models had the least to track.

\paragraph{Emotion VA tracks the expected ordering.}
Emotion VA differentiates groups even more cleanly ($H(3) = 118.7$, $p < 10^{-25}$, $\eta^2 = 0.053$). Mean VA: Mixed 0.345, Positive 0.286, Stable 0.256, Negative 0.195. Conversations with strong emotional movement on either or both axes (Mixed, Positive) are easier for models to track in valence-arousal space than Stable conversations or those drifting Negative.

\paragraph{Pairwise Accuracy is inverted.}
Unlike the Composite or Emotion VA, Pairwise Accuracy is \emph{highest} on Negative conversations (0.558) and lower on Positive (0.532), Mixed (0.530), and Stable (0.529); the Stable--Negative gap is significant ($p < 0.01$). This inversion is consistent with a hypothesis that participants experiencing a negative emotional shift produced sharper preference signals (clearer preferred-vs-not-preferred response distinctions), making preference prediction easier even though the conversation was harder on other dimensions.

\paragraph{Binary OM Accuracy is approximately flat across the three main groups.}
The 3-group (Stable / Positive / Negative) omnibus on Binary OM Accuracy is non-significant ($H(2) = 4.41$, $p = 0.11$); only Mixed (0.880) stands out. This is consistent with the main-paper finding that observed-behavior assessment is the most stable component of the benchmark across difficulty conditions.

\paragraph{Draft Judge mirrors the Composite ordering.}
Draft Judge shows the same Mixed > Positive > Negative > Stable ordering as the Composite ($H(3) = 39.5$, $p < 10^{-7}$, $\eta^2 = 0.017$), with Stable conversations producing the lowest-quality drafts (0.791) and Mixed the highest (0.821). Per the main paper's caveats, Draft Judge reflects judge-model preferences (Mistral Large) and should be interpreted alongside the other quality signals, not as a primary indicator.

A full per-model Kruskal-Wallis breakdown and pairwise group contrast tables are available in the analysis output (\texttt{panas\_shift\_output.txt} in the code repository).

\subsection{Conversation Length Effects}
\label{sec:supp-length}

Conversation length in the benchmark spans 5 to 11 turns (median 6); 162 of 200 conversations are 6 turns or fewer (Short) and 38 are longer (Long). This section reports a sanity check: do model rankings or absolute scores depend on conversation length?

Pooled across 11 models, no model shows a Composite difference between Short and Long conversations that survives Holm-Bonferroni correction. Per-model $\Delta$(Short $-$ Long) values span $-1.18$ (Opus 4.6) to $+1.30$ (Mistral Lg) Composite points, all with $p_\text{adj} = 1.0$ and small effect sizes ($|r| \leq 0.09$). The same null pattern holds for Pairwise Accuracy, Binary OM/HP Accuracy, and Emotion VA at the per-model level.

We interpret this as evidence that conversation length is approximately orthogonal to model quality on this benchmark: the model ordering observed in the main results is preserved across short and long conversations, and absolute scores are not systematically inflated or deflated by length. This is consistent with the turn-drift analysis in Section~\ref{sec:analysis} of the main paper, which finds significant within-conversation degradation (Binary OM Acc) but does not find that longer conversations produce systematically different aggregate scores.

A caveat: the Long-conversation subset is small ($n = 38$, distributed across all 11 models so per-model $n$ is the full 38), which limits statistical power for detecting modest length effects. Future versions of the benchmark with broader length distributions could revisit this analysis.

\subsection{Per-Model Breakdown by Conversation Topic}
\label{sec:supp-topic-permodel}

The main paper (Section~\ref{sec:analysis}) reports topic-level pooled results showing Romantic Relationships as the most-difficult topic and modest topic effects overall ($\eta^2 \leq 0.022$). Table~\ref{tab:supp-topic-permodel} gives the full 11-model $\times$ 10-topic Composite matrix, allowing inspection of whether the topic-difficulty ordering is preserved at the per-model level.

\begin{table}[H]
\centering
\caption{Composite Score by model and conversation topic (Default Mode, $n = 200$ across all topics, varying per-topic). Topic abbreviations: Pol = Politics, Mny = Money, Wrk = Work/School, Fam = Family, Hob = Hobbies, Ent = Entertainment Media, Fri = Friends, Rel = Religion, Phy = Physical Health, Rom = Romantic Relationships.}
\label{tab:supp-topic-permodel}
\small
\setlength{\tabcolsep}{4pt}
\begin{tabular}{l rrrrrrrrrr}
\toprule
\textbf{Model} & \textbf{Pol} & \textbf{Mny} & \textbf{Wrk} & \textbf{Fam} & \textbf{Hob} & \textbf{Ent} & \textbf{Fri} & \textbf{Rel} & \textbf{Phy} & \textbf{Rom} \\
\midrule
Opus 4.7      & \textbf{56.2} & 54.1 & 53.5 & 54.9 & 54.2 & 53.1 & 53.0 & 54.0 & 53.3 & 50.5 \\
Opus 4.6      & 55.6 & 54.9 & 53.9 & \textbf{56.5} & 56.3 & 53.9 & 53.7 & 53.6 & 53.6 & 51.5 \\
Haiku 4.5     & 52.3 & 54.0 & \textbf{54.8} & 52.6 & 52.5 & 53.7 & 54.2 & 52.4 & 51.5 & 50.3 \\
Sonnet 4.6    & 51.7 & 50.7 & 50.7 & \textbf{51.9} & 49.9 & 50.0 & 49.7 & 49.8 & 50.2 & 48.5 \\
Gemini 3.1    & 51.9 & 53.4 & 53.7 & 53.5 & 53.2 & \textbf{54.4} & 54.0 & 51.6 & 53.3 & 49.7 \\
Mistral Lg    & 52.6 & 52.9 & \textbf{53.5} & 50.5 & 52.3 & 52.3 & 52.8 & 51.5 & 50.6 & 49.1 \\
GPT-5.5       & 54.8 & 54.0 & 54.1 & 54.5 & 54.4 & \textbf{55.1} & 53.2 & 52.9 & 52.3 & 52.1 \\
GPT-5.4       & \textbf{53.1} & 50.8 & 51.1 & 52.3 & 50.6 & 49.3 & 49.7 & 48.2 & 49.4 & 48.6 \\
Qwen 2.5 72B  & 52.1 & 53.4 & \textbf{53.8} & 51.9 & 52.5 & 53.1 & 53.6 & 51.2 & 51.9 & 48.9 \\
Grok 4        & 51.2 & \textbf{51.3} & 49.6 & 49.1 & 51.2 & 50.5 & 49.9 & 51.2 & 49.3 & 48.3 \\
MiMo-v2-Pro   & 55.0 & \textbf{55.8} & 54.7 & 54.8 & 53.7 & 53.1 & 54.5 & 52.9 & 52.8 & 49.4 \\
\midrule
\textbf{Pooled} & 53.3 & 53.2 & 53.0 & 53.0 & 52.8 & 52.6 & 52.6 & 51.8 & 51.6 & 49.7 \\
\bottomrule
\end{tabular}
\end{table}

\paragraph{Romantic Relationships is the worst topic for 10 of 11 models.}
Romantic Relationships is each model's worst topic except for GPT-5.4, where Religion (48.2) edges out Romantic (48.6). Across all other 10 models, Romantic produces the lowest Composite by a margin of 0.7 to 4.6 points. The pattern holds at the pooled level (Romantic 49.7 vs. next-worst Physical Health 51.6) and across model tiers (top-tier Opus 4.6 drops 5.0 points from its best-topic Family 56.5 to Romantic 51.5; bottom-tier Grok 4 drops 3.0 points from its best Money 51.3 to Romantic 48.3).

\paragraph{Best-topic identity varies across models.}
Whereas the worst topic is highly consistent, the best topic is not: model peaks span Politics (Opus 4.7, GPT-5.4), Family (Opus 4.6, Sonnet 4.6), Work/School (Haiku 4.5, Mistral Lg, Qwen), Entertainment Media (Gemini 3.1, GPT-5.5), and Money (Grok 4, MiMo). This suggests that what makes a topic ``easy'' for emotional inference depends on model-specific competencies, while what makes Romantic Relationships uniformly hard is a structural feature of the data (interpersonal complexity, ambiguous preference signals) that no model in the current pool overcomes.

Per-topic Kruskal-Wallis omnibus tests, full pairwise contrasts, and Emotion VA / Binary Acc breakdowns are in \texttt{topic\_subset\_output.txt}.

\subsection{Per-Model Breakdown by Participant Diagnosis Group}
\label{sec:supp-diagnosis-permodel}

The main paper (Section~\ref{sec:analysis}) reports diagnosis-level pooled results. Conversations are grouped by self-reported participant diagnosis: \textbf{None} (no diagnoses reported, $n = 107$ conversations from 6 participants), \textbf{Any} (one or more diagnoses, $n = 93$ from 5 participants), and two overlapping subsets of Any: \textbf{AnxDep} (anxiety and/or depression, $n = 73$) and \textbf{ASD/ADHD} (autism spectrum disorder and/or ADHD, $n = 43$). Table~\ref{tab:supp-diagnosis-permodel} gives the per-model Composite breakdown for each group.

\begin{table}[H]
\centering
\caption{Composite Score by model and participant diagnosis group (Default Mode). Group sizes are the number of conversations contributing to each per-model mean. AnxDep and ASD/ADHD are non-disjoint subsets of Any.}
\label{tab:supp-diagnosis-permodel}
\small
\setlength{\tabcolsep}{6pt}
\begin{tabular}{l rrrr | rrr}
\toprule
 & \multicolumn{4}{c|}{\textbf{Composite (mean)}} & \multicolumn{3}{c}{\textbf{$\Delta$ vs.\ None}} \\
\textbf{Model} & \textbf{None} & \textbf{Any} & \textbf{AnxDep} & \textbf{ASD/ADHD} & \textbf{Any} & \textbf{AnxDep} & \textbf{ASD/ADHD} \\
\midrule
Opus 4.7      & 53.01 & 54.37 & 56.00 & 53.09 & $+1.36$ & $+2.99$\textsuperscript{*} & $+0.08$ \\
Opus 4.6      & 53.79 & 54.89 & 56.35 & 53.69 & $+1.09$ & $+2.56$ & $-0.10$ \\
Haiku 4.5     & 52.44 & 53.29 & 54.75 & 52.60 & $+0.85$ & $+2.31$ & $+0.17$ \\
Sonnet 4.6    & 49.53 & 51.18 & 52.90 & 50.27 & $+1.65$ & $+3.37$\textsuperscript{**} & $+0.74$ \\
Gemini 3.1    & 52.87 & 52.94 & 53.75 & 52.31 & $+0.06$ & $+0.88$ & $-0.57$ \\
Mistral Lg    & 52.90 & 50.49 & 51.83 & 49.65 & $-2.41$ & $-1.07$ & $-3.25$ \\
GPT-5.5       & 52.50 & 55.10 & 56.23 & 54.73 & $+2.60$\textsuperscript{*} & $+3.73$\textsuperscript{**} & $+2.23$ \\
GPT-5.4       & 48.15 & 52.64 & 54.08 & 52.06 & $+4.48$\textsuperscript{***} & $+5.93$\textsuperscript{***} & $+3.91$ \\
Qwen 2.5 72B  & 53.26 & 51.05 & 51.30 & 52.06 & $-2.21$ & $-1.96$ & $-1.20$ \\
Grok 4        & 49.89 & 50.43 & 51.41 & 50.41 & $+0.54$ & $+1.52$ & $+0.52$ \\
MiMo-v2-Pro   & 53.02 & 54.29 & 55.23 & 53.25 & $+1.27$ & $+2.21$ & $+0.23$ \\
\midrule
\textbf{Pooled} & 51.94 & 52.79 & 53.98 & 52.19 & $+0.84$\textsuperscript{**} & $+2.04$\textsuperscript{***} & $+0.25$ \\
\bottomrule
\end{tabular}
\par\medskip
\textsuperscript{*} $p_\text{adj} < 0.05$; \textsuperscript{**} $p_\text{adj} < 0.01$; \textsuperscript{***} $p_\text{adj} < 0.001$ (Mann-Whitney U with Holm-Bonferroni across 11 models per metric per comparison).
\end{table}

\paragraph{The AnxDep boost is consistent and large; the ASD/ADHD effect is near zero.}
Conversations from participants reporting anxiety or depression yield higher per-conversation Composite scores than those from no-diagnosis participants for 9 of 11 models (exceptions: Mistral Lg, Qwen 2.5 72B). The AnxDep boost is largest for GPT-5.4 ($+5.93$, $p < 0.001$), GPT-5.5 ($+3.73$, $p < 0.01$), Sonnet 4.6 ($+3.37$, $p < 0.01$), and Opus 4.7 ($+2.99$, $p < 0.05$). At the pooled level, AnxDep produces a $+2.04$ Composite advantage ($p < 0.001$, $\eta^2 = 0.18$ medium effect).

By contrast, the ASD/ADHD vs.\ None contrast is small and non-significant for every model after correction (largest absolute effect: Mistral Lg at $-3.25$, $p_\text{adj} > 0.05$; pooled $\Delta = +0.25$, $p_\text{adj} = 0.36$). This is the dissociation noted in the main paper: anxiety/depression conversations expose performance differences that ASD/ADHD conversations do not.

\paragraph{Interpretation.}
The direction of the AnxDep boost (higher Composite for diagnosed participants) is initially counterintuitive but consistent with the topic-effect interpretation: AnxDep participants may produce richer affective signals (more frequent mood-shift tags, more pronounced PANAS shifts, more distinguishable preferred-vs.-observed gaps), which provides more data for model predictions to be evaluated against. The two models that buck this pattern (Mistral Lg, Qwen 2.5 72B) are also the two with the lowest Pairwise Accuracy in the main results, suggesting they fail to exploit the additional signal that AnxDep conversations provide.

Per-metric breakdowns (Emotion VA, Binary OM/HP, Pairwise Acc by diagnosis group), additional pairwise tests, and full per-model rows for all metrics are in \texttt{diagnosis\_subset\_output.txt}.

\subsection{Emotion Tagging Accuracy vs.\ Intensity Accuracy}
\label{sec:supp-emotion-intensity}

The main paper reports two emotion-tracking metrics, Emotion F1 (set-level tag accuracy) and Emotion VA (continuous valence-arousal similarity), but does not separately report intensity prediction error. The benchmark also captures intensity ratings on a 7-point Likert scale (Section~\ref{sec:design}); models that match an HP's emotion tag still need to predict its intensity. Table~\ref{tab:supp-intensity} reports per-model intensity MAE alongside the two main-paper metrics.

\begin{table}[H]
\centering
\caption{Per-model emotion metrics including intensity MAE (1--7 scale). Lower intensity MAE is better; an MAE of 1.3 corresponds to being off by approximately one intensity level on average. Bold marks the best (lowest) intensity MAE.}
\label{tab:supp-intensity}
\small
\begin{tabular}{l rrr r}
\toprule
\textbf{Model} & \textbf{Intensity MAE} & \textbf{Emotion F1} & \textbf{Emotion VA} & \textbf{Hit Rate} \\
\midrule
Gemini 3.1     & \textbf{1.262} & 0.133 & 0.278 & 0.192 \\
Haiku 4.5      & 1.313 & 0.136 & 0.276 & 0.206 \\
Qwen 2.5 72B   & 1.319 & 0.106 & 0.257 & 0.163 \\
GPT-5.4        & 1.342 & 0.138 & 0.265 & 0.213 \\
MiMo-v2-Pro    & 1.342 & 0.138 & 0.263 & 0.215 \\
Sonnet 4.6     & 1.352 & 0.138 & 0.251 & 0.229 \\
Opus 4.7       & 1.361 & 0.140 & 0.255 & 0.229 \\
Grok 4         & 1.376 & 0.135 & 0.269 & 0.193 \\
Opus 4.6       & 1.390 & 0.138 & 0.250 & 0.230 \\
Mistral Lg     & 1.405 & 0.137 & 0.227 & 0.249 \\
GPT-5.5        & 1.435 & 0.141 & 0.261 & 0.224 \\
\bottomrule
\end{tabular}
\end{table}

\paragraph{Intensity MAE has a narrow range and is approximately decoupled from tag accuracy.}
Intensity MAE spans 1.26 to 1.44 across the 11 models, a range of 0.18 on a 7-point scale. The models with the best (lowest) intensity prediction error are not the same as those leading on Emotion F1 or VA: Gemini 3.1 leads on intensity and is also tied for the VA lead, but Qwen 2.5 72B has the third-lowest intensity MAE despite its outlier-low Emotion F1 (0.106). Conversely, GPT-5.5 has the highest intensity MAE (1.44) but a competitive F1 (0.141, second-highest). Two models can match HP tags equally well yet differ in how accurately they place those tags on the intensity scale.

\paragraph{Hit Rate is reported for completeness but excluded from main paper metrics.}
The Hit Rate column shows set-level recall of HP mood-shift tags (intersection over ground-truth size). It is computed by the scorer but excluded from the Composite and main-paper reporting in favor of Emotion F1, which combines precision and recall and is the more standard set-comparison metric. The Hit Rate ordering does not match the F1 ordering exactly (e.g., Mistral Lg is highest on Hit Rate at 0.249 but trails on F1), reflecting the precision-recall trade-off.

%% ════════════════════════════════════════════════════════════════════════════

\section{Robustness Checks}
\label{sec:supp-robustness}

This section addresses concerns raised in the main paper's Limitations (Section~\ref{sec:limitations}) and methodology note (Section~\ref{sec:analysis}) with explicit robustness analyses.

\subsection{Pairwise Accuracy Stratified by Original Model}
\label{sec:supp-om-stratification}

A potential concern with using Pairwise Accuracy as a primary metric is that the EM is comparing three response variants (original, model-generated, human-edited) where the ``original'' was produced by an Original Model (OM) drawn from the eight-model OM pool described in Section~\ref{sec:method}. If the EM rankings simply tracked overlap between the EM's identity and the OM's identity (e.g., an Anthropic EM rating Anthropic-OM responses more favorably), the headline ranking could be an artifact of OM-EM coupling rather than genuine preference-prediction skill.

Table~\ref{tab:supp-om-stratified} reports Pairwise Accuracy stratified by OM identity. Each row is a different OM; columns are the 11 EMs in the main results.

\begin{table}[H]
\centering
\caption{Pairwise Accuracy stratified by Original Model (OM). Each cell is the mean Pairwise Accuracy across all conversations whose OM was the row model, evaluated by the column EM. The stability of column rankings across rows establishes that EM rankings are not driven by OM-EM identity overlap. EM column abbreviations: Op4.6 = Opus~4.6, GPT55 = GPT-5.5, Op4.7 = Opus~4.7, MiMo = MiMo-v2-Pro, Gem = Gemini~3.1, Hai = Haiku~4.5, Qw = Qwen~2.5~72B, Mst = Mistral~Lg, So = Sonnet~4.6, GPT54 = GPT-5.4, Gk = Grok~4.}
\label{tab:supp-om-stratified}
\scriptsize
\setlength{\tabcolsep}{2.8pt}
\begin{tabular}{lrrrrrrrrrrr}
\toprule
OM & Op4.6 & GPT55 & Op4.7 & MiMo & Gem & Hai & Qw & Mst & So & GPT54 & Gk \\
\midrule
claude-3-haiku ($n=6$)            & 0.691 & 0.489 & 0.700 & 0.528 & 0.418 & 0.476 & 0.216 & 0.500 & 0.479 & 0.350 & 0.361 \\
claude-3.5-sonnet ($n=7$)         & 0.655 & 0.494 & 0.653 & 0.625 & 0.522 & 0.497 & 0.465 & 0.515 & 0.472 & 0.382 & 0.395 \\
gemini-2.0-flash ($n=9$)          & 0.748 & 0.552 & 0.742 & 0.710 & 0.440 & 0.585 & 0.563 & 0.601 & 0.533 & 0.431 & 0.566 \\
gemini-2.5-flash ($n=9$)          & 0.719 & 0.515 & 0.719 & 0.700 & 0.354 & 0.591 & 0.460 & 0.512 & 0.430 & 0.396 & 0.508 \\
gpt-4-turbo-preview ($n=28$)      & 0.628 & 0.536 & 0.635 & 0.539 & 0.508 & 0.561 & 0.440 & 0.511 & 0.509 & 0.466 & 0.438 \\
gpt-4o ($n=8$)                    & 0.760 & 0.565 & 0.757 & 0.721 & 0.536 & 0.640 & 0.452 & 0.550 & 0.533 & 0.386 & 0.468 \\
gpt-4o-mini ($n=119$)             & 0.605 & 0.540 & 0.619 & 0.577 & 0.519 & 0.552 & 0.444 & 0.509 & 0.518 & 0.500 & 0.452 \\
inflection-3-pi ($n=14$)          & 0.708 & 0.554 & 0.700 & 0.704 & 0.509 & 0.609 & 0.530 & 0.545 & 0.526 & 0.435 & 0.552 \\
\midrule
\textit{Default-mode mean}        & \textit{0.637} & \textit{0.538} & \textit{0.646} & \textit{0.598} & \textit{0.503} & \textit{0.560} & \textit{0.450} & \textit{0.517} & \textit{0.512} & \textit{0.470} & \textit{0.461} \\
\bottomrule
\end{tabular}
\end{table}

\paragraph{Rank stability across OMs.}
Spearman rank correlations of EM Pairwise Accuracy rankings across the 8 OM strata: mean $\rho = +0.82$, range $[+0.62, +0.95]$, computed across all $\binom{8}{2} = 28$ OM pairs. Within every OM stratum the two Opus models occupy two of the top three EM positions on Pairwise Accuracy, and Qwen, GPT-5.4, and Grok occupy the bottom three positions on this metric in every stratum except one. The dissociation reported in the main paper---preference-ranking skill is largely separable from first-person behavioral classification, and the Opus family is strong on the former while trailing on the latter---is therefore not an artifact of OM-EM identity coupling. The skill ordering on Pairwise Accuracy is stable across the conversational substrate, even though the substrate (which OM the HP happened to converse with) varies in absolute difficulty. There is no systematic in-family advantage; if anything the direction reverses. Opus EMs' absolute Pairwise Accuracy is higher on out-of-family OM strata --- gpt-4o ($0.760$ for Opus 4.6, $0.757$ for Opus 4.7), gemini-2.0-flash ($0.748$, $0.742$), inflection-3-pi ($0.708$, $0.700$) --- than on the in-family Anthropic OM strata (claude-3-haiku: $0.691$, $0.700$; claude-3.5-sonnet: $0.655$, $0.653$). One reading is that prior-generation Anthropic OMs produced response styles closer to the human-edited ``golden'' targets, leaving less preference-ranking signal for Anthropic EMs to exploit; out-of-family OMs produced more distinguishable response variants, yielding higher Pairwise Accuracy regardless of EM identity. Either way, the headline ranking is not driven by EM-OM provider overlap. OM substrate quality is itself heterogeneous from the HP perspective: HP-reported Four Branch ratings of OMs in the pool range from approximately 4.2 to 6.5 (mean across the 4 branches, conditioned on uneven per-OM sample sizes), confirming the OM substrate spans real variation in EI quality without changing the conclusion above.

\subsection{Per-Participant Headline Findings}
\label{sec:supp-per-participant}

The main paper notes (Section~\ref{sec:analysis}) that per-conversation tests treat conversations as independent observations even though each participant contributes 2--50 conversations, which may overstate effective sample size. Table~\ref{tab:supp-per-participant} addresses this directly: for each of the 11 participants in the Default Mode 200-conversation evaluation, we recompute the headline metrics within their conversations only and identify which model ranks first per metric.

\begin{table}[H]
\centering
\caption{Per-participant headline replication. For each HP we report: $n$, the number of conversations they contributed (per model); the within-participant mean Pairwise Accuracy of the two Opus models; the within-participant mean Binary HP Accuracy of Mistral Large and Opus 4.7; and the model that ranks first within that participant on each of three metrics. The pooled-data finding---Opus family leads Pairwise; Mistral Large is competitive on Binary HP---is observable within most individual participants.}
\label{tab:supp-per-participant}
\footnotesize
\setlength{\tabcolsep}{4pt}
\begin{tabular}{lrrrrrlll}
\toprule
HP & $n$ & Op4.7 PW & Op4.6 PW & Mst BinHP & Op4.7 BinHP & Top PW & Top BinHP & Top Comp \\
\midrule
A00001 & 11 & 0.607 & 0.575 & 0.829 & 0.780 & Opus 4.7 & GPT-5.4 & Opus 4.7 \\
A00002 & 41 & 0.574 & 0.556 & 0.793 & 0.713 & Opus 4.7 & Gemini 3.1 & Qwen 2.5 72B \\
A00003 & 50 & 0.654 & 0.642 & 0.861 & 0.810 & Opus 4.7 & GPT-5.5 & Opus 4.6 \\
A00005 & 2  & 0.578 & 0.621 & 0.822 & 0.791 & GPT-5.4 & Mistral Lg & GPT-5.5 \\
A00006 & 4  & 0.583 & 0.506 & 0.818 & 0.757 & GPT-5.5 & Grok 4 & Opus 4.7 \\
A00007 & 10 & 0.542 & 0.497 & 0.881 & 0.841 & Gemini 3.1 & GPT-5.5 & Gemini 3.1 \\
B00008 & 20 & 0.656 & 0.657 & 0.787 & 0.691 & MiMo-v2-Pro & GPT-5.5 & GPT-5.5 \\
B00009 & 37 & 0.775 & 0.774 & 0.839 & 0.783 & Opus 4.7 & Qwen 2.5 72B & Opus 4.6 \\
B00010 & 12 & 0.640 & 0.656 & 0.860 & 0.831 & Opus 4.6 & Mistral Lg & Opus 4.6 \\
B00011 & 10 & 0.569 & 0.613 & 0.779 & 0.738 & Opus 4.6 & Haiku 4.5 & Opus 4.6 \\
B00013 & 3  & 0.720 & 0.752 & 0.727 & 0.636 & Opus 4.6 & Haiku 4.5 & Opus 4.6 \\
\bottomrule
\end{tabular}
\end{table}

\paragraph{Headline persistence at the participant level.}
An Opus model (4.6 or 4.7) is the top-ranked model on Pairwise Accuracy for 7 of 11 HPs, and an Opus model is the top-ranked model on Composite for 7 of 11 HPs. The four exceptions on Pairwise are concentrated among the smaller-$n$ participants ($n \leq 10$ for three of four), where within-participant means are noisier. The dissociation between preference ranking and first-person binary classification reported in the main paper is also visible at the per-participant level: Opus 4.7's within-participant Binary HP Accuracy is below Mistral Large's in 9 of 11 HPs, mirroring the pooled gap.

The headline patterns are not an artifact of participant pooling. The numerical magnitudes vary, particularly for HPs with fewer than 10 conversations, but the directional findings hold within most individuals.

\subsection{Draft Judge Selection}
\label{sec:supp-rejudge}

Draft~Judge scores in the main paper rely on a single judge LLM (Mistral~Large), which the main paper flags as the rationale for treating Draft~Judge as a supplementary rather than primary metric (Section~\ref{sec:analysis}). To support sensitivity analysis under alternative judge choices, the released code repository ships a standalone re-judging utility (\texttt{rejudge.py}) that re-runs LLM-as-judge scoring on already-completed evaluation result files without re-executing the EM. The utility supports (i) substituting a different judge model on existing outputs, (ii) running multiple judges and aggregating scores (mean for continuous dimensions; majority vote for binary alignment), and (iii) writing rejudged outputs to a separate directory for side-by-side comparison. Researchers extending the benchmark are encouraged to use \texttt{rejudge.py} to verify that Draft~Judge findings of interest are not artifacts of the specific judge LLM chosen here.

%% ════════════════════════════════════════════════════════════════════════════

\section{Extended Methodology}
\label{sec:supp-methods-extended}

This section provides extended rationale and quantitative detail for the methodological decisions and limitations enumerated in main paper Section~\ref{sec:limitations}. Subsections cover topic-assignment design, participant-pool distribution, self-report validation, disclosure constraints, annotation burden, and detailed evaluation-protocol guidance.

\subsection{Topic Assignment Strategy}
\label{sec:supp-topic-assignment}

An adaptive topic-assignment strategy --- using participants' pre-conversation PANAS scores combined with their self-reported topic attitudes --- was contemplated for this round to balance affective-valence coverage (positive, negative, neutral). Implementation was deferred due to limited development resources for the conversational interface; topics were instead assigned via simple randomization from the 50-topic list, with no per-participant repeats. Future iterations may benefit from revisiting adaptive assignment as a mechanism for improving coverage across emotional conditions and ensuring more even distribution of high-intensity affective shifts.

\subsection{Participant Pool and Sampling Distribution}
\label{sec:supp-pool-distribution}

The $n = 11$ pool was kept intentionally small (selected from a larger applicant pool) to allow tighter experimental control during workflow piloting. Per-participant conversation counts range from 2 to 50 (mean 18.2, median 11), creating two analysis implications: (1) statistical tests treating conversations as independent observations may overstate effective sample size --- the rationale for the methodology note in main paper Section~\ref{sec:analysis}; and (2) per-participant subgroup analyses are limited in resolution for participants with smaller contributions.

\subsection{Reliance on Self-Reported Data}
\label{sec:supp-self-report}

The benchmark relies heavily on self-reported data, including participant demographics, mental health diagnostic categories, conversation-topic attitudes, and PANAS affective ratings. None of these were independently verified. Participants were instructed to engage in authentic conversations grounded in their actual experiences and to avoid role play. However, the validity of any individual conversation cannot be fully verified, and informal participant feedback indicated that some conversations, while grounded in real experiences, were framed in the present tense despite referring to past events. Additionally, given the potentially sensitive nature of some topics, participants may have withheld information or modified responses due to discomfort, reducing the realism of those conversations. We mitigated these concerns through pre-screening for willingness to discuss sensitive categories, in-session safeguards (topic reassignment, withdrawal-at-any-time), and post-collection PII review, but the underlying reliance on self-report is a structural limitation of any benchmark grounded in genuine human-model interaction. Cross-validating self-reports with independent psychometric instruments was outside the scope of this v1.0 release.

\subsection{Participant Comfort and Disclosure Constraints}
\label{sec:supp-disclosure}

Although the recruitment pre-screening explicitly sought participants who reported willingness to discuss a range of sensitive topics, and in-session safeguards (topic reassignment, withdrawal rights, clear privacy communication) were implemented, it is reasonable to expect that some participants experienced discomfort during specific conversations. This may have led to selective disclosure, topic avoidance, or modification of responses, particularly for the more emotionally charged categories (Money, Politics, Romantic Relationships). Such effects are difficult to quantify but represent an inherent challenge in studies involving emotionally salient or personal content. The Romantic Relationships outlier finding in the main paper (this topic is the most-difficult topic for 10 of 11 models, see Appendix~\ref{sec:supp-topic-permodel}) may partially reflect such constraints: participants who chose to engage with this topic may have done so under more guarded conditions than for less personally charged topics. We did not collect post-conversation discomfort ratings; future versions of the benchmark would benefit from doing so.

\subsection{Conversation Length and Annotation Burden}
\label{sec:supp-annotation-burden}

The conversation-length distribution skews to the 5--6 turn minimum, with only 38 of 200 conversations exceeding 6 turns --- a direct consequence of per-turn annotation burden. Each turn requires the participant to: review and refine mood-shift tags, complete binary judgments across an LLM-suggested subset (with the option to revise), draft a ``golden'' response, and complete pairwise comparisons across three response variants. For a 10-turn conversation, the cumulative annotation work is substantial. Feedback on interface usability was collected and informed iterative interface improvements during the data-collection window; reducing per-turn friction is a priority for future collection rounds. The conversation-length finding in Section~\ref{sec:analysis} of the main paper (length is approximately orthogonal to model quality) suggests this skew did not introduce systematic bias, but it does limit the benchmark's ability to characterize model performance on extended dialogues.

\subsection{Evaluation Protocol and Fair Use}
\label{sec:supp-fair-use}

Section~\ref{sec:usage} states the usage guidelines for published results. Below we provide rationale for each rule.

\begin{itemize}
  \item \textbf{No training on benchmark data.} The evaluated model must not have been fine-tuned, instruction-tuned, or otherwise trained on any conversations, annotations, or metadata from this dataset. Fine-tuning on the benchmark would compromise its diagnostic value: any future release of a fine-tuned model could appear to ``solve'' the benchmark while actually only memorizing it.
  \item \textbf{No access to human annotations during evaluation.} The EM must not have access to turn-level binary judgments, pairwise preferences, PANAS shift labels, or any other human annotations when generating outputs. These annotations constitute the ground truth against which EM outputs are scored; allowing access during evaluation would short-circuit the inference task and inflate scores artificially.
  \item \textbf{Report evaluation mode.} Results must specify which evaluation mode was used (Default, Omniscient, or Verbose). Default Mode corresponds to the primary benchmark; Omniscient and Verbose Mode results are informative but should not be presented as the primary benchmark score without qualification. Reporting an Omniscient Mode score as a headline number obscures the fact that the EM had access to participant background information that a deployed system would not.
  \item \textbf{No test set contamination.} All conversations were collected through March 2026 and have not been publicly released in full prior to publication of this paper. Users are nonetheless encouraged to verify that evaluated models were not exposed to substantively similar content during training, particularly for model versions trained after March 2026.
\end{itemize}

These guidelines are not technically enforced but constitute the expected standard for results reported in published work using this benchmark. Authors deviating from any of these guidelines (e.g., evaluating in Omniscient Mode without qualification, or fine-tuning on the dataset) should explicitly state and justify the deviation.

%% ════════════════════════════════════════════════════════════════════════════

\section{Human-Baseline Pilot}
\label{sec:supp-human-baseline}

To establish a preliminary point of comparison between Evaluated Models and human predictors, three independent annotators were each asked to predict the original Human Participants' (HPs') annotations on a 7-conversation subset randomly sampled from the 200-conversation Default Mode evaluation. Annotators saw the full conversation transcript and the same anonymized response variants the EMs see, but were not told which response was the OM's actual reply, did not see the HPs' demographic profiles, and had no access to the HPs' actual annotations. The pilot covers Pairwise Accuracy, Binary HP Accuracy, Emotion F1, and the conversation-wide questions (Q1 Goal Identification, Q2 Emotion Clarity, Q3 Conversational Fit). Multi-rater characterization across the full 200-conversation benchmark is left to future work; the present pilot is intended as scoping evidence, not as a stable human ceiling.

\subsection{Per-Rater Accuracy vs.\ HP Ground Truth}
\label{sec:supp-human-baseline-accuracy}

Table~\ref{tab:supp-human-baseline-accuracy} reports each annotator's mean accuracy across the 7 pilot conversations against HP ground truth, alongside the per-conversation best EM mean (the average across conversations of the per-conversation best-performing model) and the pooled across-EM mean for context.

\begin{table}[H]
\centering
\caption{Human-baseline pilot accuracy across 7 conversations. Annotator scores compared against HP ground truth; EM columns show the per-conversation best EM (averaged across the 7 convs) and the pooled mean across all 11 EMs $\times$ 7 convs (Default Mode).}
\label{tab:supp-human-baseline-accuracy}
\footnotesize
\setlength{\tabcolsep}{6pt}
\begin{tabular}{lrrrrr}
\toprule
\textbf{Metric} & \textbf{anno 1} & \textbf{anno 2} & \textbf{anno 3} & \textbf{best EM/conv (mean)} & \textbf{all-EM mean} \\
\midrule
Pairwise Accuracy   & 0.500 & \textbf{0.722} & 0.631 & 0.665 & 0.542 \\
Binary HP Accuracy  & \textbf{0.814} & 0.802 & 0.744 & 0.853 & 0.804 \\
Emotion F1          & 0.179 & 0.192 & 0.168 & --- & --- \\
Q1 Goals (Jaccard)  & 0.190 & 0.119 & 0.071 & --- & --- \\
Q2 Clarity (exact)  & 0.429 & 0.571 & \textbf{0.714} & --- & --- \\
Q3 Fit (exact)      & 0.571 & \textbf{0.714} & 0.286 & --- & --- \\
\bottomrule
\end{tabular}
\end{table}

\begin{figure}[H]
\centering
\includegraphics[width=0.95\linewidth]{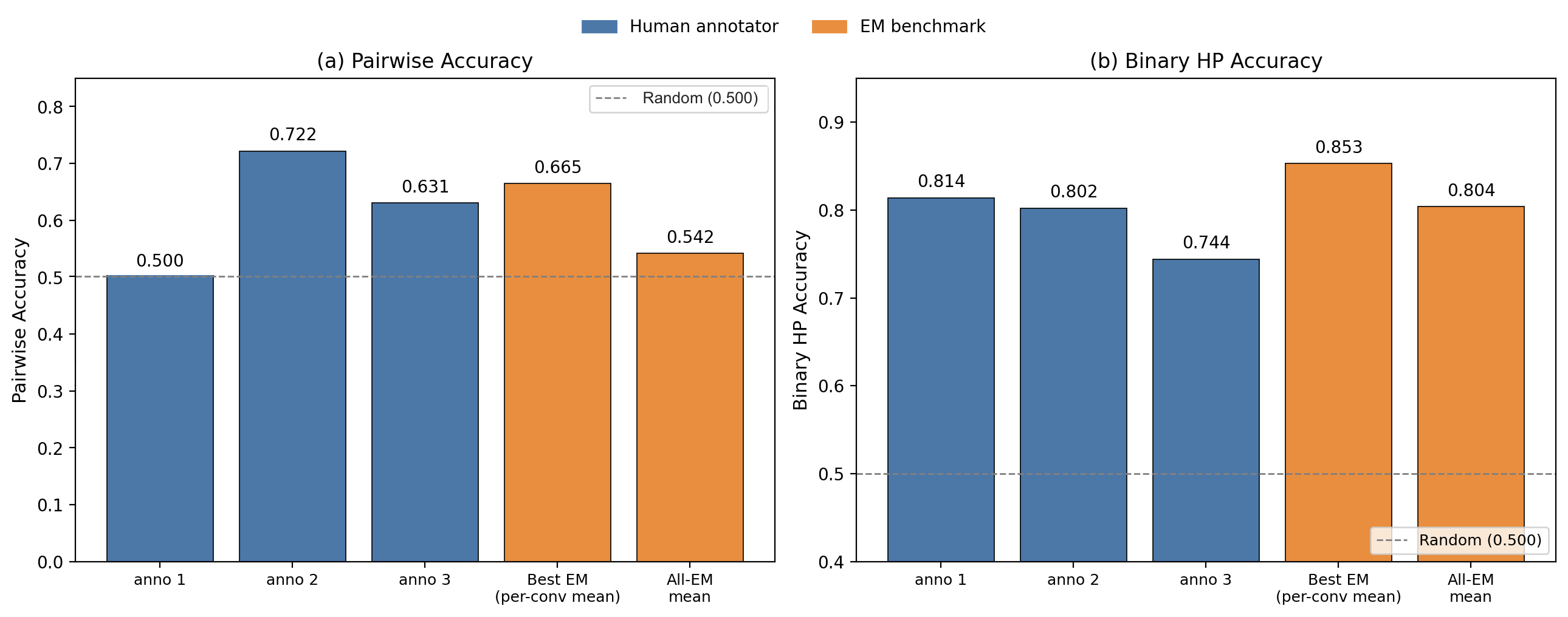}
\caption{Human-baseline pilot accuracy (3 annotators, 7 conversations) compared against EM benchmarks from the full Default Mode evaluation. (a) On Pairwise Accuracy, the strongest annotator (anno 2: 0.722) exceeds the per-conversation best EM mean (0.665). (b) On Binary HP Accuracy, all three annotators sit at or below the per-conversation best EM mean (0.853); the strongest annotator (anno 1: 0.814) is within 4 percentage points.}
\label{fig:supp-hbl-em-comparison}
\end{figure}

\paragraph{Annotators bracket the EM range and exceed it on Pairwise.}
On Pairwise Accuracy, the strongest annotator (anno 2: 0.722) exceeds the per-conversation best EM mean (0.665), suggesting humans can outperform contemporary LLMs at preference prediction on these conversations (Figure~\ref{fig:supp-hbl-em-comparison}). The weakest annotator (anno 1: 0.500) trails substantially, however, indicating large between-annotator variance. On Binary HP Accuracy, all three annotators are at or below the per-conversation best EM mean (0.744--0.814 vs.\ 0.853), with the strongest annotator landing within 4 percentage points of the EM ceiling.

\paragraph{Annotators outperform EMs on Emotion F1.}
All three annotators (range 0.168--0.192) exceed the best EM Emotion F1 (0.141, GPT-5.5) on the full 200-conversation benchmark. This is consistent with the main paper's observation that Emotion F1 is the metric where models are weakest; even a single-rater human prediction of HP-reported mood-shift tags yields higher set F1 than any model in the pool.

\paragraph{Annotators collapse on Q1 Goal Identification.}
On Q1 Goal Identification, all three annotators score 0.07--0.19, far below the best EM (GPT-5.4: 0.71) on the full 200. Q1 asks predictors to identify what the HP was looking for in the conversation (e.g.,~\textit{vent / get advice / feel less alone}); annotators predict from the conversation alone, while EMs may capture goal structure through information accumulated across the conversational trajectory in ways human annotators do not. This is exploratory and the gap may also reflect annotator strategy (annotators may have under-selected goals from the multi-select list).

\subsection{Inter-Rater Agreement}
\label{sec:supp-human-baseline-irr}

Table~\ref{tab:supp-human-baseline-irr} reports pairwise agreement across the three rater pairs. Cohen's $\kappa$ is reported for binary metrics where the chance baseline is meaningful; raw agreement \% is reported for all metrics.

\begin{table}[H]
\centering
\caption{Inter-rater agreement across the three rater pairs. Cohen's $\kappa$ shown for Binary HP and Pairwise (binary winner). Q1 reports Jaccard set similarity; emotion tags report mean per-turn Jaccard.}
\label{tab:supp-human-baseline-irr}
\footnotesize
\begin{tabular}{lrrrr}
\toprule
\textbf{Pair} & \textbf{Pairwise (\%/$\kappa$)} & \textbf{Binary HP (\%/$\kappa$)} & \textbf{Emotion (Jaccard)} & \textbf{Q2/Q3 exact (\%)} \\
\midrule
anno 1 vs.\ anno 2 & 61.2\% / +0.21 & 76.4\% / +0.44 & 25.4\% & 28.6\% / 28.6\% \\
anno 1 vs.\ anno 3 & 49.1\% / $-$0.02 & 70.9\% / +0.38 & 21.4\% & 42.9\% / 57.1\% \\
anno 2 vs.\ anno 3 & 62.6\% / +0.23 & 76.6\% / +0.51 & 20.7\% & 42.9\% / 28.6\% \\
\midrule
\textit{Mean} & \textit{57.6\%} & \textit{74.6\%} & \textit{22.5\%} & \textit{38.1\% / 38.1\%} \\
\bottomrule
\end{tabular}
\end{table}

Binary HP shows moderate agreement (Cohen's $\kappa = +0.38$ to $+0.51$); Pairwise shows fair-to-poor agreement (Cohen's $\kappa = -0.02$ to $+0.23$), reflecting genuine disagreement on preference among humans even on the same conversation. The annotator pair with the lowest Pairwise agreement (anno 1 vs.\ anno 3, $\kappa = -0.02$) is also the pair with the largest demographic distance (anno 1: F, 18--24, with anxiety; anno 3: M, 25--34, no diagnoses).

\subsection{Self-Rating Disclosure}
\label{sec:supp-human-baseline-self-rating}

One of the three annotators (anno 3) is the HP for one of the seven pilot conversations (\texttt{df8737a9}, participant ID A00002). They were asked to predict the same annotations they had originally provided as the HP for that conversation. We report this transparently rather than excluding the conversation: the comparison illuminates rater consistency as much as the prediction task itself.

Anno 3's accuracy on their own conversation is only modestly above their mean across the other six conversations: Binary HP $+7.1$ pp (0.805 vs.\ 0.734 mean other), Pairwise $-3.6$ pp (0.600 vs.\ 0.636 mean other), Q2 Clarity 100\% match (their own self-reported clarity), and Q3 Fit \emph{0\%} match (the HP did not reproduce their own original Q3 rating when re-asked weeks later). This suggests participants do not perfectly recall their own past responses to subjective questions, and that the HP-as-ground-truth standard itself includes within-individual response noise. Removing this conversation from anno 3's mean shifts their pooled scores by less than 0.02 on every metric, so the headline numbers are not meaningfully affected.

\subsection{Annotator-Trait Matching (Exploratory)}
\label{sec:supp-human-baseline-trait-match}

Did annotators predict more accurately for HPs whose demographic and psychometric profiles resembled their own? We computed per-trait similarity scores between each (annotator, HP) pair and correlated them with annotator accuracy across the 21 (annotator, conversation) cells for the categorical traits (gender, age range, education, diagnosis bucket), and across 14 cells for the continuous psychometric traits where two of the three annotators have full TIPI / WHO-5 / PROMIS / ASRS / AQ-10 scores available. With $n \leq 21$, only $|r|>0.43$ reaches conventional two-tailed significance ($p<0.05$); all results below should be read as exploratory.

\textbf{No trait-similarity correlation survives at the significance threshold consistently across metrics.} The strongest categorical signal is shared education level with Binary HP Accuracy ($r=+0.54$, just above threshold); shared gender shows $r=+0.40$ on the same metric (below threshold). An earlier two-rater preliminary analysis suggested a much stronger gender-match signal ($r=+0.78$ with $n=14$), but adding the third annotator (a Male predicting predominantly Female HPs at competitive accuracy) collapsed the apparent effect. We report this trajectory transparently to discourage over-interpretation of small-sample trait-match findings in EI evaluation.

The diagnosis-bucket analyses (AnxDep match, ASD/ADHD match) are largely uninterpretable in this pilot: the 7 conversations include only 1 AnxDep HP and 1 ASD/ADHD HP, so any correlation involving these matches is driven by single conversations. The continuous-trait correlations (TIPI subscales, WHO-5, PROMIS scales) similarly do not reach the significance threshold and lack a theoretical interpretation we are willing to defend at this sample size.

The honest summary is that trait similarity between annotator and HP did \emph{not} robustly predict prediction accuracy in this 7-conversation pilot. This is a useful preliminary negative finding: it argues against the strong claim that EI evaluation requires demographically-matched annotators, while leaving open the possibility that effects exist at sample sizes this pilot cannot detect.

\subsection{Pilot Limitations}

\begin{itemize}
  \item \textbf{Sample size.} Three annotators across seven conversations yields 21 (rater, conversation) cells (14 with full demographics for the trait-match analysis). Underpowered for inferential conclusions; the pilot is exploratory.
  \item \textbf{HP demographic distribution.} The 7 pilot HPs span a narrow demographic range (predominantly Female, ages 25--34, US, no diagnoses), limiting the variation available for trait-matching and gender/age effects.
  \item \textbf{Annotator demographic distribution.} The 3 annotators do not cover all the demographic combinations represented among the 11 HPs in the full benchmark.
  \item \textbf{One self-rating case.} Anno 3 is the HP for one conversation; we report this transparently and show their accuracy on that conversation is not anomalous (\S\ref{sec:supp-human-baseline-self-rating}).
  \item \textbf{Single-task framing.} Annotators predicted HP responses from the conversation alone, without access to the HPs' pre-conversation PANAS or psychometric profile (matching Default Mode for EMs). A fuller pilot might include a comparison with annotators who do see the HP profile.
\end{itemize}

A larger-scale human-baseline characterization (more annotators, full 200-conversation evaluation, balanced annotator demographic distribution, formal IRR including weighted-$\kappa$ on ordinal Q3) is planned for future iterations of the benchmark.

%% ════════════════════════════════════════════════════════════════════════════

\section{Additional Figures and Tables}
\label{sec:supp-figures}

This section collects the full per-mode figure galleries, full per-metric tables (Default, Verbose, Omniscient), correlation matrices, and the dense per-model breakdowns whose main-body summaries appear in Section~\ref{sec:results} and Section~\ref{sec:analysis}.

\subsection{Full Per-Metric Results Tables}
\label{sec:supp-full-tables}

Table~\ref{tab:supp-default-full} reports the full per-model means across all primary metrics for the Default Mode evaluation ($n = 200$). Tables~\ref{tab:supp-verbose-full} and~\ref{tab:supp-omn-full} report the same metrics for Verbose and Omniscient modes respectively, allowing direct three-way comparison without the abbreviation-heavy mode-effect tables typically used in the main body.

\begin{table}[H]
\centering
\caption{Default Mode per-model means across primary metrics ($n=200$ conversations per model). Sorted by Composite descending.}
\label{tab:supp-default-full}
\footnotesize
\setlength{\tabcolsep}{3.5pt}
\begin{tabular}{lrrrrrrrr}
\toprule
\textbf{Model} & \textbf{Comp} & \textbf{PW Acc} & \textbf{K$\tau$} & \textbf{EmoVA} & \textbf{EmoF1} & \textbf{BinOM} & \textbf{BinHP} & \textbf{Judge} \\
\midrule
Opus 4.6      & 54.30 & 0.637 & $+$0.301 & 0.250 & 0.138 & 0.844 & 0.774 & 0.844 \\
GPT-5.5       & 53.71 & 0.538 & $+$0.024 & 0.261 & 0.141 & 0.854 & 0.821 & 0.826 \\
Opus 4.7      & 53.64 & 0.646 & $+$0.338 & 0.255 & 0.140 & 0.843 & 0.767 & 0.833 \\
MiMo-v2-Pro   & 53.61 & 0.598 & $+$0.213 & 0.263 & 0.138 & 0.849 & 0.797 & 0.782 \\
Gemini 3.1    & 52.90 & 0.503 & $-$0.015 & 0.278 & 0.133 & 0.859 & 0.818 & 0.813 \\
Haiku 4.5     & 52.83 & 0.560 & $+$0.124 & 0.276 & 0.136 & 0.832 & 0.802 & 0.806 \\
Qwen 2.5 72B  & 52.23 & 0.450 & $-$0.127 & 0.257 & 0.106 & 0.859 & 0.817 & 0.691 \\
Mistral Lg    & 51.78 & 0.517 & $+$0.029 & 0.227 & 0.137 & 0.862 & 0.827 & 0.796 \\
Sonnet 4.6    & 50.29 & 0.512 & $-$0.030 & 0.251 & 0.138 & 0.841 & 0.791 & 0.814 \\
GPT-5.4       & 50.24 & 0.470 & $-$0.135 & 0.265 & 0.138 & 0.839 & 0.799 & 0.807 \\
Grok 4        & 50.14 & 0.461 & $-$0.125 & 0.269 & 0.135 & 0.836 & 0.779 & 0.795 \\
\bottomrule
\end{tabular}
\end{table}

\begin{table}[H]
\centering
\caption{Verbose Mode per-model means across primary metrics, computed on the 50-conversation Verbose evaluation subsample ($n = 50$ per model). Sorted by Composite descending. Mistral Large degrades on Pairwise Accuracy under Verbose Mode (main paper Section~\ref{sec:analysis}); the absolute drop is visible in the K$\tau$ column.}
\label{tab:supp-verbose-full}
\footnotesize
\setlength{\tabcolsep}{3.5pt}
\begin{tabular}{lrrrrrrrr}
\toprule
\textbf{Model} & \textbf{Comp} & \textbf{PW Acc} & \textbf{K$\tau$} & \textbf{EmoVA} & \textbf{EmoF1} & \textbf{BinOM} & \textbf{BinHP} & \textbf{Judge} \\
\midrule
Opus 4.7      & 53.95 & 0.657 & $+$0.350 & 0.257 & 0.149 & 0.804 & 0.749 & 0.785 \\
Opus 4.6      & 53.50 & 0.632 & $+$0.309 & 0.234 & 0.130 & 0.842 & 0.769 & 0.844 \\
GPT-5.5       & 53.44 & 0.532 & $-$0.009 & 0.249 & 0.129 & 0.839 & 0.802 & 0.823 \\
MiMo-v2-Pro   & 52.71 & 0.574 & $+$0.141 & 0.252 & 0.120 & 0.851 & 0.802 & 0.790 \\
Qwen 2.5 72B  & 51.48 & 0.477 & $-$0.047 & 0.237 & 0.093 & 0.852 & 0.815 & 0.728 \\
Haiku 4.5     & 51.47 & 0.540 & $+$0.093 & 0.262 & 0.114 & 0.828 & 0.794 & 0.803 \\
Gemini 3.1    & 51.21 & 0.500 & $-$0.001 & 0.272 & 0.141 & 0.829 & 0.788 & 0.809 \\
Sonnet 4.6    & 49.73 & 0.500 & $-$0.076 & 0.235 & 0.128 & 0.840 & 0.786 & 0.815 \\
Grok 4        & 49.70 & 0.481 & $-$0.059 & 0.267 & 0.122 & 0.823 & 0.769 & 0.803 \\
Mistral Lg    & 49.66 & 0.480 & $-$0.009 & 0.215 & 0.127 & 0.854 & 0.818 & 0.810 \\
GPT-5.4       & 49.10 & 0.463 & $-$0.193 & 0.255 & 0.123 & 0.827 & 0.785 & 0.798 \\
\bottomrule
\end{tabular}
\end{table}

\begin{table}[H]
\centering
\caption{Omniscient Mode per-model means across primary metrics, computed on the 25-conversation Omniscient evaluation subsample ($n = 25$ per model). Sorted by Composite descending. Note the Opus 4.7 Draft Judge collapse to 0.506 (main paper Section~\ref{sec:analysis}); all other models' Judge scores remain in the 0.7--0.85 range.}
\label{tab:supp-omn-full}
\footnotesize
\setlength{\tabcolsep}{3.5pt}
\begin{tabular}{lrrrrrrrr}
\toprule
\textbf{Model} & \textbf{Comp} & \textbf{PW Acc} & \textbf{K$\tau$} & \textbf{EmoVA} & \textbf{EmoF1} & \textbf{BinOM} & \textbf{BinHP} & \textbf{Judge} \\
\midrule
Opus 4.6      & 55.01 & 0.656 & $+$0.366 & 0.276 & 0.153 & 0.827 & 0.777 & 0.846 \\
Opus 4.7      & 54.78 & 0.695 & $+$0.411 & 0.308 & 0.194 & 0.770 & 0.685 & 0.506 \\
Gemini 3.1    & 53.75 & 0.491 & $-$0.059 & 0.313 & 0.150 & 0.837 & 0.795 & 0.821 \\
Haiku 4.5     & 53.38 & 0.570 & $+$0.146 & 0.285 & 0.133 & 0.810 & 0.797 & 0.806 \\
MiMo-v2-Pro   & 53.00 & 0.612 & $+$0.237 & 0.289 & 0.153 & 0.833 & 0.787 & 0.779 \\
GPT-5.5       & 52.85 & 0.531 & $+$0.020 & 0.285 & 0.151 & 0.819 & 0.798 & 0.836 \\
Qwen 2.5 72B  & 52.67 & 0.478 & $-$0.018 & 0.263 & 0.118 & 0.818 & 0.794 & 0.708 \\
Mistral Lg    & 51.53 & 0.529 & $+$0.052 & 0.225 & 0.153 & 0.843 & 0.804 & 0.794 \\
Grok 4        & 50.31 & 0.491 & $-$0.051 & 0.276 & 0.130 & 0.805 & 0.762 & 0.788 \\
Sonnet 4.6    & 49.34 & 0.501 & $-$0.066 & 0.261 & 0.141 & 0.825 & 0.769 & 0.802 \\
GPT-5.4       & 49.06 & 0.450 & $-$0.179 & 0.268 & 0.125 & 0.811 & 0.785 & 0.802 \\
\bottomrule
\end{tabular}
\end{table}

\subsection{Mode-Effect Detail Figures}
\label{sec:supp-mode-detail}

\begin{figure}[H]
\centering
\includegraphics[width=\linewidth]{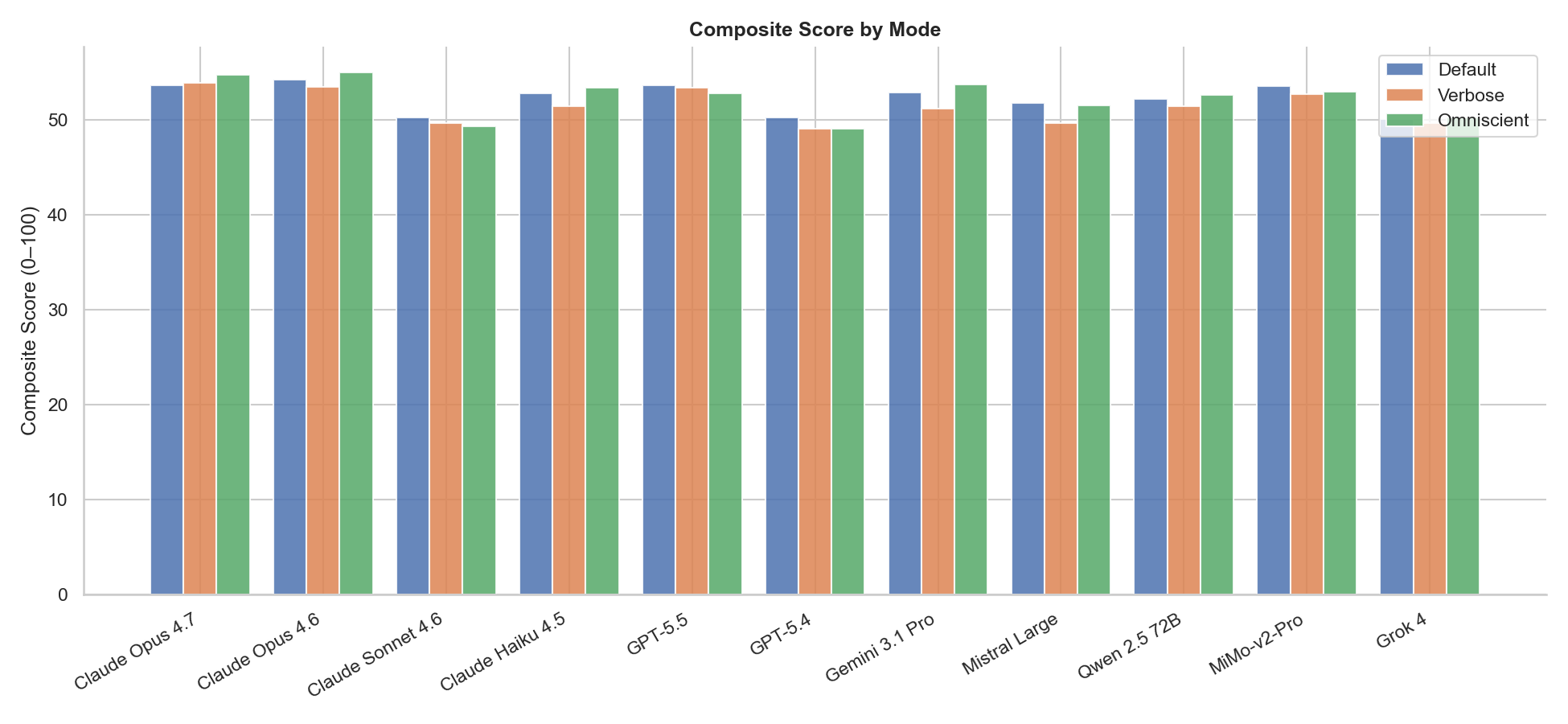}
\caption{Composite Score per model, broken out by evaluation mode (Default, Verbose, Omniscient). Most models cluster within a 1--2 point band across modes; Opus 4.7 Omniscient is the largest cross-mode outlier.}
\label{fig:supp-composite-by-mode}
\end{figure}

\begin{figure}[H]
\centering
\includegraphics[width=\linewidth]{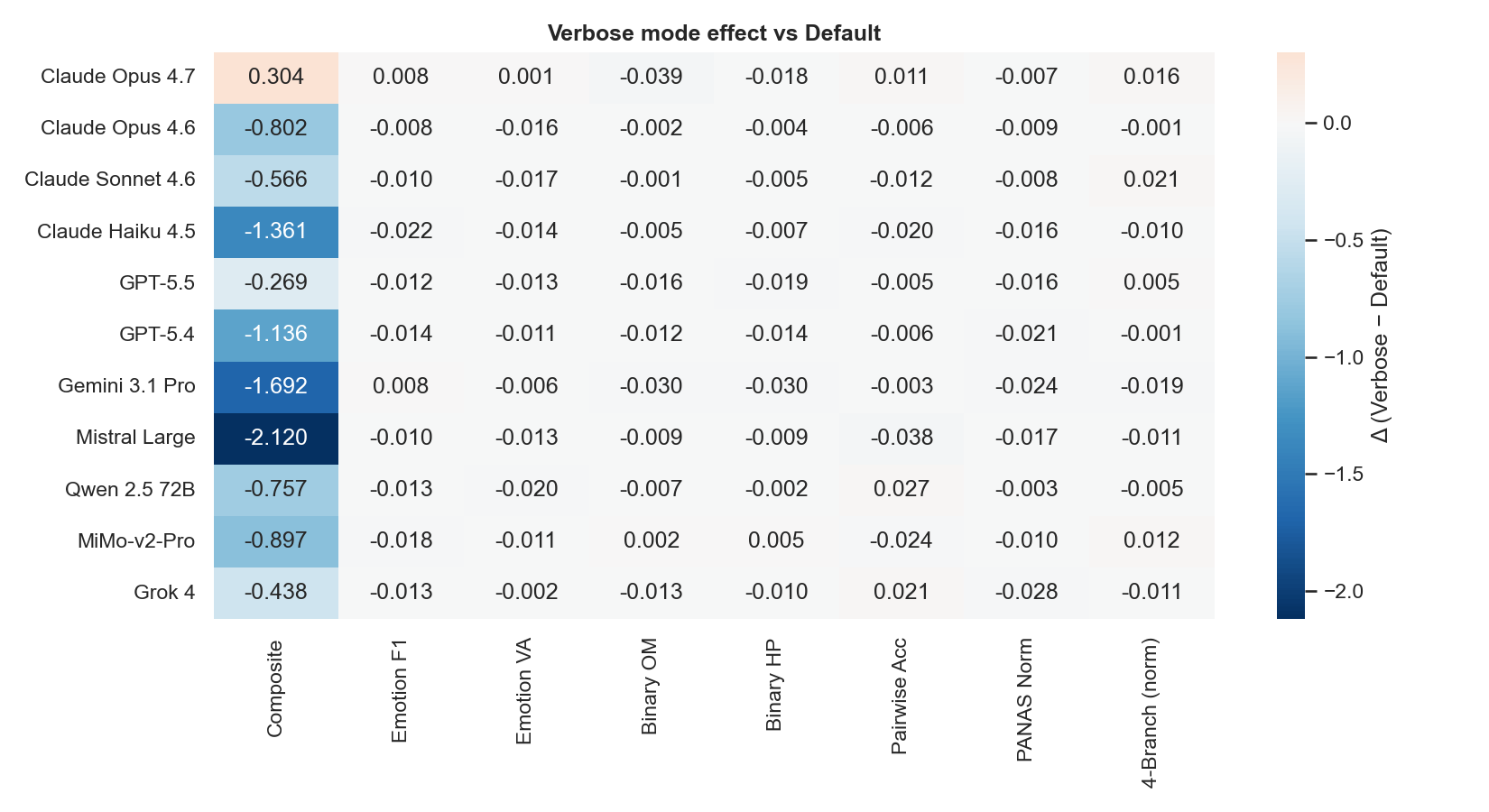}
\caption{Verbose -- Default difference per model (rows) and metric (columns). Positive (red) cells indicate Verbose Mode improves on the metric; negative (blue) cells indicate degradation. Mistral Large's Composite degradation ($-2.12$) is the single most prominent cell.}
\label{fig:supp-mode-verbose-heatmap}
\end{figure}

\begin{figure}[H]
\centering
\includegraphics[width=\linewidth]{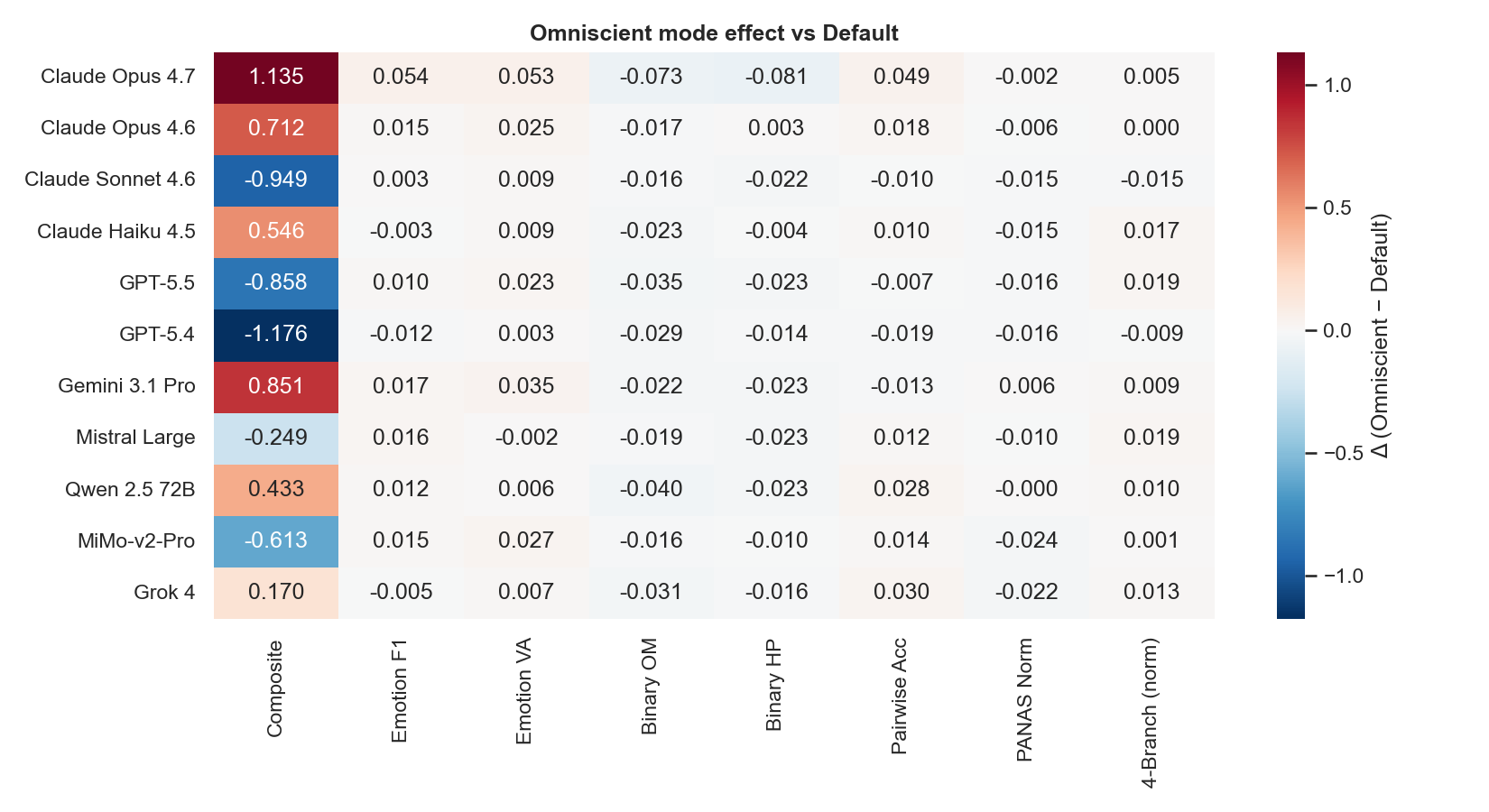}
\caption{Omniscient -- Default difference per model (rows) and metric (columns). Positive (red) cells indicate Omniscient Mode improves on the metric; negative (blue) cells indicate degradation. Composite shifts vary by model (Opus 4.7 $+1.14$, GPT-5.5 $-0.86$); per-metric shifts are mostly near-zero, supporting the main paper's conclusion that profile access does not systematically improve performance.}
\label{fig:supp-mode-omn-heatmap}
\end{figure}

The main paper (Section~\ref{sec:analysis}) reports mode-effect findings as significant exceptions; the figures in this subsection show the full $11 \times \text{metric}$ effect matrix and rank dynamics across modes. Composite Score per model broken out by mode is in Figure~\ref{fig:supp-composite-by-mode}; Verbose~$-$~Default and Omniscient~$-$~Default difference matrices are in Figures~\ref{fig:supp-mode-verbose-heatmap} and~\ref{fig:supp-mode-omn-heatmap}. Verbose and Omniscient evaluations were run on smaller conversation subsamples ($n = 50$ and $n = 25$ per model respectively) than the Default Mode evaluation ($n = 200$); per-model means in this section reflect those subsample sizes.

\paragraph{Within-family rank flip: Opus 4.6 $\leftrightarrow$ Opus 4.7 under Verbose Mode.}
On the Default Mode 200-conversation evaluation, Opus 4.6 leads Opus 4.7 on Composite ($54.30$ vs.\ $53.64$). Under Verbose Mode the ordering reverses: Opus 4.7 leads Opus 4.6 on Composite ($53.95$ vs.\ $53.50$), Pairwise Accuracy ($0.657$ vs.\ $0.632$), and Kendall $\tau$ ($+0.350$ vs.\ $+0.309$). The flip is consistent across the three turn-level discrimination metrics and is unique to the Anthropic family pair: no other within-provider model pair in the panel switches relative rank between Default and Verbose Mode. Reasoning traces appear to confer a selective advantage on Opus 4.7's preference prediction without correspondingly improving Opus 4.6.

\paragraph{Opus 4.7 Omniscient Mode: best preference prediction in the panel, judge-format collapse.}
The Omniscient Mode subsample produces Opus 4.7's highest absolute Pairwise Accuracy and Kendall $\tau$ across any model in any mode in this benchmark ($0.695$ and $+0.411$ respectively, up from $0.646$ and $+0.338$ in Default Mode and $0.657$ and $+0.350$ in Verbose). At the same time, Draft Judge crashes to $0.506$ from $0.833$ (Default), while every other model's Omniscient-Mode Judge score remains in the $0.71$--$0.85$ band. The dissociation between the two effects is informative: when scored by metrics grounded in HP-vs.-EM agreement (Pairwise, Kendall $\tau$, Emotion VA, Emotion F1), Opus 4.7 strictly improves under profile access; only the judge-LLM-rated Draft Judge (a single-judge proxy noted as supplementary throughout the paper) drops. This supports the main paper's framing of the Opus 4.7 Omniscient Judge collapse as an output-format interaction with the judge model rather than a real degradation in inference quality.

\FloatBarrier

\subsection{Spearman Correlation Matrix Across Primary Metrics}
\label{sec:supp-corr}

Table~\ref{tab:supp-corr-matrix} reports the full Spearman rank correlation matrix among the eight primary metrics, computed across all 2200 Default Mode per-conversation observations (11 models $\times$ 200 conversations). The main paper cites a handful of pairwise correlations from this matrix in Section~\ref{sec:analysis}; the full matrix is included here for transparency and to support meta-analytic uses of the benchmark data.

\begin{table}[H]
\centering
\caption{Spearman rank correlations among primary metrics, computed across all 2200 Default Mode per-conversation observations. Magnitudes $\geq 0.3$ in bold (excluding the diagonal). Three correlation clusters are visible: Pairwise Acc $\leftrightarrow$ K$\tau$ ($+0.89$), Emotion VA $\leftrightarrow$ Emotion F1 ($+0.61$), and Binary OM $\leftrightarrow$ Binary HP ($+0.60$). Pairwise/K$\tau$ correlations with Binary OM/HP are weakly negative, reinforcing the dissociation discussed in main paper Section~\ref{sec:analysis}.}
\label{tab:supp-corr-matrix}
\footnotesize
\setlength{\tabcolsep}{4pt}
\begin{tabular}{lrrrrrrrr}
\toprule
 & Comp & PWAcc & K$\tau$ & EmoVA & EmoF1 & BinOM & BinHP & Judge \\
\midrule
Comp   & $+$1.00 & \textbf{$+$0.43} & \textbf{$+$0.35} & \textbf{$+$0.48} & \textbf{$+$0.44} & \textbf{$+$0.37} & \textbf{$+$0.34} & $+$0.25 \\
PWAcc  & \textbf{$+$0.43} & $+$1.00 & \textbf{$+$0.89} & $-$0.04 & $+$0.03 & $-$0.03 & $-$0.10 & $+$0.12 \\
K$\tau$ & \textbf{$+$0.35} & \textbf{$+$0.89} & $+$1.00 & $-$0.06 & $+$0.02 & $-$0.04 & $-$0.11 & $+$0.08 \\
EmoVA  & \textbf{$+$0.48} & $-$0.04 & $-$0.06 & $+$1.00 & \textbf{$+$0.61} & $+$0.11 & $+$0.11 & $+$0.12 \\
EmoF1  & \textbf{$+$0.44} & $+$0.03 & $+$0.02 & \textbf{$+$0.61} & $+$1.00 & $+$0.04 & $+$0.00 & $+$0.09 \\
BinOM  & \textbf{$+$0.37} & $-$0.03 & $-$0.04 & $+$0.11 & $+$0.04 & $+$1.00 & \textbf{$+$0.60} & $+$0.23 \\
BinHP  & \textbf{$+$0.34} & $-$0.10 & $-$0.11 & $+$0.11 & $+$0.00 & \textbf{$+$0.60} & $+$1.00 & $+$0.20 \\
Judge  & $+$0.25 & $+$0.12 & $+$0.08 & $+$0.12 & $+$0.09 & $+$0.23 & $+$0.20 & $+$1.00 \\
\bottomrule
\end{tabular}
\end{table}

\FloatBarrier

\subsection{Default Mode Headline Figure Gallery}
\label{sec:supp-default-gallery}

\begin{figure}[!htbp]
\centering
\includegraphics[width=\linewidth]{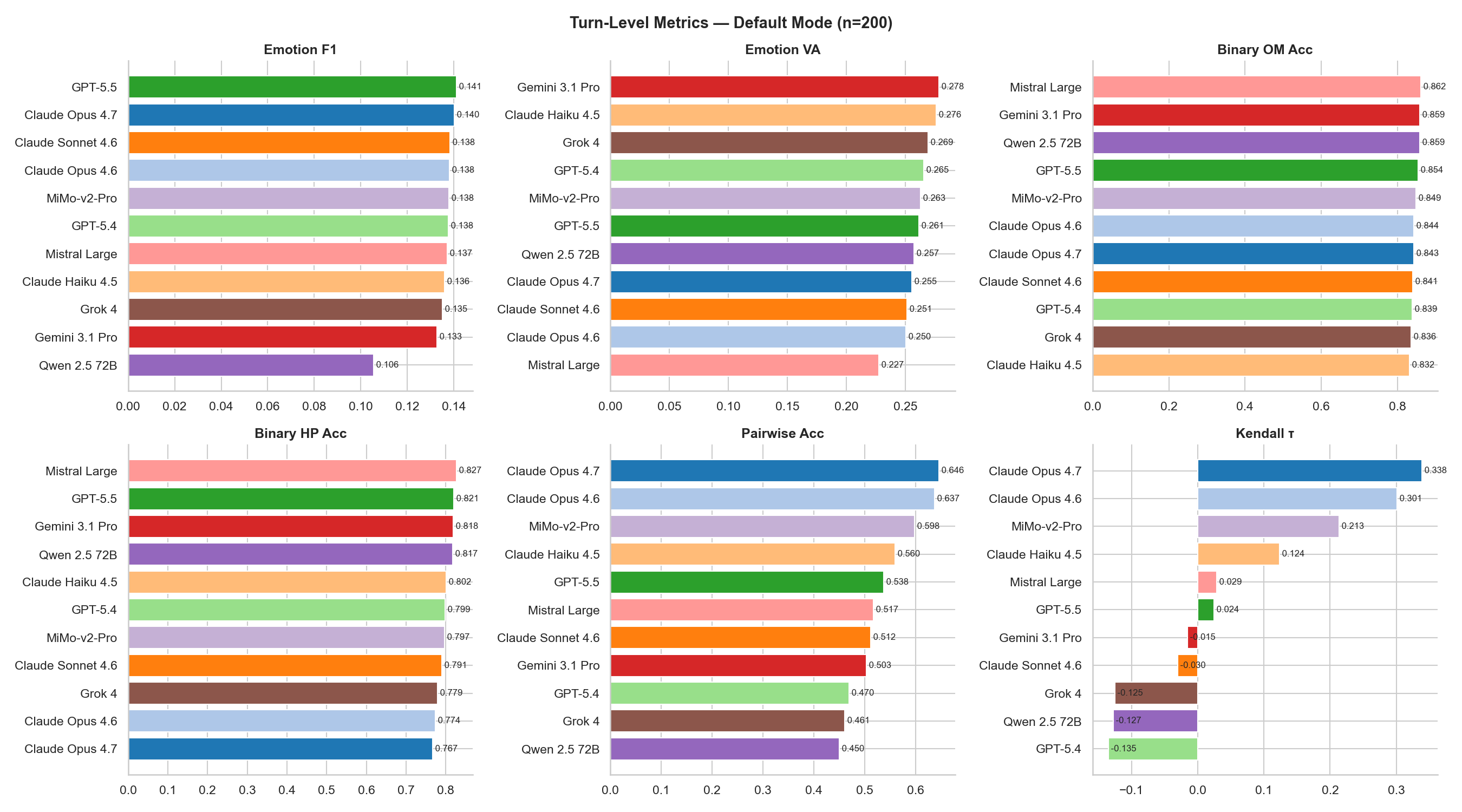}
\caption{Turn-level metric grid: Emotion F1, Emotion VA, Binary OM/HP Accuracy, Pairwise Accuracy, Kendall $\tau$ (Default Mode).}
\label{fig:supp-def-turn-grid}
\end{figure}

\begin{figure}[!htbp]
\centering
\includegraphics[width=\linewidth]{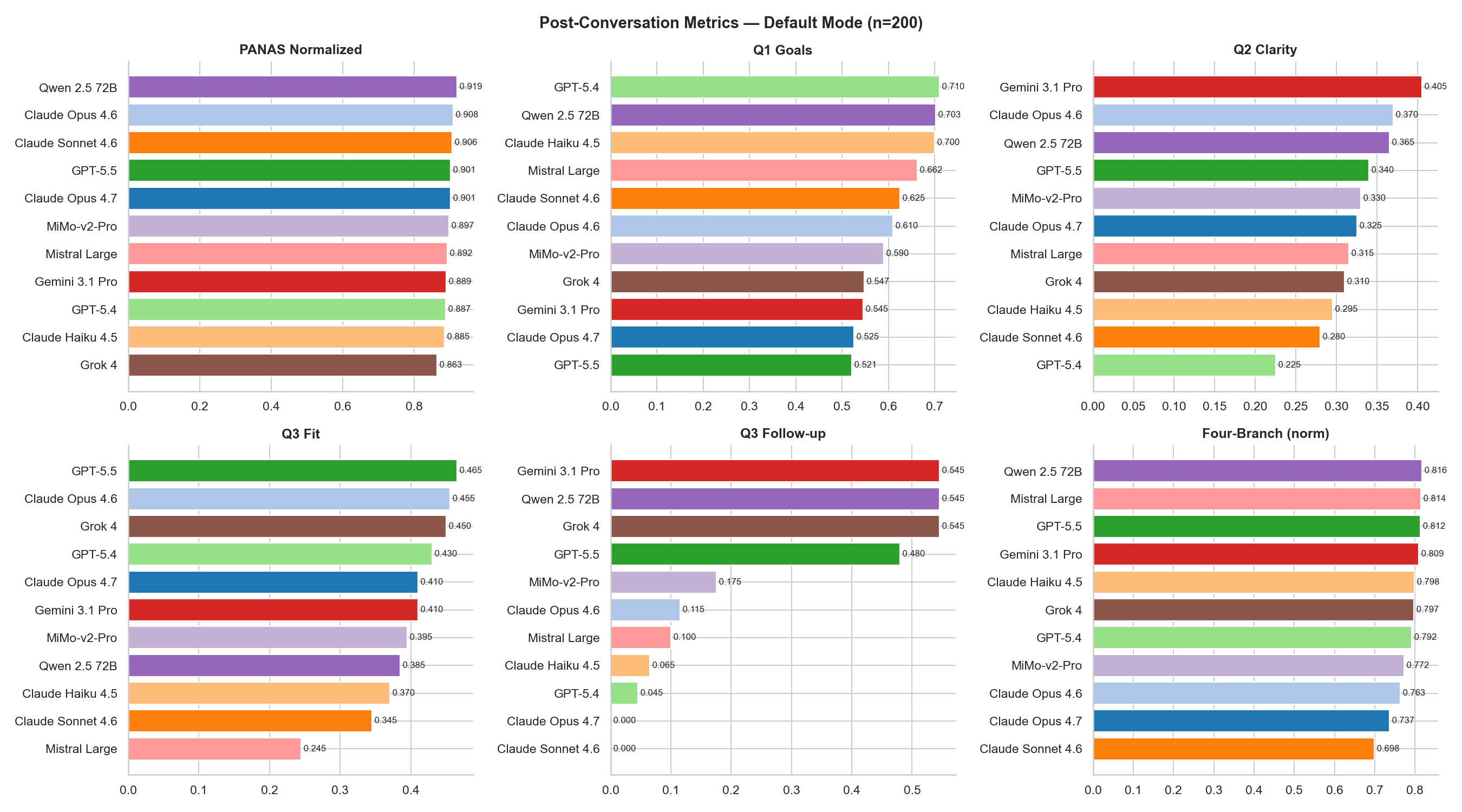}
\caption{Post-conversation metric grid: PANAS Norm, Q1 Goal Identification, Q2 Emotion Clarity, Q3 Conversational Fit, Q3 Follow-up, Four-Branch (Default Mode).}
\label{fig:supp-def-post-grid}
\end{figure}

\begin{figure}[!htbp]
\centering
\includegraphics[width=0.6\linewidth]{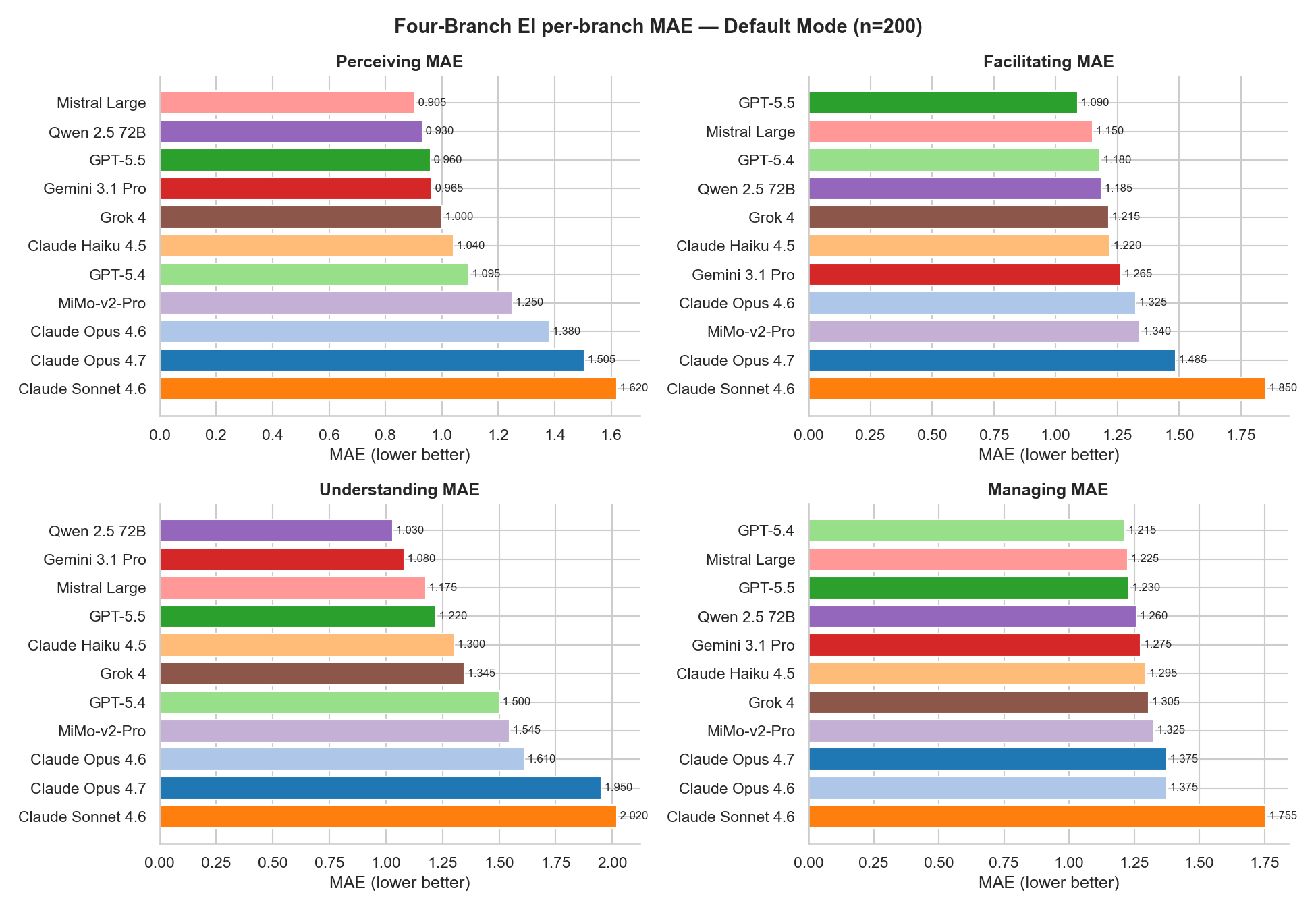}
\caption{Four-Branch MAE per branch (Perceiving, Understanding, Facilitating, Managing) for each model. ``Understanding'' is consistently the hardest branch and ``Perceiving'' the easiest, a shared pattern across models.}
\label{fig:supp-def-fourbranch-per}
\end{figure}

\begin{figure}[!htbp]
\centering
\includegraphics[width=0.85\linewidth]{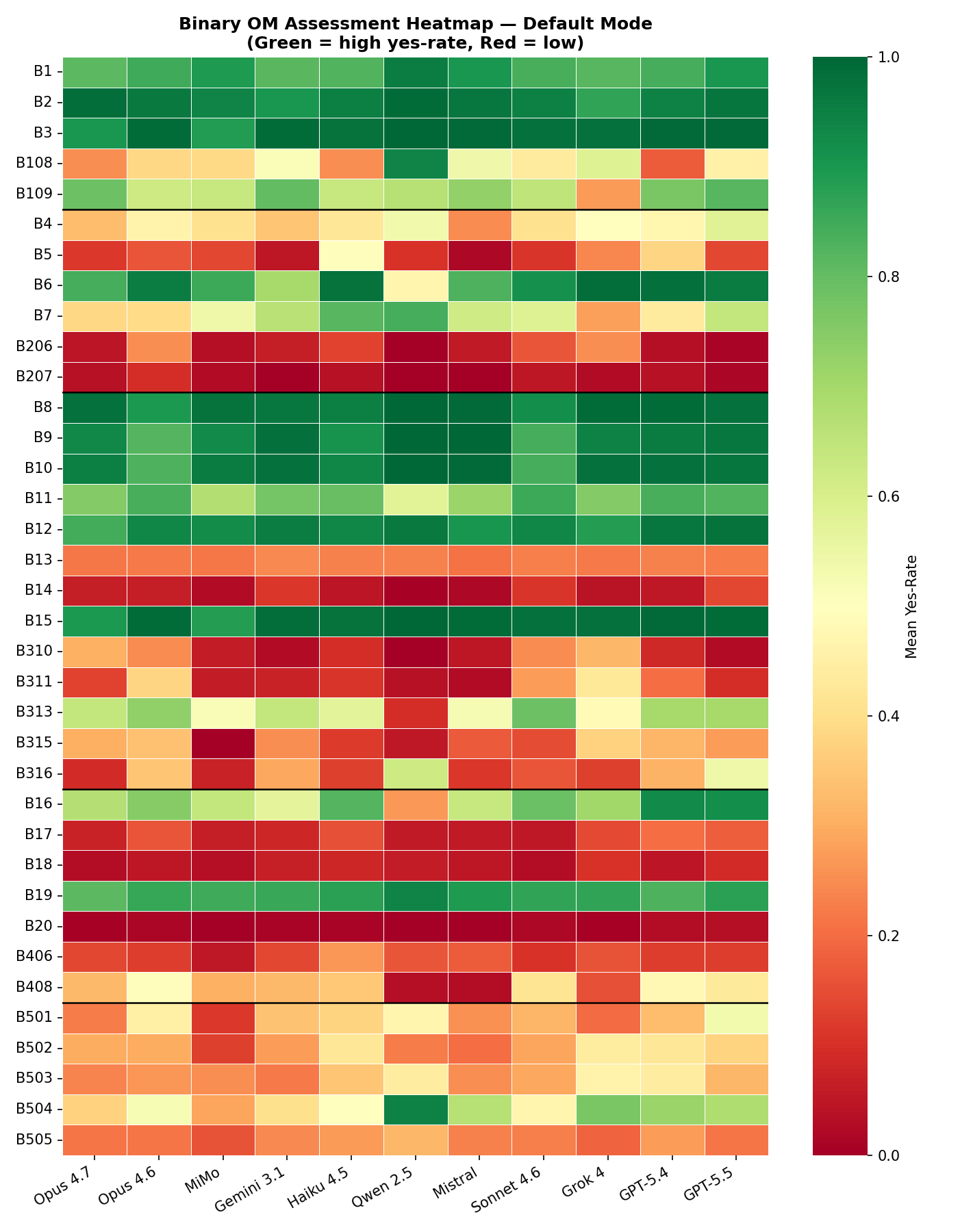}
\caption{Binary OM yes-rate heatmap: rows = behavior question (B-codes, grouped by question family with horizontal dividers), columns = evaluated model, cell value = mean rate at which the EM answered ``yes'' to that behavior question across all 200 Default Mode conversations. Green indicates consistently high yes-rates; red indicates consistently low yes-rates. The figure shows the response distribution underlying the Binary OM Accuracy metric, not agreement with HP.}
\label{fig:supp-def-binary-heatmap}
\end{figure}

\begin{figure}[!htbp]
\centering
\includegraphics[width=0.85\linewidth]{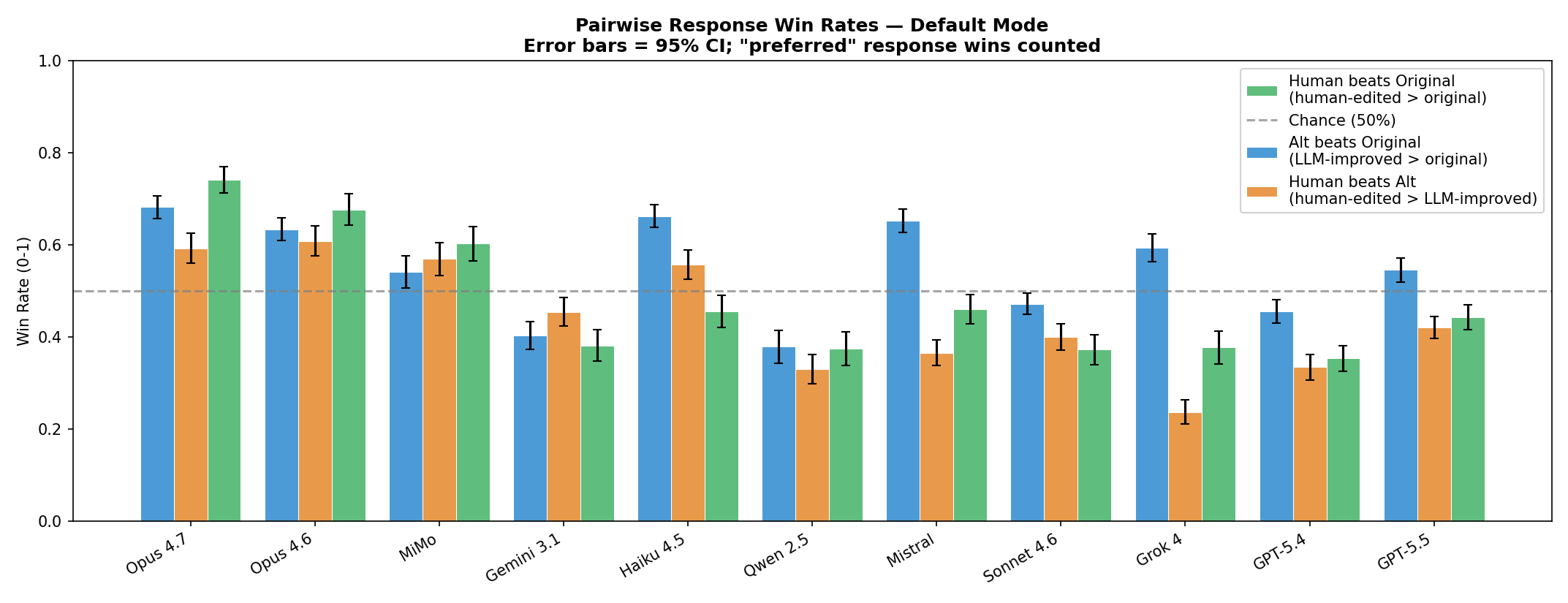}
\caption{Pairwise response win rates per model (Default Mode). For each model, three bars show the fraction of conversations in which the ``preferred'' response variant wins a head-to-head selection by the EM: Alt (LLM-improved) over Original (blue), Human-edited over Alt (orange), and Human-edited over Original (green). Error bars are 95\% CIs; dashed line is chance (0.5). Note that the ``winner'' label is assigned by canonical variant tier (Human > Alt > Original), not by per-turn HP preference.}  
\label{fig:supp-def-pwa-matrix}
\end{figure}

Figures~\ref{fig:supp-def-turn-grid}--\ref{fig:supp-def-pwa-matrix} expand the Default Mode per-metric breakdowns. The turn-level grid (Figure~\ref{fig:supp-def-turn-grid}) and post-conversation grid (Figure~\ref{fig:supp-def-post-grid}) show all per-model rankings on the metrics aggregated into the Composite. The four-branch per-branch decomposition (Figure~\ref{fig:supp-def-fourbranch-per}) supports the main paper's observation that ``Understanding'' is the hardest branch across all models. The pairwise win-rate matrix (Figure~\ref{fig:supp-def-pwa-matrix}) and binary-agreement heatmap (Figure~\ref{fig:supp-def-binary-heatmap}) show the per-pair structure underlying the discrimination claims in Section~\ref{sec:analysis} of the main paper. The Composite ranking itself appears in main-paper Figure~\ref{fig:composite-pairwise} and is not re-plotted here.

\FloatBarrier

\subsection{Verbose Mode Figure Gallery}
\label{sec:supp-verbose-gallery}

\begin{figure}[!htbp]
\centering
\includegraphics[width=0.6\linewidth]{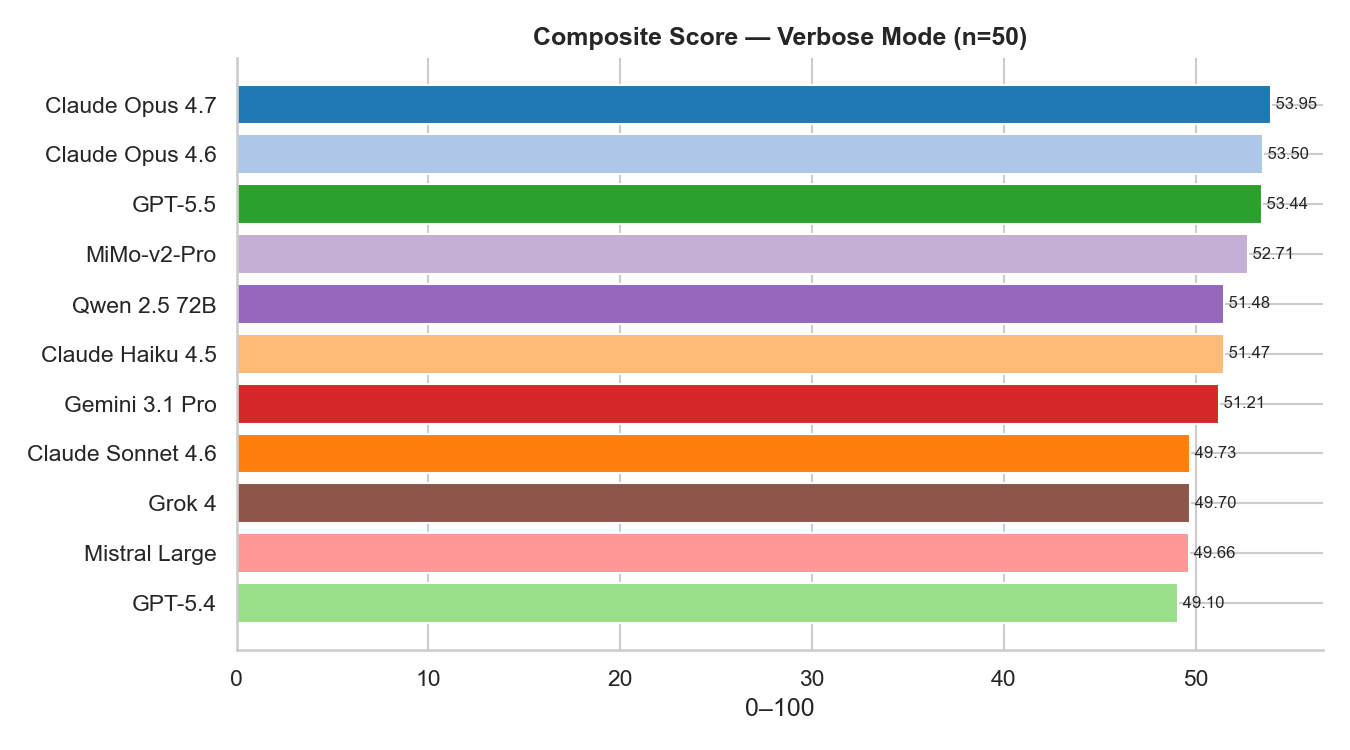}
\caption{Composite Score per model under Verbose Mode evaluation.}
\label{fig:supp-verb-composite}
\end{figure}

\begin{figure}[!htbp]
\centering
\includegraphics[width=\linewidth]{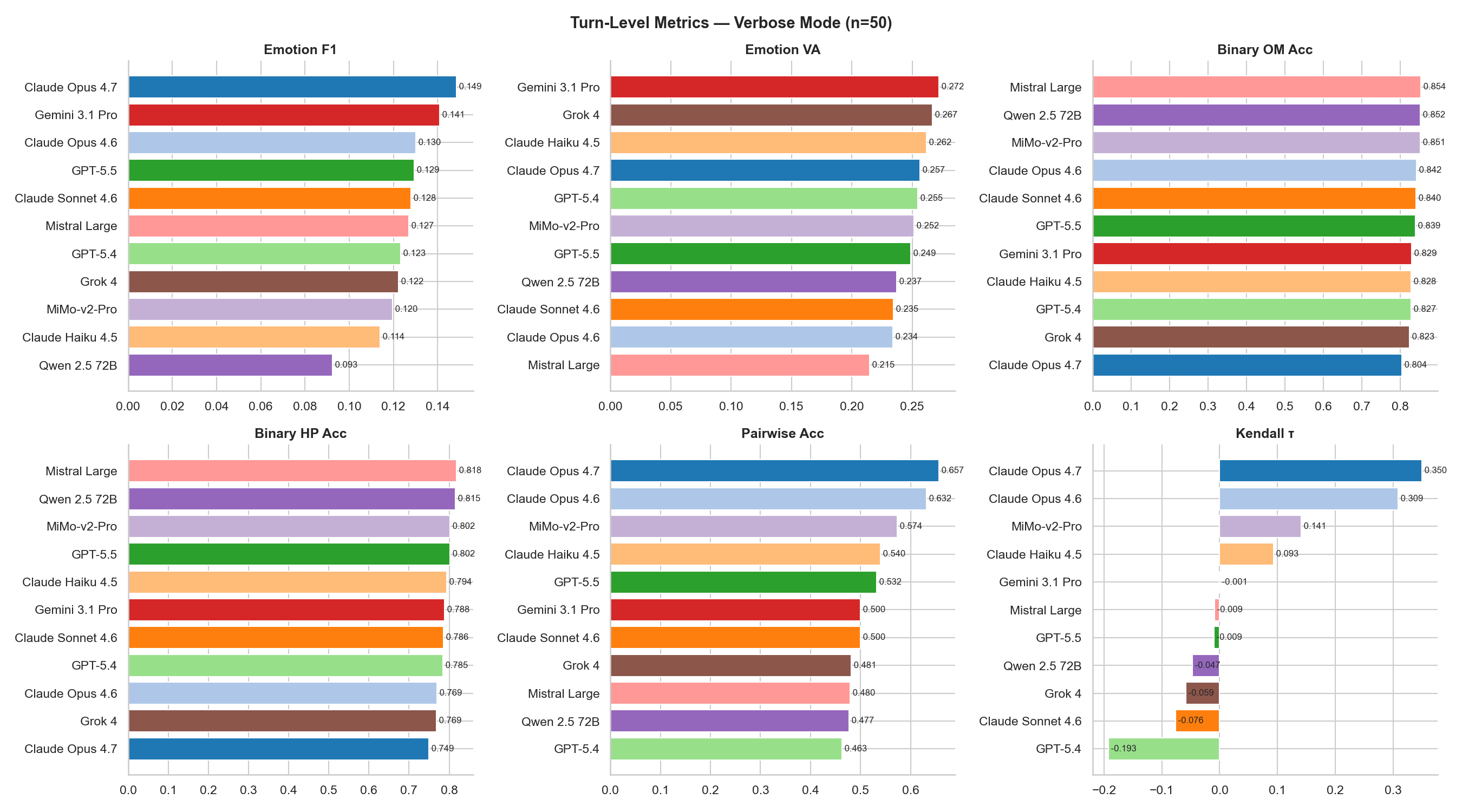}
\caption{Turn-level metric grid under Verbose Mode.}
\label{fig:supp-verb-turn-grid}
\end{figure}

\begin{figure}[!htbp]
\centering
\includegraphics[width=\linewidth]{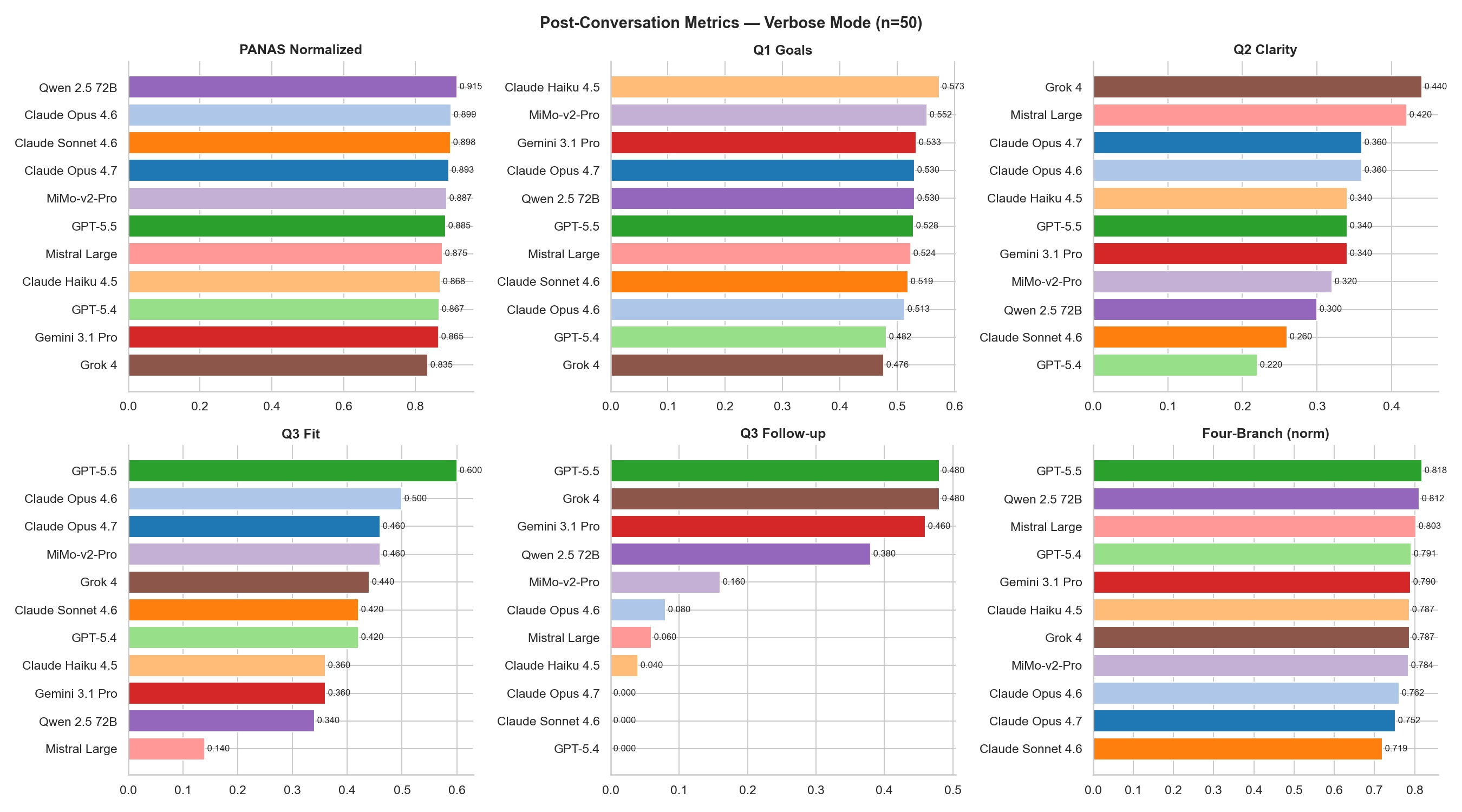}
\caption{Post-conversation metric grid under Verbose Mode.}
\label{fig:supp-verb-post-grid}
\end{figure}

The Verbose Mode evaluation runs on a 50-conversation subsample (vs.\ 200 in Default Mode), so absolute values in this gallery are noisier than the Default Mode set. The figures support the Verbose-mode findings discussed in main-paper Section~\ref{sec:analysis} and detailed in Section~\ref{sec:supp-mode-detail}: the Mistral~Large preference-prediction degradation (Figure~\ref{fig:supp-verb-turn-grid}, Pairwise/Kendall~$\tau$ panels) and the Opus~4.6~$\leftrightarrow$~4.7 within-family rank flip on Composite (Figure~\ref{fig:supp-verb-composite}). Post-conversation metrics (PANAS, Q1--Q3, Four-Branch) under Verbose Mode are in Figure~\ref{fig:supp-verb-post-grid}.

\FloatBarrier

\subsection{Omniscient Mode Figure Gallery}
\label{sec:supp-omn-gallery}

\begin{figure}[!htbp]
\centering
\includegraphics[width=\linewidth]{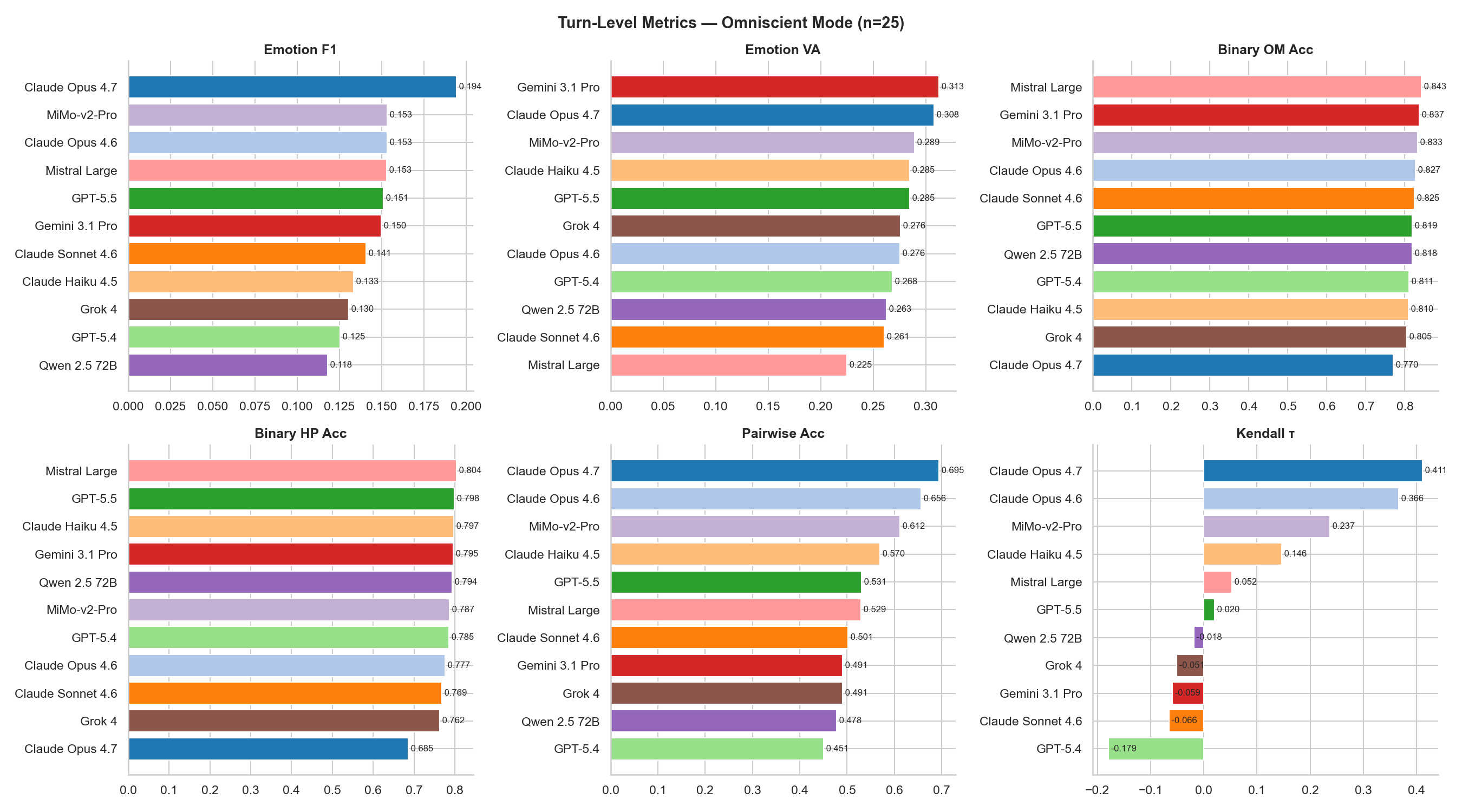}
\caption{Turn-level metric grid under Omniscient Mode.}
\label{fig:supp-omn-turn-grid}
\end{figure}

\begin{figure}[!htbp]
\centering
\includegraphics[width=\linewidth]{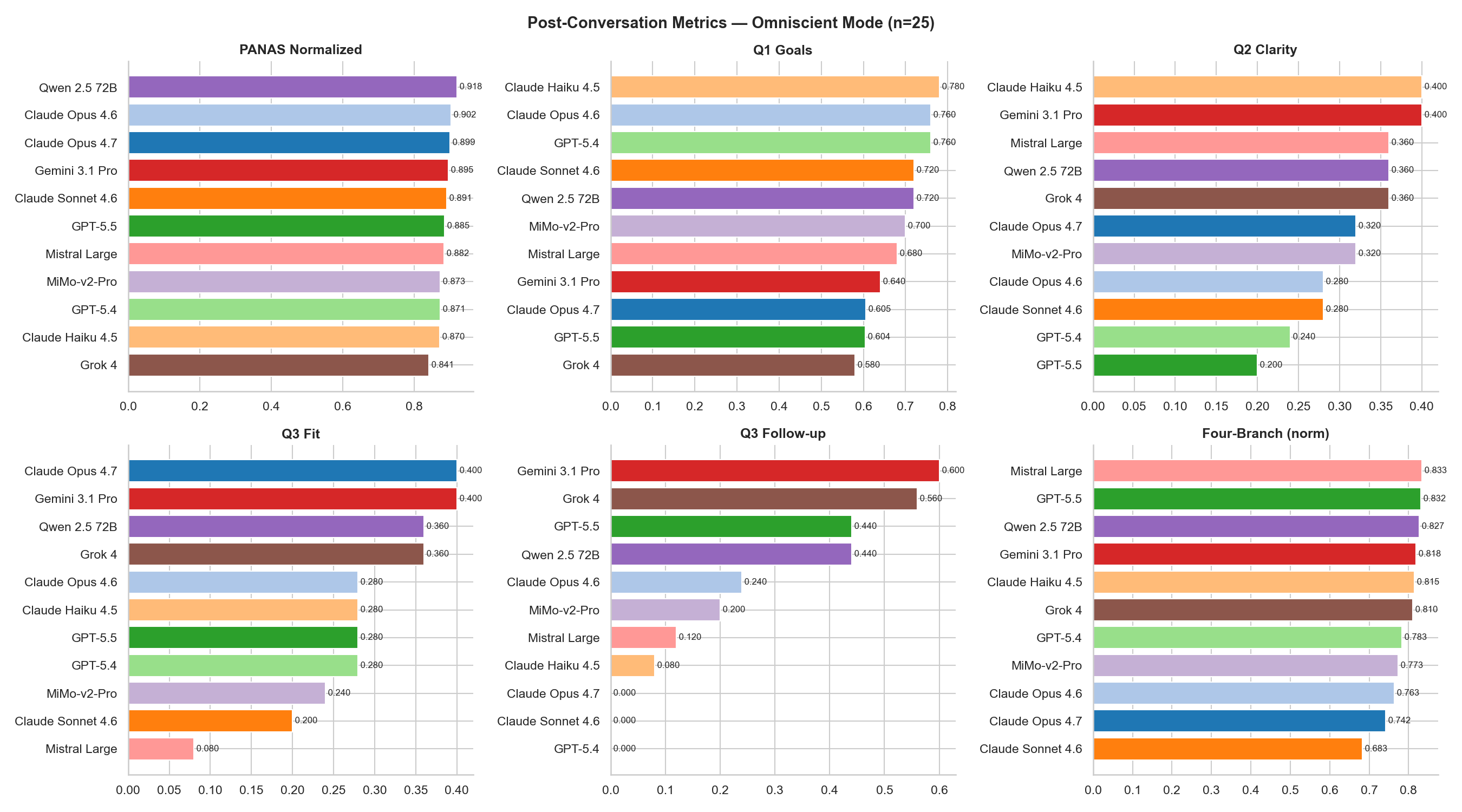}
\caption{Post-conversation metric grid under Omniscient Mode.}
\label{fig:supp-omn-post-grid}
\end{figure}

The Omniscient Mode evaluation provides each EM with the HP psychometric profile at evaluation time on a 25-conversation subsample. The figures support the Omniscient-mode findings discussed in main-paper Section~\ref{sec:analysis} and Section~\ref{sec:supp-mode-detail}: the Opus~4.7 Draft~Judge collapse (Figure~\ref{fig:supp-omn-post-grid}, Draft~Judge panel) is concurrent with Opus~4.7's strongest Pairwise/Kendall~$\tau$ values across any model in any mode (Figure~\ref{fig:supp-omn-turn-grid}), supporting the main paper's framing of the Judge drop as an output-format artifact rather than a real degradation in inference quality.

\FloatBarrier

%% ════════════════════════════════════════════════════════════════════════════

\section{Cost and Runtime}
\label{app:cost}

Table~\ref{tab:cost-runtime} reports measured API cost and cumulative call duration per model for each evaluation mode. Costs are dollar amounts charged by the respective providers during data collection (early 2026), aggregated from per-conversation logs in the runner output. Durations are the cumulative time spent inside API calls across all conversations for that model and mode; actual wall-clock with concurrent calls is substantially shorter, as the runner issues calls in parallel where rate limits permit.

\begin{table}[H]
\centering
\caption{Per-model API cost and cumulative call duration. Default Mode: $n = 200$ conversations; Verbose Mode: $n = 50$; Omniscient Mode: $n = 25$. ``Total \$'' is the sum across all three modes; the \textbf{full-benchmark cost across all 11 models is approximately \$1{,}679}. Models sorted by Default-mode cost per conversation.}
\label{tab:cost-runtime}
\small
\setlength{\tabcolsep}{4pt}
\begin{tabular}{l rr rr rr r}
\toprule
 & \multicolumn{2}{c}{\textbf{Default}} & \multicolumn{2}{c}{\textbf{Verbose}} & \multicolumn{2}{c}{\textbf{Omniscient}} & \\
\textbf{Model} & \textbf{\$/conv} & \textbf{hrs} & \textbf{\$/conv} & \textbf{hrs} & \textbf{\$/conv} & \textbf{hrs} & \textbf{Total \$} \\
\midrule
Opus 4.7      & 1.62 & 9.0  & 1.67 & 1.9 & 1.63 & 0.7 & 448.50 \\
Opus 4.6      & 1.23 & 18.8 & 1.31 & 2.7 & 1.29 & 2.4 & 343.99 \\
GPT-5.5       & 0.83 & 14.9 & 0.91 & 4.8 & 0.85 & 2.1 & 231.51 \\
Sonnet 4.6    & 0.74 & 11.4 & 0.78 & 2.3 & 0.78 & 1.1 & 207.21 \\
Gemini 3.1    & 0.64 & 24.7 & 0.69 & 6.0 & 0.67 & 2.6 & 179.02 \\
GPT-5.4       & 0.29 & 8.5  & 0.31 & 2.2 & 0.27 & 1.0 & 80.27  \\
Haiku 4.5     & 0.25 & 6.7  & 0.35 & 3.5 & 0.26 & 0.8 & 74.38  \\
Grok 4        & 0.17 & 4.1  & 0.22 & 1.0 & 0.17 & 0.5 & 49.24  \\
MiMo-v2-Pro   & 0.11 & 11.4 & 0.11 & 2.1 & 0.11 & 1.8 & 29.19  \\
Mistral Lg    & 0.09 & 12.6 & 0.11 & 5.3 & 0.09 & 2.2 & 25.55  \\
Qwen 2.5 72B  & 0.04 & 15.1 & 0.03 & 3.1 & 0.04 & 1.7 & 10.15  \\
\midrule
\multicolumn{7}{r}{\textbf{Total (all models): \$1{,}201.52 (Default) + \$323.40 (Verbose) + \$154.34 (Omniscient)}} & \textbf{\$1{,}679.27} \\
\bottomrule
\end{tabular}
\end{table}

\paragraph{Cost spread across models.}
Per-conversation Default Mode cost spans approximately 40$\times$ from Qwen 2.5 72B (\$0.04) to Opus 4.7 (\$1.62). The variation is dominated by per-token API pricing rather than call count: every model receives the same call structure under the runner protocol (5 calls per turn plus 1 post-conversation call, totaling roughly 26 calls for a 5-turn conversation). Input tokens scale with growing conversation history across turns, so token usage is approximately matched across models for the same conversation; the cost difference reflects each provider's input/output token rates.

\paragraph{Mode-level cost differences.}
Verbose Mode adds reasoning traces but does not change the call count, so per-conversation cost increases primarily via additional output tokens for the traces (visible by comparing the Default and Verbose \$/conv columns; the increment is typically a few percent of total cost). Omniscient Mode adds participant background to the system prompt, slightly increasing input-token cost without affecting the call count.

\paragraph{Wall-clock vs.\ cumulative call duration.}
The ``hrs'' columns sum sequential call duration across all conversations for that (model, mode) pair. Real wall-clock with concurrent calls is substantially shorter --- typical Default Mode runs completed in 1--3 wall-clock hours per model on a single workstation, depending on per-provider concurrency limits. Mode-by-mode the cumulative-duration figures are most useful as a relative reference: Gemini 3.1 (slowest at 24.7 hrs cumulative Default) runs approximately 6$\times$ longer per call than Grok 4 (fastest at 4.1 hrs).